\documentclass[twoside,11pt]{article}

\usepackage[preprint]{jmlr2e}
\usepackage{multirow}
\usepackage[utf8]{inputenc}
\usepackage[T1]{fontenc}
\usepackage{url}
\usepackage{booktabs}
\usepackage{amsfonts}
\usepackage{nicefrac}
\usepackage{microtype}
\usepackage{xcolor}
\usepackage{longtable}
\usepackage{amsmath}
\usepackage{subcaption}
\usepackage{algorithm}
\usepackage{algorithmic}
\usepackage{lastpage}

\newif\ifdraftlinks
\draftlinkstrue 

\ifdraftlinks
  \hypersetup{
    colorlinks=true,
    citecolor=blue,
    linkcolor=blue,
    urlcolor=blue
  }
\else
  \hypersetup{hidelinks}
\fi

\jmlrheading{0}{2026}{1-\pageref{LastPage}}{8/26}{}{under-review}{Minh-Khoi Pham, Luca Cotugno, Dan Cernei, Alina S\^irbu, Stefano Masi, Giuseppe Prencipe, Alessandro Pingitore, Patrizia Landi, Working Group on Uric Acid and Cardiovascular Risk of the Italian Society of Hypertension, Tai Tan Mai, Martin Crane, and Marija Bezbradica}
\ShortHeadings{TabFMs for Time-to-Event Prediction}{Minh-Khoi Pham et al.}
\firstpageno{1}

\begin{document}

\title{Adaptation Interfaces for In-Context Tabular Foundation Models in Time-to-Event Prediction}

\author{\name Minh-Khoi Pham\thanks{Corresponding author} \\
\email minhkhoi.pham@adaptcentre.ie \\
\addr ADAPT Centre, School of Computing, Dublin City University\\
DCU Glasnevin Campus, Dublin 9, D09 V209, Ireland
\AND
\name Luca Cotugno, Dan Cernei, Alina S\^irbu \\
\email \{luca.cotugno, dan.cernei\}@studio.unibo.it, alina.sirbu@unibo.it \\
\addr Department of Computer Science and Engineering, University of Bologna, Italy
\AND
\name Stefano Masi, Giuseppe Prencipe \\
\email \{stefano.masi, giuseppe.prencipe\}@unipi.it \\
\addr University of Pisa, Italy
\AND
\name Alessandro Pingitore, Patrizia Landi \\
\email \{alessandro.pingitore, patrizia.landi\}@cnr.it \\
\addr Institute of Clinical Physiology, CNR, Pisa, Italy
\AND
\name Tai Tan Mai, Martin Crane, Marija Bezbradica \\
\email \{tai.mai, martin.crane, marija.bezbradica\}@dcu.ie \\
\addr School of Computing, Dublin City University, Ireland
}

\maketitle

\begin{abstract}%
Tabular foundation models (TabFMs) achieve strong performance on structured data, particularly for standard classification and regression problems. Yet, extending them to censored time-to-event prediction is challenging because it requires properly handling censoring and event-time dynamics. Building on our prior work, we further link TabFMs with CoxPH and DeepHit and revise the context-resampled training procedure. We evaluate temporal zero-shot reformulation, classification-based fine-tuning, and survival-head adaptation using frozen TabFM backbones on 74 single-risk data sets, and we additionally study 4 competing-risk data sets. Zero-shot inference is effective on smaller single-risk data sets, whereas supervised adaptation becomes increasingly advantageous as data sets scale. Cox provides the most reliably strong interface, especially for Integrated Brier Score (IBS) on larger data sets. DeepHit is relatively stronger for the time-dependent Concordance Index ($C_{td}$) than for IBS, while cause-specific MTLR ranks highest among the TabFM survival heads in the four-data-set competing-risk analysis. Classification fine-tuning becomes more competitive with zero-shot inference as data sets grow but remains weaker for probabilistic prediction. Overall, our results indicate that effective TabFM transfer depends on the data regime and on the statistical structure represented by the chosen adaptation interface. The implementation scripts used for this work are available at \url{https://github.com/kaylode/survival-fm}.
\end{abstract}


\section{Introduction}
\label{sec:introduction}

Time-to-event prediction addresses not only if an event happens, but also the timing of that event, and is widely applied in domains such as healthcare, finance, and marketing, where making predictions from heterogeneous tabular data remains an ongoing challenge~\citep{chi2026obstructive}. Traditional survival-analysis methods, including Kaplan--Meier estimation and Cox proportional hazards (CoxPH), continue to serve as key baselines~\citep{kaplan1958nonparametric,cox1972regression}, but they provide limited flexibility for capturing complex relationships between covariates and time. As a result, a wide variety of machine-learning methods have been developed for censored time-to-event prediction.

Tabular foundation models (TabFMs), including TabPFN~\citep{hollmann2023tabpfn}, TabDPT~\citep{ma2025tabdpt}, and TabICL~\citep{qu2025tabicl}, provide pretrained priors that transfer across heterogeneous tabular prediction tasks. Their standard interfaces, however, assume fully observed classification or regression targets. Time-to-event prediction violates this interface in several ways: outcomes may be censored, prediction targets depend on time, and competing-risk settings involve multiple mutually exclusive event types. The challenge is therefore not simply whether a TabFM can be used for time-to-event prediction, but how its pretrained prior should be adapted to a downstream task with fundamentally different statistical structure. This calls for learning objectives or prediction mechanisms that differ from standard classification and regression formulations~\citep{kalbfleisch2002statistical}. 

Existing work has approached this mismatch from two directions. One strategy reformulates survival prediction as censoring-aware classification over discretized time horizons, allowing pretrained tabular classifiers to be used directly without changing their underlying task interface~\citep{kim2026tabular}. This provides a convenient route for generic classifiers, but introduces subject--time expansion, class imbalance, dependence on the choice of discretization horizons, and the native interface constraints of the underlying classifier. Another direction incorporates time-to-event structure during pretraining itself, producing prior-fitted foundation models specifically designed for survival inference~\citep{seletkov2026survival,qi2026survivalpfn}. In our previous work~\citep{pham2026tabular}, we established multi-task logistic regression (MTLR)-head adaptation over frozen TabFM representations and evaluated it on standard public clinical survival benchmarks and two large ICU cohorts. That study was clinically focused, single-risk, and restricted to MTLR as the survival-specific TabFM head. The present work treats that MTLR interface as a prior baseline and asks how alternative adaptation interfaces behave across a substantially broader benchmark, additional survival objectives, probabilistic evaluation, and competing risks.

In this work, we study time-to-event prediction with pretrained TabFMs as an \emph{adaptation interface problem under objective mismatch}. We compare three interfaces for transferring a generic pretrained tabular prior to censored prediction: (i) the existing zero-shot horizon-wise reformulation; (ii) classification fine-tuning on temporally expanded survival examples; and (iii) survival-head adaptation, where CoxPH and DeepHit are compared directly with the previously established MTLR interface. In the supervised interfaces, the pretrained backbone remains frozen and adaptation is performed through the task-specific classification or survival head. Our central question is therefore: \emph{which adaptation interfaces best transfer generic pretrained TabFMs to time-to-event prediction?}

\clearpage

Our main contributions are:
\begin{itemize}
    \item We formulate the transfer of generic TabFMs to time-to-event prediction as an adaptation-interface problem and systematically compare existing zero-shot reformulation, censoring-aware classification fine-tuning, and survival-head adaptation across a broad multi-domain benchmark.
    \item We extend survival-head adaptation beyond the previously studied MTLR interface to CoxPH and DeepHit under context-resampled, context-conditioned training, and compare these heads in terms of discrimination and probabilistic prediction across data regimes while retaining MTLR as a reference baseline.
    \item We extend the analysis to competing risks, comparing survival-head interfaces only, and distinguish cause-specific adaptation from jointly normalized cause--time modeling to test whether the conclusions from single-risk transfer persist when multiple event types are present.
\end{itemize}

\section{Related Work}
\label{sec:related}

\subsection{Modern Survival Modeling}
Survival analysis has progressed from classical statistical methods such as Kaplan--Meier estimation, CoxPH, and tree-based survival models toward increasingly flexible machine-learning approaches for censored time-to-event prediction~\citep{cox1972regression,ishwaran2008random}. Deep survival models including DeepSurv, MTLR, DeepHit, and related continuous- and discrete-time extensions expand this design space by relaxing proportional-hazards assumptions, modeling individualized event-time distributions, and supporting competing risks through specialized objectives~\citep{katzman2018deepsurv,yu2011learning,kvamme2019time-to-event,lee2018deephit,nagpal2021deep}. More recent architectures incorporate transformers, attention mechanisms, latent-variable formulations, and longitudinal encoders~\citep{wang2022survtrace,mesinovic2026dysurv}. These methods show the importance of survival-specific statistical structure, but they are predominantly trained from scratch for individual tasks.

\subsection{Foundation Models for Tabular Learning}
\label{subsec:tabfm_related}
TabFMs shift structured-data prediction from fitting a separate model for each data set toward broad pretraining followed by in-context prediction. The present benchmark uses three representative pretrained backbones with deliberately different pretraining and inference designs. \textbf{TabPFN v2} is a prior-fitted transformer pretrained on synthetic tabular tasks and performs prediction by conditioning on labeled examples supplied in context~\citep{hollmann2025tabpfn}. \textbf{TabDPT} combines real-data pretraining with retrieval-based conditioning, explicitly using retrieved examples to support transfer to unseen tables~\citep{ma2025tabdpt}. \textbf{TabICL v1} uses a two-stage architecture that first builds fixed-dimensional row embeddings through column-then-row processing and then applies a transformer for scalable in-context learning~\citep{qu2025tabicl}. These are the exact model generations used in our experiments; for ease of reading, we refer to them hereafter without version numbers: TabPFN, TabDPT, and TabICL.


The appeal of these models is that a reusable pretrained prior can reduce task-specific optimization and perform strongly in small- and medium-data regimes. At the same time, recent broader evaluations caution against treating this advantage as universal. BeyondArena finds that current TabFMs are particularly strong on tiny- to medium-sized IID problems, whereas conventional tree ensembles and supervised deep models can regain an advantage on non-IID, large, or high-dimensional data sets~\citep{purucker2026beyondiid}. Adaptation and retrieval have therefore become increasingly important for extending pretrained tabular priors beyond their most favorable operating regime~\citep{liu2025tabpfn,pham2026retrieval-aligned}. Native in-context interfaces can also impose architectural constraints; for example, TabFM's native classifier is limited to a fixed number of output classes in its released interface~\citep{kong2026tabfm}. Such constraints motivate distinguishing direct reuse of a pretrained prediction interface from adaptation strategies that retain the backbone while replacing the downstream head. Their standard objectives, however, still do not directly represent censoring or event-time distributions.

\subsection{Foundation Models for Censored Time-to-Event Prediction}
Recent work addresses the mismatch between generic foundation model interfaces and censored time-to-event prediction through several distinct routes. \citet{kim2026tabular} reformulate survival analysis as a sequence of censoring-aware binary classification problems, allowing existing TabFMs to be used directly through their pretrained classification interface. Because this route leaves the native classifier unchanged, it also inherits backbone-specific interface restrictions; the horizon-wise formulation itself is binary, but the broader output interface remains fixed by the pretrained classifier. In contrast, Survival In-Context (SIC)~\citep{seletkov2026survival} and SurvivalPFN~\citep{qi2026survivalpfn} introduce survival structure during prior-fitted pretraining, amortizing survival inference directly rather than adapting a generic classification-pretrained model post hoc. These approaches demonstrate the value of incorporating event timing directly into the foundation model learning objective, but require survival- or time-to-event-aware pretraining rather than reusing an otherwise generic tabular foundation model unchanged.


Our setting is complementary: rather than redesigning the pretraining distribution or requiring a survival-native foundation model, we study how generic pretrained TabFMs can be transferred through different downstream interfaces. Our earlier study~\citep{pham2026tabular} introduced MTLR-based adaptation over frozen TabFM representations and evaluated it alongside zero-shot inference on public clinical survival benchmarks and two large ICU cohorts. In the present work, that MTLR interface is retained as a reference baseline rather than presented as a new contribution. We remove the ICU cohorts from the benchmark and broaden the study to 74 heterogeneous single-risk data sets, alternative CoxPH and DeepHit adaptation heads, censoring-aware classification fine-tuning, probabilistic evaluation with IBS, context-resampled training, data-regime analysis, and competing-risk formulations.

\section{Methods}
\label{sec:methods}

We consider three transfer regimes: (i) zero-shot in-context reformulation, with no parameter updates; (ii) classification fine-tuning on temporally expanded examples; and (iii) survival-head adaptation with CoxPH, DeepHit, or the previously established MTLR interface~\citep{pham2026tabular}. The horizon-wise zero-shot mechanism follows prior work~\citep{kim2026tabular,pham2026tabular}. In both supervised regimes, the pretrained TabFM backbone remains frozen and only the task-specific classification or survival head is optimized. Temporal discretization and temporal expansion are distinct operations: classification adaptation expands each subject into multiple horizon-specific examples, whereas MTLR and DeepHit discretize the event-time distribution internally without horizon-wise data expansion. For discrete-time formulations, the time intervals are selected adaptively from the observed training event-time distribution. This allows finer temporal resolution when observed events are abundant while avoiding sparsely supervised bins on smaller data sets.

\subsection{Problem Setup}

We consider right-censored survival data $\mathcal{D}=\{(\mathbf{x}_i,\tilde T_i,\delta_i)\}_{i=1}^N$, where $\mathbf{x}_i\in\mathbb{R}^d$ denotes covariates, $T_i$ is the event time, $C_i$ is the censoring time, $\tilde T_i=\min(T_i,C_i)$ is the observed time, and $\delta_i=\mathbf{1}[T_i\le C_i]$ indicates whether the event was observed. In competing-risk settings, we replace the binary event indicator with $\Delta_i\in\{0,1,\ldots,C\}$, where $\Delta_i=0$ denotes censoring and $\Delta_i=c$ denotes an observed event of cause $c$. The target is either the survival function $S(t\mid\mathbf{x})=P(T>t\mid\mathbf{x})$ or, for cause $c$, the cumulative incidence function $F_c(t\mid\mathbf{x})=P(T\le t,\Delta=c\mid\mathbf{x})$. To assess generalization while tuning model hyperparameters, we adopt a nested cross-validation scheme in which each outer fold is further split into training and validation subsets $\mathcal{D}_{\mathrm{train}}$ and $\mathcal{D}_{\mathrm{val}}$, with the remaining outer fold reserved for testing $\mathcal{D}_{\mathrm{test}}$.

Let $f_{\theta^\star}$ denote a pretrained in-context learning TabFM. In zero-shot inference, $\theta^\star$ remains fixed and predictions are obtained through the pretrained classification interface. For both classification fine-tuning and survival-head adaptation, the reported experiments also keep $\theta^\star$ fixed but optimize a new task-specific head. The resulting representations remain context dependent because the pretrained TabFM is conditioned on sampled labeled support examples rather than used as a one-time deterministic feature extractor.

\subsection{Context-Conditioned Training and Inference}
Context conditioning is shared by the classification and survival-head interfaces. Our earlier MTLR study conditioned each representation on the training cohort as a fixed support set~\citep{pham2026tabular}. Here, supervised training instead uses sampled support contexts and resamples them across optimization steps, making the downstream head robust to variation in the in-context support while preventing a query from appearing in its own context. Specifically, each query minibatch $\mathcal{B}\subset\mathcal{D}_{\mathrm{train}}$ is paired with
\[
\mathcal{D}_{\mathrm{context}}^{\mathrm{train}}\subseteq\mathcal{D}_{\mathrm{train}}\setminus\mathcal{B}.
\]

For \textbf{survival-head adaptation}, each sampled support subject is supplied to the pretrained backbone through its native labeled-context interface. Because the pretrained TabFMs expect classification-style labels rather than censored time-to-event targets, we encode the support label as the binary observed-event indicator $e_j=\mathbf{1}[\delta_j>0]$.
For a sampled support index set $\mathcal C$, the query representation is therefore
\[
\mathbf{h}_i=f_{\theta^\star}\!\left(\mathbf{x}_i;\{(\mathbf{x}_j,e_j):j\in\mathcal C\}\right).
\]
The observed time $\tilde T_j$ is not supplied as a context label to the pretrained backbone; event-time and censoring information instead enter through the downstream CoxPH, MTLR, or DeepHit objective used to optimize the survival head. In competing-risk settings, all observed causes are mapped to $e_j=1$, so cause identity is introduced only by the competing-risk survival head. Contexts are resampled across optimization steps, so even with frozen pretrained parameters the representation of the same query can vary under different labeled support samples.

The \textbf{classification-based adaptations} use a different labeled context because they operate on temporally expanded subject--horizon examples. A generic classification backbone has no built-in representation of event time, so each classification query is augmented with fixed temporal features representing the requested horizon. We denote the resulting input by $\tilde{\mathbf{x}}_{ik}$. For single-risk data, support examples take the form
\[
\left(\tilde{\mathbf{x}}_{jk},y_{jk}\right),
\qquad
y_{jk}=\mathbf{1}[\delta_j=1\land \tilde T_j\le t_k],
\]
with censored subject--horizon pairs retained only while their status is known. Thus, the single-risk classification context label indicates whether an observed event has occurred by the queried horizon, whereas the survival-head interface uses only the subject-level observed-event indicator to condition the pretrained representation. The competing-risk extension uses multiclass cause labels as defined in Section~\ref{sec:competing_methods}. The detailed temporal expansion and validity mask are defined in Section~\ref{sec:classification}.

Validation queries come from $\mathcal{D}_{\mathrm{val}}$ during fine-tuning and from $\mathcal{D}_{\mathrm{test}}$ during inference, while their support context is sampled only from the training set. Neither validation nor test observations are ever included in a support context,
\[
\mathcal{D}_{\mathrm{context}}^{\mathrm{val}}\subseteq\mathcal{D}_{\mathrm{train}}, 
\qquad
\mathcal{D}_{\mathrm{context}}^{\mathrm{test}}\subseteq\mathcal{D}_{\mathrm{train}}.
\]

The two classification-based interfaces use the temporally augmented input $\tilde{\mathbf{x}}_{ik}$ but differ in whether a downstream head is learned. The underlying zero-shot mechanism is not new to this work: as in prior horizon-wise formulations~\citep{kim2026tabular,pham2026tabular}, expanded training examples act as labeled in-context demonstrations and no parameters are optimized. For a single-risk test query,
\[
\hat p_{ik}=g_{\mathrm{native}}\!\left(f_{\theta^\star}(\tilde{\mathbf{x}}_{ik};\mathcal{D}_{\mathrm{context}})\right),
\qquad
\hat S(t_k\mid\mathbf{x}_i)=1-\hat p_{ik},
\]
and the horizon-wise survival probabilities are interpolated to the evaluation grid. In \emph{classification fine-tuning}, the pretrained backbone is frozen and a task-specific classification head is optimized with the censoring-aware objective defined in Section~\ref{sec:classification}. Replacing the native head removes dependence on its fixed output dimensionality; the single-risk horizon-wise target is binary, while the competing-risk extension is multiclass.

Survival-head adaptation uses the same train/validation context separation but operates on subject representations rather than subject--horizon classification queries. For a test subject $\mathbf{x}_\ast\in\mathcal{D}_{\mathrm{test}}$,
\[
\mathbf{h}_\ast^{(r)}=f_{\theta^\star}\!\left(\mathbf{x}_\ast;\{(\mathbf{x}_j,e_j):j\in\mathcal C^{(r)}\}\right),
\qquad
\mathcal C^{(r)}\subseteq\mathcal{D}_{\mathrm{train}},
\]
and the fitted survival head produces $\hat S^{(r)}(t\mid\mathbf{x}_\ast)$, or $\hat F_c^{(r)}(t\mid\mathbf{x}_\ast)$ for competing risks. No model parameters are updated during validation or test inference. When context ensembling is enabled, inference is repeated over separately sampled training contexts and the predictions are averaged,
\[
\hat S(t\mid\mathbf{x}_\ast)=\frac{1}{R}\sum_{r=1}^{R}\hat S^{(r)}(t\mid\mathbf{x}_\ast),
\]
with analogous averaging applied independently to each cause-specific cumulative incidence function. The reported benchmarks use $R=5$ context samples for both single-risk and competing-risk inference. The zero-shot interface averages over the same number of support samples, but through its native in-context mechanism: the pretrained classifier is fed with each of five stratified subsamples of the expanded context and the predicted probabilities are averaged. Zero-shot inference is also run per horizon, so a prediction requires one in-context fit per time bin (i.e., a query with 100 time bins requires 100 forward passes).

For evaluations that require a scalar risk score, predicted trajectories are summarized after inference. In single-risk zero-shot inference,
\[
r_{\mathrm{ZS}}(\mathbf{x})=1-\hat S(t_{\max}\mid\mathbf{x}),
\]
whereas the supervised single-risk interfaces use the negative area under the predicted survival curve,
\[
r_{\mathrm{sup}}(\mathbf{x})=-\int \hat S(t\mid\mathbf{x})\,dt,
\]
so shorter predicted survival corresponds to larger risk. For supervised competing-risk models, the analogous cause-specific score integrates each predicted cumulative incidence function over the evaluation grid and normalizes by the grid span. These scalar summaries are used only by evaluations that require a one-dimensional risk score; the reported Antolini $C_{td}$ is computed directly from the predicted survival trajectories. IBS is likewise computed from the survival or cumulative-incidence trajectories themselves.

\subsection{Temporal Classification Reformulation}
\label{sec:classification}

The horizon-wise censoring-aware classification mechanism follows the existing reformulation literature~\citep{kim2026tabular,pham2026tabular}. For the present benchmark, we use an adaptive, event-balanced discretization rather than fixing a common temporal resolution across data sets. Let $\mathcal{K}=\{10,20,30,50,100\}$ denote candidate numbers of intervals. For each candidate $K$, quantile boundaries are computed from the observed (uncensored) event times and the number of events falling in each interval is counted. The procedure selects the largest feasible resolution
\[
K^\star=\max_{K\in\mathcal K}\left\{K:\min_k n_k^{(E)}\ge m\right\},
\]
where $n_k^{(E)}$ is the number of observed events in interval $k$ and $m=5$ is the default minimum event support per interval. For the selected resolution $K^\star$, the initial quantile boundaries are
\[
q_j=Q_E\!\left(\frac{j}{K^\star}\right),\qquad j=0,\ldots,K^\star,
\]
where $Q_E$ is the empirical quantile function of the observed event times. We additionally include $t=0$ as the initial temporal coordinate and remove duplicate boundaries caused by tied event times. We therefore denote the effective ordered horizon set by
\[
\mathcal T=\{t_0,\ldots,t_{K-1}\}
=\operatorname{unique}\!\left(\{0\}\cup\{q_j\}_{j=0}^{K^\star}\right),
\]
where $K=|\mathcal T|$ is the effective number of temporal coordinates. This procedure provides finer temporal resolution when observed events are abundant and coarser resolution when event information is sparse, without tuning the number of bins separately for each data set. Because duplicate quantile boundaries are removed, the effective number of temporal coordinates $K$ can be smaller than the number implied by the requested resolution $K^\star$.

Each subject is then paired with each horizon to form time-augmented examples. For single-risk data, the binary target is
\[
y_{ik}=\mathbf{1}[\delta_i=1\land \tilde T_i\le t_k].
\]
For an observed event, all horizons remain informative: horizons before the observed event time are negative and horizons at or after the event are positive. For a censored subject, horizons before censoring are known negatives, whereas later horizons have unknown event status and are excluded. The implementation therefore uses the validity mask
\[
m_{ik}=\mathbf{1}[\delta_i=1\lor \tilde T_i>t_k].
\]
The masked binary cross-entropy (BCE) objective is
\[
\mathcal{L}_{\mathrm{CE}}
=\sum_{i,k}m_{ik}\,\mathrm{BCE}\!\left(\hat p_{ik},y_{ik}\right).
\]
To control the size and class imbalance of the expanded classification data, all positive subject--horizon examples are retained and negative examples are randomly subsampled. With sampling ratio $\rho$, at most $\rho N_{+}$ negative examples are retained, where $N_{+}$ denotes the number of positive expanded examples; the reported classification fine-tuning configuration uses $\rho=0.5$. This subsampling is distinct from the event-balanced minibatch sampler used during optimization.

This construction expands the data from $N$ subjects to up to $O(NK)$ subject--horizon examples before subsampling. Temporal discretization and temporal expansion are distinct: MTLR and DeepHit also use discretized event-time grids, but they optimize structured survival objectives without expanding subjects across horizon-specific classification examples, while CoxPH operates directly through continuous-time risk sets.

\subsection{Survival Head Adaptation}
\label{section:survival}

Survival-head adaptation freezes the pretrained TabFM parameters and optimizes a survival-specific prediction head over context-conditioned representations. Our earlier work already established the MTLR variant~\citep{pham2026tabular}; we therefore retain MTLR as a reference baseline and focus the methodological comparison here on extending the same frozen-backbone interface to CoxPH and DeepHit. Gradients update only the survival-head parameters $\Phi$ and are not propagated into $\theta^\star$. The head is a single hidden layer of 64 units with dropout, followed by the output parameterization required by the corresponding objective: a scalar log-risk for CoxPH, and a $K^\star$-dimensional logit vector for MTLR and DeepHit. CoxPH operates in continuous time. DeepHit is discrete-time and uses the adaptive event-balanced quantile discretization described above. For consistency in the benchmark, the MTLR baseline is evaluated under the same current data splits, preprocessing, context-resampling protocol, and adaptive discretization, but its standard single-risk formulation is not reintroduced here.

\subsubsection{Cox Proportional Hazards (CoxPH)}
The survival head outputs a scalar log-risk $r_i=g_\Phi(\mathbf h_i)$ and minimizes the negative partial log-likelihood
\[
\mathcal{L}_{\mathrm{Cox}}
=-\sum_{i:\delta_i=1}\left[r_i-\log\sum_{j\in\mathcal{R}(T_i)}\exp(r_j)\right],
\qquad
\mathcal{R}(T_i)=\{j:\tilde T_j\ge T_i\}.
\]
Censoring is handled through the event-indexed risk sets.

\subsubsection{Single-Risk DeepHit}
For single-risk data, DeepHit predicts a discrete probability mass function $p_{ik}=P(T\in I_k\mid\mathbf{x}_i)$ over time intervals and uses the pycox DeepHitSingle objective implemented in our pipeline:
\[
\mathcal{L}_{\mathrm{DH}}^{\mathrm{SR}}
=\alpha\mathcal{L}_{\mathrm{NLL}}+(1-\alpha)\mathcal{L}_{\mathrm{rank}}.
\]
For an observed event in interval $k_i$, the likelihood contribution is $-\log p_{i,k_i}$; for censoring in interval $k_i$, it is the negative log of the remaining tail probability. The pairwise ranking term encourages subjects with earlier observed events to receive larger cumulative event probability at the corresponding horizon. Thus, the single-risk implementation combines discrete event-time likelihood and ranking supervision.

\subsection{Competing-Risk Adaptation}
\label{sec:competing_methods}

The competing-risk implementations use two distinct survival-head strategies that should not be conflated: cause-specific modeling for Cox and MTLR, and a joint cause--time distribution for DeepHit. As in the single-risk survival-head variants, the TabFM backbone remains frozen while the corresponding task-specific output layers are optimized. The horizon-wise classification interfaces admit a natural multiclass extension, described below for completeness, but we do not report competing-risk results for it; the reasons are given in Section~\ref{sec:competing_results}.

\subsubsection{Multiclass Horizon Reformulation}
For competing-risk classification, class $0$ denotes that no observed event has occurred by horizon $t_k$, while classes $1,\ldots,C$ identify the observed cause. For a valid subject--horizon pair,
\[
y^{\mathrm{CR}}_{ik}=
\begin{cases}
\Delta_i, & \Delta_i>0 \text{ and } \tilde T_i\le t_k,\\
0, & \tilde T_i>t_k.
\end{cases}
\]
As in the single-risk construction, censored subject--horizon pairs after $\tilde T_i$ are excluded because their status is unknown. At inference, the probability assigned to class $c$ at horizon $t_k$ provides the corresponding horizon-wise cause-$c$ event probability. Unlike the joint DeepHit formulation below, this horizon-wise classification construction does not define a single normalized probability mass jointly over cause and event time, and the cause-specific probabilities it produces are not constrained to be mutually consistent. 

\subsubsection{Cause-Specific Cox and MTLR}

For Cox and MTLR, competing risks are handled through separate cause-specific models. For each cause $c$, we define \[ \delta_i^{(c)}=\mathbf{1}[\Delta_i=c], \] so that events from the other causes are treated as censoring for the cause-specific task.

For Cox, the model produces one relative-risk score per cause and is optimized using the sum of the corresponding cause-specific partial likelihoods, where only events of cause $c$ contribute as cases to the $c$-th objective. At prediction time, the cause-specific Breslow baseline cumulative hazards are combined with the predicted relative risks to obtain cause-specific cumulative hazards. Overall survival is then reconstructed from the sum of these hazards, and cumulative incidence functions are obtained from the resulting hazard increments. 

For MTLR, the same cause-specific recoding is used to define a binary single-risk task for each cause, and an independent MTLR head is fitted using the corresponding single-risk discretization procedure. Because the cause-specific MTLR models are optimized independently, they do not impose a global normalization constraint across competing causes.

Thus, Cox and MTLR both use cause-specific competing-risk formulations, whereas the DeepHit variant below models a single joint distribution over cause--time outcomes.

\subsubsection{Joint Competing-Risk DeepHit}
The competing-risk DeepHit implementation predicts a joint cause--time probability mass function
\[
p_{c,k}(\mathbf{x})=P(\Delta=c,T\in I_k\mid\mathbf{x}),
\qquad
\sum_c\sum_k p_{c,k}(\mathbf{x})=1,
\]
using a joint softmax. Unlike the single-risk DeepHit implementation, the competing-risk loss used here is a joint likelihood without the pairwise ranking term. For an observed event of cause $c_i$ in interval $k_i$,
\[
\mathcal{L}_{\mathrm{obs},i}=-\log p_{c_i,k_i}(\mathbf{x}_i),
\]
whereas for censoring assigned to interval $k_i$ the implemented discrete-bin convention uses
\[
\mathcal{L}_{\mathrm{cens},i}
=-\log\left(\sum_c\sum_{k\ge k_i}p_{c,k}(\mathbf{x}_i)\right).
\]
The inclusive boundary preserves censoring supervision for observations clipped to the final discrete interval. The cumulative incidence for cause $c$ follows as $F_c(t_m\mid\mathbf{x})=\sum_{k\le m}p_{c,k}(\mathbf{x})$. This joint softmax is the formulation in our benchmark that most directly enforces probability conservation across competing causes and event times. The joint competing-risk time grid uses equal-width intervals over the observed event-time range, in contrast to the adaptive quantile-based cuts used by the single-risk discrete heads. Algorithm~\ref{alg:adaptation} summarizes the three adaptation interfaces and the points at which they differ: target construction, whether a downstream head is optimized, and which objective supplies the event-time and censoring information.

\begin{algorithm}[t]
\caption{Adaptation interfaces for transferring a pretrained TabFM to survival prediction}
\label{alg:adaptation}
\begin{algorithmic}[1]
\REQUIRE Pretrained TabFM $f_{\theta^\star}$; survival data $\mathcal{D}=\{(\mathbf{x}_i,\tilde T_i,z_i)\}_{i=1}^{N}$, where $z_i=\delta_i$ for single risk and $z_i=\Delta_i$ for competing risks; adaptation strategy $s\in\{\text{zero-shot},\text{CE},\text{survival-head}\}$
\IF{$s\in\{\text{zero-shot},\text{CE}\}$}
    \STATE \COMMENT{Classification interface; Section~\ref{sec:classification}}
    \STATE Select effective horizons $t_0<\cdots<t_{K-1}$ and construct temporally augmented inputs $\tilde{\mathbf{x}}_{ik}$
    \STATE For single risk, set $z_i=\delta_i$ and define $y_{ik}=\mathbf{1}[z_i=1\land\tilde T_i\le t_k]$ and $m_{ik}=\mathbf{1}[z_i=1\lor\tilde T_i>t_k]$
    \IF{$s=\text{zero-shot}$}
        \STATE $\hat p_{ik}\gets g_{\mathrm{native}}\!\left(f_{\theta^\star}(\tilde{\mathbf{x}}_{ik};\mathcal D_{\mathrm{context}})\right)$ \COMMENT{no parameter updates}
    \ELSE
        \STATE Freeze pretrained parameters $\theta^\star$
        \STATE Optimize the task-specific classification head on valid expanded examples
    \ENDIF
    \STATE Recover horizon-wise survival probabilities, or cause probabilities for competing risks
\ELSE
    \STATE \COMMENT{Survival head interface; Section~\ref{section:survival}}
    \STATE Freeze pretrained parameters $\theta^\star$
    \STATE Define native context labels $e_j=\mathbf{1}[z_j>0]$
    \STATE Sample $\mathcal C\subset\mathcal D_{\mathrm{train}}\setminus\mathcal B$ for query minibatch $\mathcal B$
    \STATE Compute $\mathbf h_i=f_{\theta^\star}(\mathbf x_i;\{(\mathbf x_j,e_j):j\in\mathcal C\})$
    \STATE Attach $g_\Phi\in\{\mathrm{CoxPH},\mathrm{MTLR},\mathrm{DeepHit}\}$ and optimize $\displaystyle \Phi^\star=\arg\min_{\Phi}\mathcal L_{\mathrm{surv}}\bigl(g_\Phi(\mathbf h_i),\tilde T_i,z_i\bigr)$
    \STATE Predict $\hat S(t\mid\mathbf x_i)$, or $\hat F_c(t\mid\mathbf x_i)$ for competing risks, from $g_{\Phi^\star}(\mathbf h_i)$
\ENDIF
\STATE During supervised training, use event-balanced minibatches and resample support contexts across optimization steps
\end{algorithmic}
\end{algorithm}

\section{Experimental Setup}
\label{sec:experiments}

\subsection{Data Set Taxonomy}
We curate a broad benchmark of public tabular survival data sets spanning diverse domains, sample sizes, censoring rates, and event structures. Following \citet{kim2026tabular}, the single-risk benchmark includes data sets from SurvSet~\citep{drysdale2022survset}. For competing risks, the reported main results use SUPPORT2-CR, FRAMINGHAM, PBC2, and SYNTHETIC~\citep{lee2018deephit}. Missing values are imputed using training-set medians; continuous covariates are clipped to the $[0.1\%,99.9\%]$ training quantile range and standardized using training-set statistics only. For the size-stratified analysis in Section~\ref{sec:single_results} we group data sets by cohort size into \emph{small} ($N<500$), \emph{medium} ($500\le N<4{,}000$), and \emph{large} ($N\ge 4{,}000$) regimes. Full data set characteristics are reported in Table~\ref{tab:datasets_survset} for the single-risk benchmark and Table~\ref{tab:datasets_cr} for the competing-risk benchmark.

For TabFM interfaces whose native feature capacity is smaller than the input dimensionality, we apply principal component analysis (PCA) before backbone inference. Within each outer cross-validation fold, PCA is fitted only on the training partition and the learned projection is applied unchanged to validation and test samples. If $d$ denotes the original number of covariates and $d_{\max}$ the feature capacity exposed by the corresponding backbone wrapper, PCA is applied only when $d>d_{\max}$, with the number of retained components additionally bounded by the available number of training observations. Data sets already satisfying the native feature-capacity constraint are passed without PCA reduction. For classification-based interfaces, the capacity calculation also reserves dimensions for the appended temporal features. Backbones that natively support the observed feature dimensionality, such as the TabICL used here, do not invoke this PCA reduction.

\subsection{Pretrained Backbones and Baselines}
We evaluate TabPFN, TabDPT, and TabICL under the adaptation framework above, together with non-pretrained neural and classical survival baselines. Supervised TabFM variants keep the pretrained backbone frozen; classification fine-tuning is denoted by -CE, while survival-head variants are denoted by -Cox, -DH, and -MTLR. The -MTLR variants implement the adaptation interface established in our earlier work~\citep{pham2026tabular} and are treated here as reference baselines under the current benchmark protocol. Other baselines include CoxPH, random survival forests (RSF), gradient-boosted survival models (GBSA), DeepSurv, SurvTRACE~\citep{wang2022survtrace}, and DySurv~\citep{mesinovic2026dysurv}. Two further non-pretrained neural baselines, denoted MLP-MTLR and MLP-DH, pair the same multilayer-perceptron trunk used by DeepSurv with the MTLR and DeepHit heads; they are the matched non-pretrained counterparts used for the backbone comparison in Section~\ref{sec:objective_backbone_effects}. The competing-risk benchmark adds SurvBoost, a gradient-boosted competing-risk baseline. All baselines are described in Appendix~\ref{sec:appendix_models}.

\subsection{Training Details}
All methods are evaluated with 5-fold cross-validation and fixed seed. We use batch size 128 for both single-risk and competing-risk experiments. Classification fine-tuning uses a learning rate of $10^{-5}$ and 5 fine-tuning epochs in the shared wrapper. Cox, DeepHit, and MTLR heads use learning rate $10^{-5}$ for up to 50 epochs in the benchmark configuration. Context-conditioned supervised forward passes use at most 512 labeled support examples; when fewer eligible examples are available, the full eligible support pool is used. During training, the support context is sampled from the training partition while excluding the current query minibatch. Observed events can be sparse, so supervised training uses event-balanced minibatches with a minimum quota of observed-event examples; the remaining positions are sampled from the full training pool and the resulting minibatch is shuffled before optimization. Hyperparameter search is applied to non-pretrained baselines with model-specific search spaces; each tuned baseline receives a 20-trial Optuna search within every training fold. No search is performed for any TabFM variant---zero-shot, classification fine-tuning, and survival-head adaptation all use the fixed settings above on every data set. Full implementation settings and baseline search spaces are summarized in Appendix~\ref{sec:finetuning_setup} and Table~\ref{tab:hyperparameters}.

\subsection{Evaluation Protocol}
We report the Antolini time-dependent concordance index $C_{td}$~\citep{antolini2005time-dependent} and Integrated Brier Score (IBS)~\citep{graf1999assessment} for discrimination and probabilistic prediction accuracy, respectively. IBS evaluates squared error of predicted survival probabilities over time under inverse-probability-of-censoring weighting; it is sensitive to both calibration and discrimination and should not be interpreted as a calibration-only metric. Mean dynamic area under the ROC curve (AUC), evaluated at the 25th, 50th, and 75th percentile time horizons, is reported only as a supplementary metric in the full-results appendix and is not used for the main comparative claims. IBS is integrated over 100 evenly spaced points spanning the observed follow-up, and the AUC horizons are percentiles of the observed event times in the test fold. 

\section{Results}
\label{sec:results}

\subsection{Adaptation and Backbone Effects}
\label{sec:objective_backbone_effects}

\begin{figure}[hbt]
\centering
\includegraphics[width=\linewidth]{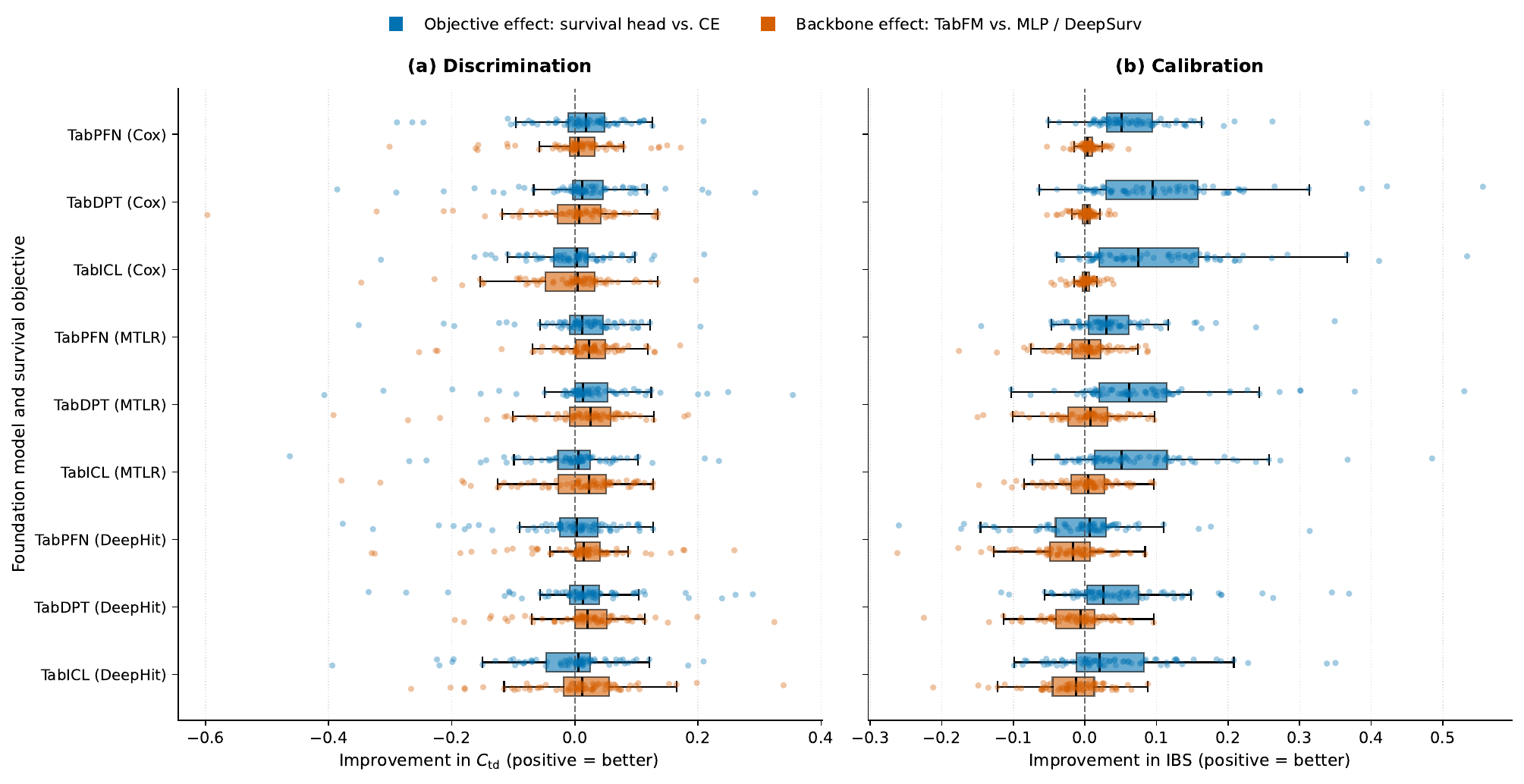}
\caption{\textbf{Adaptation and backbone effects across 74 single-risk data sets.} Blue compares survival-head adaptation with CE using the same frozen TabFM backbone; orange compares TabFM-based survival models with matched non-pretrained neural baselines. Panel (a) shows changes in $C_{\mathrm{td}}$ and panel (b) changes in IBS, sign-reversed so that positive values indicate improvement.}
\label{fig:effect_distribution_combined}
\end{figure}

We first separate two sources of performance variation: the effect of changing the adaptation interface while keeping the pretrained TabFM backbone fixed, and the effect of replacing a non-pretrained neural backbone with a pretrained TabFM under the same survival objective. Figure~\ref{fig:effect_distribution_combined} summarizes these as an \emph{adaptation effect}, comparing survival-head adaptation with CE under the same frozen TabFM backbone, and a \emph{backbone effect}, comparing a TabFM-based survival model with its matched non-pretrained neural counterpart.

For $C_{\mathrm{td}}$, both effects are generally centered near or above zero but vary substantially across data sets and model families. The separation is clearer for IBS: Cox shows the strongest and most consistent improvement over CE, MTLR also benefits with greater variability, while DeepHit gains are more pronounced for $C_{\mathrm{td}}$ than for IBS. These results indicate that the value of pretrained representations depends on the downstream survival head and evaluation metric, rather than yielding a universal advantage across objectives.

\subsection{Performance Across Data Regimes}
\label{sec:single_results}

\begin{figure}[hbt]
\centering
\includegraphics[width=0.7\linewidth]{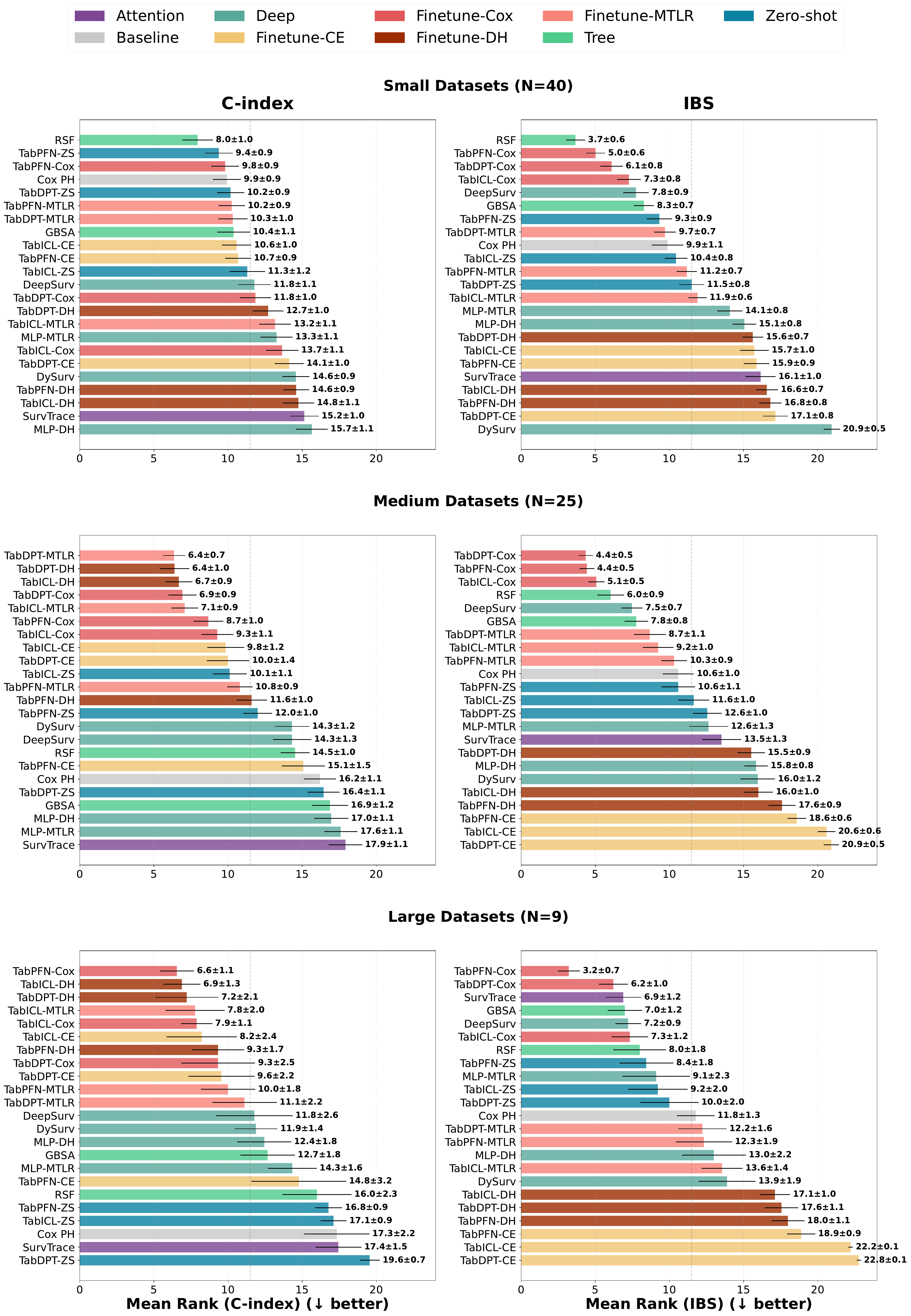}
\caption{Size-stratified mean-rank comparison for single-risk survival analysis. Lower rank is better. Results are aggregated across 40 small, 25 medium, and 9 large data sets and compare discrimination using time-dependent $C_{td}$ with probabilistic prediction using IBS.}
\label{fig:ranking_overall}
\end{figure}

We next examine how the relative performance of the full model interfaces changes with data-set size. Figure~\ref{fig:ranking_overall} shows a clear interaction between adaptation interface and data-set size; per-data-set results are reported in Table~\ref{tab:core_metrics}. On small data sets, RSF ranks best overall and zero-shot TabPFN remains competitive for $C_{\mathrm{td}}$. On medium data sets, supervised TabFM survival heads move toward the top of the rankings, with MTLR and DeepHit competitive for $C_{\mathrm{td}}$ and Cox-based adaptations among the strongest for IBS. On large data sets, Cox-adapted TabFMs lead for both $C_{\mathrm{td}}$ and IBS, while zero-shot and CE variants fall substantially behind. Supplementary pairwise win matrices and matched-backbone comparisons in Appendix~\ref{sec:pairwise_backbone} support this size-dependent pattern and show that the contribution of the pretrained backbone varies with the survival objective.

Across metrics, Cox is the most consistently strong interface for IBS and becomes especially competitive for $C_{td}$ on large data sets. DeepHit is relatively stronger for $C_{td}$ than for IBS, while MTLR remains competitive, particularly in the medium and large regimes. Across the data-size regimes, classification fine-tuning becomes more competitive relative to zero-shot inference as data sets grow, but remains weaker than the strongest survival heads for IBS.

\subsection{Transfer to Competing Risks}
\label{sec:competing_results}

We finally examine whether the single-risk ordering persists when multiple mutually exclusive event types must be modeled.

\begin{figure}[hbt]
\centering
\includegraphics[width=\linewidth]{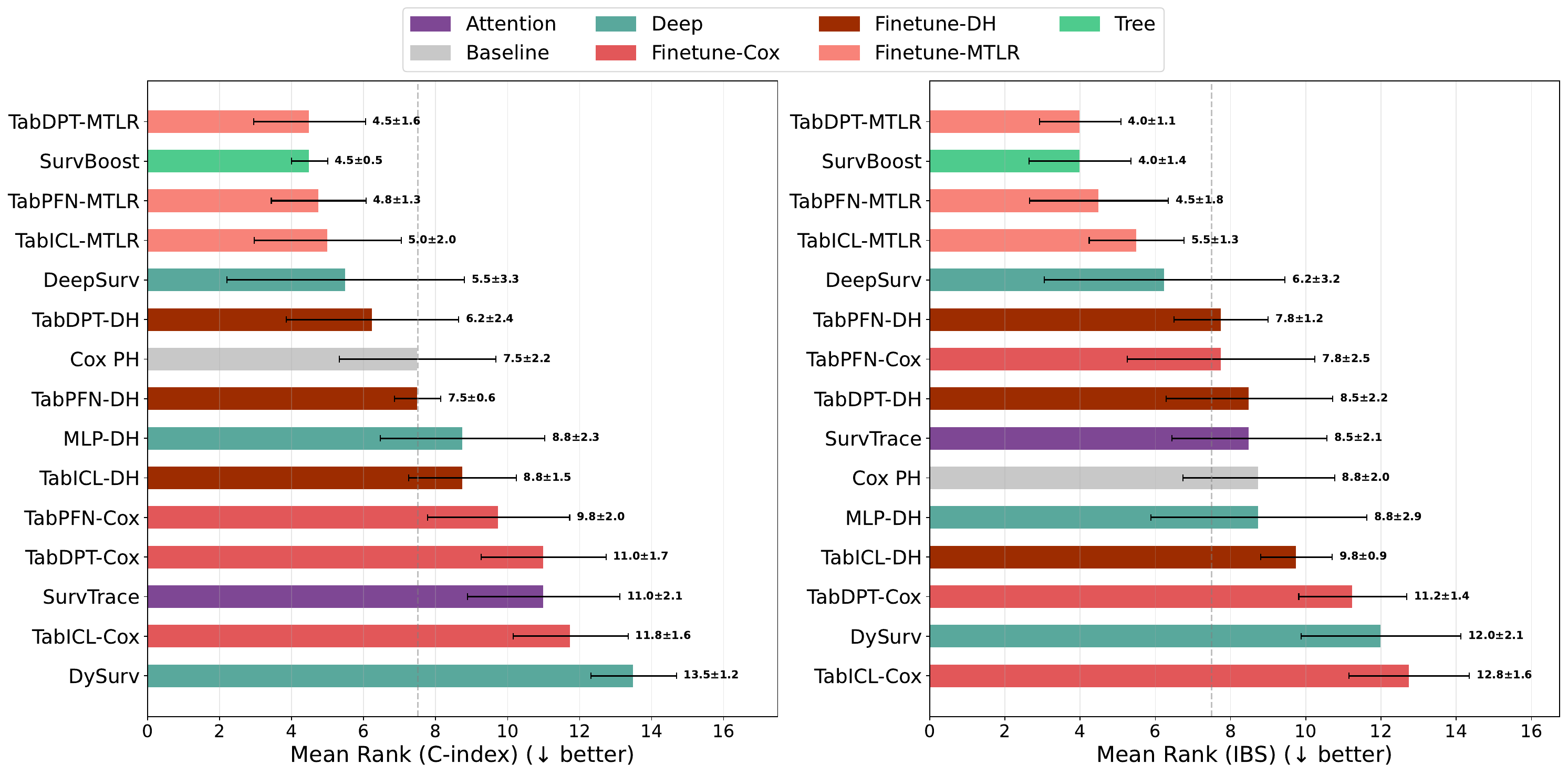}
\caption{\textbf{Mean-rank comparison for competing-risk survival} across four data sets. Lower rank is better. The comparison includes the three survival-head interfaces and classical, tree-based, and non-pretrained neural baselines; horizon-wise classification interfaces are excluded from this arm.}
\label{fig:ranking_competing_overall}
\end{figure}

Figure~\ref{fig:ranking_competing_overall} compares the three survival-head interfaces with competing-risk baselines, with per-cause results reported in Table~\ref{tab:detailed_metrics_cr}. Cause-specific MTLR is the strongest TabFM interface on both IBS and $C_{td}$, performing comparably to the strongest tree-based baseline, while joint DeepHit ranks in the middle of the field and cause-specific Cox is generally weaker. This differs from the single-risk ordering in Section~\ref{sec:single_results}, where Cox is the strongest interface overall. We interpret this ordering cautiously because only four competing-risk data sets were available for evaluation, making the mean ranks more sensitive to individual data sets and insufficient to establish a stable global ordering among the heads. The strong MTLR result also does not arise from joint probability normalization, since the cause-specific MTLR models are fitted independently; joint normalization is enforced only by the competing-risk DeepHit softmax over cause--time outcomes. The observed ranking therefore suggests an interaction between objective, discretization, backbone, and competing-risk formulation under a limited benchmark rather than a definitive ordering of the three interfaces.

\section{Discussion}
\label{sec:discussion}

Overall, the results support viewing survival transfer as an adaptation interface problem rather than simply a choice of downstream loss. Native zero-shot TabFM inference is most competitive on smaller data sets, whereas supervised adaptation becomes increasingly useful as data sets grow. This pattern is consistent with broader TabFM studies that report strong native in-context performance on smaller data sets and a reduced advantage as data sets become larger or more complex~\citep{purucker2026beyondiid}. Importantly, weaker zero-shot performance on larger data sets does not imply that the pretrained representation is no longer useful. Keeping the same frozen backbone while changing the downstream interface can substantially improve performance. Classification fine-tuning also becomes more competitive with zero-shot inference as data sets grow, but the stronger results from survival heads, particularly for IBS, suggest that learning a new head alone is not sufficient; the survival structure represented by that head also matters.

The survival heads show different performance profiles. Cox is the most consistently strong interface in the single-risk benchmark, particularly for IBS and for concordance on larger data sets. One possible explanation is that the pretrained representation already captures nonlinear structure, allowing the comparatively simple Cox head to provide an effective mapping to relative risk, although its proportional-hazards assumption remains restrictive. DeepHit is relatively stronger for $C_{td}$ than for IBS, which is consistent with the ranking component in its single-risk objective. This illustrates that strong risk ordering does not necessarily translate into equally accurate survival probabilities. MTLR remains competitive, particularly in the medium and large regimes, but the broader benchmark shows that its advantage is not consistent across backbones, metrics, and data regimes. Together, these results suggest that Cox, MTLR, DeepHit, and CE emphasize different aspects of survival prediction rather than following one fixed ordering.

The competing-risk experiments suggest that the single-risk ordering does not necessarily carry over when several mutually exclusive event types are modeled. Across the four evaluated data sets, cause-specific MTLR ranks highest among the TabFM survival heads on both metrics, while cause-specific Cox ranks lower. Because this analysis contains only four data sets, the ordering should be treated as preliminary and may be sensitive to individual data sets. The stronger MTLR result also cannot be explained simply by joint probability normalization, since the reported MTLR models are fitted independently for each cause, whereas joint normalization is enforced by the competing-risk DeepHit formulation. The results therefore suggest that the choice of survival objective interacts with the event structure and modeling formulation, rather than identifying one head as generally preferable.

Several limitations should be considered when interpreting these results. Mean ranks can hide important differences between individual data sets, and the competing-risk benchmark is much smaller than the single-risk benchmark. The competing-risk comparison is also limited to survival-head adaptation, so its conclusions apply only to the comparison among survival heads. In the single-risk analysis, CE and survival-head adaptation differ in both target construction and objective, so their comparison should not be interpreted as a pure loss-function ablation. Similarly, the data-size analysis compares different data sets and therefore does not isolate sample size from dimensionality, censoring, event prevalence, or domain. Explanations for the behavior of individual heads should therefore be treated as hypotheses consistent with the observed results rather than causal conclusions. Finally, the competing-risk comparison should not be interpreted as isolating the effect of the survival head alone. Cox and MTLR use cause-specific formulations, whereas DeepHit uses a jointly normalized cause--time distribution. The observed ordering may therefore reflect both the head family and the way competing risks are represented.

Future work could use more controlled experiments to separate the effects of target construction, censoring treatment, ranking supervision, and sample size. Larger competing-risk benchmarks would also be needed to determine whether the observed ordering among survival heads is stable. It would also be useful to combine context adaptation with survival-specific heads and to compare adaptation of generic TabFMs directly with survival-aware pretraining.

\section{Conclusion}

We study how generic tabular foundation models can be transferred to censored time-to-event prediction through different adaptation interfaces. Across 74 single-risk data sets, native zero-shot inference remains competitive in the small-data regime, while supervised adaptation becomes increasingly useful on medium and large data sets. This pattern cannot be explained simply by a loss of value in the pretrained representation: with task-specific heads, frozen TabFM backbones support several of the strongest methods in the larger data regimes. Relative to the MTLR adaptation established in our earlier clinical study~\citep{pham2026tabular}, the broader benchmark shows that Cox is the most consistently strong interface for single-risk prediction, particularly for IBS and for concordance on larger data sets, while DeepHit is relatively stronger for discrimination than for probabilistic prediction. In the competing-risk analysis, based on four data sets, cause-specific MTLR performs best among the TabFM survival heads, although this ordering should be interpreted cautiously given the limited benchmark. Overall, the results indicate that successful TabFM transfer depends on the amount of available data, the evaluation target, and how directly the downstream interface represents survival structure.

\acks{This research was conducted with the financial support of Taighde Éireann-Research Ireland under Grant Agreement No. 13/RC/2106\_P2 at ADAPT, the Research Ireland Centre for AI-Driven Digital Content Technology at DCU funded through the Research Ireland Research Centres Programme. For the purpose of Open Access, the author has applied a CC BY public copyright licence to any author-accepted manuscript version arising from this submission. We declare no competing interests.}

\clearpage
\appendix

\section{Data Set Summary}
\begin{longtable}{rp{2.5cm}rrrlll}
\caption{\textbf{Data set characteristics.} $N$: number of subjects. Event rate: proportion of uncensored observations. Bins: number of adaptive time intervals used by the single-risk discrete-time formulations. SurvSet data sets are collected from~\cite{drysdale2022survset}.} \label{tab:datasets_survset} \\
\toprule
\textbf{ID} & \textbf{Data set} & \textbf{$N$} & \textbf{Feat.} & \textbf{Event rate} & \textbf{Bins} & \textbf{Domain} & \textbf{Regime} \\
\midrule
\endfirsthead
\multicolumn{8}{c}{Table continued} \\
\toprule
\textbf{ID} & \textbf{Data set} & \textbf{$N$} & \textbf{Feat.} & \textbf{Event rate} & \textbf{Bins} & \textbf{Domain} & \textbf{Regime} \\
\midrule
\endhead
\midrule
\multicolumn{8}{r}{\textit{Continued on next page}} \\
\endfoot
\bottomrule
\endlastfoot
1 & ACATH & 2,230 & 3 & 0.66 & 10 & Cardiology & medium \\
2 & ACTG & 1,151 & 17 & 0.08 & 10 & Infectious disease & medium \\
3 & AIDS & 938 & 5 & 0.15 & 10 & Infectious disease & medium \\
4 & AIDS2 & 2,810 & 12 & 0.62 & 50 & Infectious disease & medium \\
5 & AML BULL & 115 & 6283 & 0.57 & 10 & Haematology & small \\
6 & BERGAMASCHI & 82 & 10 & 0.34 & 10 & Oncology & small \\
7 & BREAST & 100 & 4 & 0.26 & 10 & Oncology & small \\
8 & BURN & 154 & 13 & 0.31 & 10 & Critical care & small \\
9 & CANCER & 169 & 28 & 0.72 & 20 & Oncology & small \\
10 & CGD & 128 & 23 & 0.34 & 10 & Immunology & small \\
11 & CHOP & 412 & 3833 & 0.40 & 20 & Haematology & small \\
12 & COLON & 911 & 12 & 0.48 & 50 & Oncology & medium \\
13 & COST & 518 & 13 & 0.78 & 50 & Oncology & medium \\
14 & CSL & 2,035 & 6 & 0.12 & 30 & Hepatology & medium \\
15 & D.OROPHA.REC & 192 & 24 & 0.72 & 20 & Oncology & small \\
16 & DATADIVAT1 & 5,943 & 16 & 0.16 & 50 & Nephrology & large \\
17 & DATADIVAT2 & 1,837 & 4 & 0.32 & 50 & Nephrology & medium \\
18 & DATADIVAT3 & 4,267 & 16 & 0.06 & 30 & Nephrology & large \\
19 & DATAOVARIAN1 & 910 & 162 & 0.60 & 20 & Oncology & medium \\
20 & DBCD & 295 & 4919 & 0.27 & 10 & Oncology & small \\
21 & DIABETES & 394 & 4 & 0.39 & 30 & Endocrinology & small \\
22 & DIALYSIS & 6,805 & 72 & 0.24 & 10 & Nephrology & large \\
23 & DIVORCE & 3,371 & 4 & 0.31 & 100 & Social/Economics & medium \\
24 & DLBCL & 240 & 7399 & 0.57 & 20 & Haematology & small \\
25 & E1684 & 284 & 3 & 0.69 & 30 & Oncology & small \\
26 & EPILEPTIC & 681 & 5 & 0.16 & 20 & Neurology & medium \\
27 & FLCHAIN & 6,521 & 32 & 0.30 & 100 & Haematology & large \\
28 & FRAMINGHAM & 4,658 & 17 & 0.31 & 100 & Cardiology & large \\
29 & FRTCS & 1,391 & 5 & 0.05 & 10 & Cardiology & medium \\
30 & GBSG2 & 686 & 9 & 0.44 & 50 & Oncology & medium \\
31 & GLIOMA & 37 & 4 & 0.62 & 10 & Oncology & small \\
32 & GRACE & 1,000 & 5 & 0.32 & 10 & Cardiology & medium \\
33 & GSE1992 & 123 & 15541 & 0.28 & 10 & Oncology & small \\
34 & GSE3143 & 158 & 8663 & 0.32 & 10 & Oncology & small \\
35 & GSE4335 & 113 & 12815 & 0.34 & 10 & Oncology & small \\
36 & HDFAIL & 52,422 & 36 & 0.06 & 100 & Social/Economics & large \\
37 & HEART & 69 & 4 & 0.65 & 10 & Cardiology & small \\
38 & HEARTVALVE & 444 & 23 & 0.05 & 10 & Cardiology & small \\
39 & HEPATOCELLULAR & 101 & 43 & 0.45 & 10 & Oncology & small \\
40 & LEUKSURV & 1,043 & 29 & 0.84 & 30 & Haematology & medium \\
41 & MCLCLEANED & 92 & 574 & 0.70 & 10 & Haematology & small \\
42 & MELANOMA & 205 & 5 & 0.28 & 10 & Oncology & small \\
43 & MICRO.CENSURE & 117 & 81 & 0.22 & 10 & Oncology & small \\
44 & NKI70 & 144 & 76 & 0.33 & 10 & Oncology & small \\
45 & NSBCD & 115 & 549 & 0.33 & 10 & Oncology & small \\
46 & NWTCO & 4,028 & 9 & 0.14 & 50 & Oncology & large \\
47 & OLDMORT & 6,495 & 13 & 0.30 & 10 & All-cause mortality & large \\
48 & OVA & 358 & 11 & 0.74 & 50 & Oncology & small \\
49 & OVARIAN & 26 & 4 & 0.46 & 10 & Oncology & small \\
50 & PBC & 312 & 6 & 0.40 & 20 & Hepatology & small \\
51 & PBC3 & 327 & 19 & 0.17 & 10 & Hepatology & small \\
52 & PHARMACOSMOKING & 113 & 16 & 0.68 & 10 & Medicine & small \\
53 & PHPL04K8A & 442 & 21 & 0.53 & 30 & Oncology & small \\
54 & PROSTATE & 467 & 25 & 0.70 & 20 & Urology & small \\
55 & PROSTATESURVIVAL & 14,065 & 6 & 0.06 & 50 & Urology & large \\
56 & RDATA & 1,040 & 6 & 0.53 & 50 & Oncology & medium \\
57 & RETINOPATHY & 394 & 11 & 0.39 & 20 & Ophthalmology & small \\
58 & RHC & 634 & 73 & 0.42 & 30 & Critical care & medium \\
59 & ROSSI & 973 & 27 & 0.07 & 10 & Social/Economics & medium \\
60 & ROTT2 & 2,982 & 12 & 0.43 & 100 & Oncology & medium \\
61 & SCANIA & 1,931 & 8 & 0.56 & 100 & All-cause mortality & medium \\
62 & SMARTO & 1,021 & 34 & 0.11 & 10 & Cardiology & medium \\
63 & STAGEC & 136 & 18 & 0.38 & 10 & Urology & small \\
64 & SUPPORT2 & 530 & 65 & 0.53 & 30 & Critical care & medium \\
65 & TRACE & 1,878 & 6 & 0.51 & 100 & Cardiology & medium \\
66 & UIS & 579 & 14 & 0.81 & 50 & Psychiatry & medium \\
67 & UNEMPDUR & 3,241 & 6 & 0.61 & 10 & Social/Economics & medium \\
68 & UNEMPLOYMENT & 450 & 7 & 0.56 & 10 & Social/Economics & small \\
69 & VDV & 78 & 4705 & 0.44 & 10 & Oncology & small \\
70 & VETERAN & 137 & 8 & 0.93 & 20 & Oncology & small \\
71 & VLBW & 257 & 41 & 0.14 & 10 & Neonatology & small \\
72 & WPBC & 194 & 32 & 0.24 & 10 & Oncology & small \\
73 & Z243 & 100 & 23 & 1.00 & 10 & Medicine & small \\
74 & ZINC & 431 & 20 & 0.19 & 10 & Oncology & small \\
\midrule
 & \textit{Min} & 26 & 3 & 0.05 & 10 &  &  \\
 & \textit{Max} & 52,422 & 15541 & 1.00 & 100 &  &  \\
 & \textit{Mean} & 2,027 & 901 & 0.41 & 28 &  &  \\
 & \textit{IQR\textsubscript{25--75}} & [146, 1,331] & [6, 36] & [0.24, 0.57] & [10, 30] &  &  \\
\end{longtable}

\begin{longtable}{rp{2.5cm}rrrlll}
\caption{\textbf{Data set characteristics for data sets with competing risks.} $N$: number of subjects. Event rate: proportion of uncensored observations. Bins: number of discrete time intervals used by discrete-time models; interval construction depends on the formulation, with cause-specific MTLR using the corresponding adaptive single-risk discretization and joint DeepHit using equal-width intervals over the observed event-time range.} \label{tab:datasets_cr} \\
\toprule
\textbf{ID} & \textbf{Data set} & \textbf{$N$} & \textbf{Feat.} & \textbf{Event rate} & \textbf{Bins} & \textbf{Domain} & \textbf{Regime} \\
\midrule
\endfirsthead
\multicolumn{8}{c}{Table continued} \\
\toprule
\textbf{ID} & \textbf{Data set} & \textbf{$N$} & \textbf{Feat.} & \textbf{Event rate} & \textbf{Bins} & \textbf{Domain} & \textbf{Regime} \\
\midrule
\endhead
\midrule
\multicolumn{8}{r}{\textit{Continued on next page}} \\
\endfoot
\bottomrule
\endlastfoot
\multicolumn{8}{c}{\textit{Data Sets with Competing Risks}} \\
\midrule
1 & FRAMINGHAM & 4,273 & 18 & 0.42 ($K$=2) & 100 & Cardiology & large  \\
2 & PBC2 & 1,945 & 16 & 0.45 ($K$=2) & 50 & Hepatology & medium\\
3 & SUPPORT2-CR  & 9,105 & 37 & 0.68 ($K$=2) & 30 & Critical care & large  \\
4 & SYNTHETIC & 30,000 & 12 & 0.50 ($K$=2) & 10 & Synthetic & large\\
\midrule
 & \textit{Min} & 1,945 & 12 & 0.42 & 10 &  &  \\
 & \textit{Max} & 30,000 & 37 & 0.68 & 100 &  &  \\
 & \textit{Mean} & 11,331 & 21 & 0.51 & 48 &  &  \\
 & \textit{IQR\textsubscript{25--75}} & [3,691, 14,329] & [15, 23] & [0.44, 0.55] & [25, 62] &  &  \\
\end{longtable}

\section{Model Details}
\label{sec:appendix_models}

This section summarizes the non-TabFM models evaluated in the main experiments. Unless otherwise stated, classical and deep baselines are implemented using \texttt{scikit-survival} or \texttt{pycox}, while DySurv and SurvTRACE are adapted from their original codebases. The three TabFM backbones and exact versions used in the benchmark are described in Section~\ref{subsec:tabfm_related}.

\noindent\textbf{Cox Proportional Hazards (Cox PH)} assumes $h(t\mid\mathbf{x})=h_0(t)\exp(\beta^\top\mathbf{x})$, where $h_0(t)$ is an unspecified baseline hazard and $\beta^\top\mathbf{x}$ defines relative risk. Parameters are estimated by maximizing the regularized partial likelihood; the benchmark implementation uses fixed regularization rather than an Optuna search.

\noindent\textbf{Random Survival Forests (RSF)} extends random forests to censored outcomes using log-rank splitting criteria and ensemble averaging. Each tree estimates cumulative hazard via the Nelson--Aalen estimator, and predictions are aggregated across trees.

\noindent\textbf{Gradient Boosting Survival Analysis (GBSA)}
GBSA applies gradient boosting to survival objectives such as Cox partial likelihood, sequentially adding weak learners through functional gradient descent.

\noindent\textbf{SurvBoost} is the competing-risk gradient-boosting baseline. For each cause it fits a gradient-boosted survival model with the remaining observed causes treated as censoring, and evaluates the resulting cause-specific survival functions on a shared time grid. It is the strongest non-pretrained competitor in the competing-risk benchmark and is reported only there.

\noindent\textbf{DeepSurv} generalizes Cox PH by replacing the linear predictor with a neural network, modeling $h(t\mid\mathbf{x})=h_0(t)\exp(f_\theta(\mathbf{x}))$. Training minimizes the negative Cox partial log-likelihood.

\noindent\textbf{DeepHit} models a discrete event-time distribution. In the single-risk experiments we use likelihood plus ranking supervision; in the competing-risk TabFM implementation, the reported joint DeepHit objective is the normalized cause--time likelihood described in Section~\ref{sec:competing_methods}.

\noindent\textbf{MLP-MTLR} and \textbf{MLP-DH} are non-pretrained counterparts to the TabFM survival-head variants. Both use the same multilayer-perceptron trunk as DeepSurv over the raw standardized covariates, replacing the Cox head with the MTLR and DeepHit objectives respectively and using the same adaptive event-balanced discretization as the corresponding TabFM heads. They isolate the contribution of the pretrained representation: comparing TabPFN-MTLR with MLP-MTLR, or TabPFN-DH with MLP-DH, holds the survival objective and the time grid fixed and varies only whether the representation comes from a pretrained tabular prior or a trunk trained from scratch. DeepSurv plays the same role for the Cox head.

\noindent\textbf{SurvTRACE} applies self-attention to tabular features and is pretrained using masked feature modeling before survival fine-tuning.

\noindent\textbf{DySurv} combines neural survival modeling with latent-variable inference to capture nonlinear and heterogeneous risk dynamics.

\clearpage

\section{Training and Hyperparameter Details}
\label{sec:finetuning_setup}

No hyperparameter search is performed for the TabFM variants; all three interfaces use fixed settings across data sets. Table~\ref{tab:hyperparameters} summarizes these settings. The non-pretrained baselines with model-specific search spaces are tuned independently within each training fold using 20 Optuna trials, with the search spaces reported in Table~\ref{tab:baseline_search_spaces}.

\begin{table}[hb]
\centering
\small
\caption{Strategy-level settings used for the TabFM interfaces. A dash denotes a setting that is not applicable.}
\label{tab:hyperparameters}

\resizebox{\linewidth}{!}{
\begin{tabular}{lccc}
\toprule
\textbf{Setting} &
\textbf{Zero-shot} &
\textbf{Classification FT} &
\textbf{Survival-head} \\
\midrule

\multicolumn{4}{l}{\emph{Backbone}} \\
Gradient updates
    & None
    & None (frozen)
    & None (frozen) \\
Trainable head
    & Native classifier
    & Classification
    & Survival \\
Head architecture
    & --
    & \multicolumn{2}{c}{1 hidden layer, 64 units, dropout 0.1} \\

\midrule
\multicolumn{4}{l}{\emph{Optimization}} \\
Optimizer
    & --
    & \multicolumn{2}{c}{AdamW; weight decay $10^{-3}$} \\
Learning rate
    & --
    & $10^{-5}$
    & $10^{-5}$ \\
Schedule
    & --
    & \multicolumn{2}{c}{Cosine annealing to $0.01\times$ initial LR} \\
Gradient clipping
    & --
    & \multicolumn{2}{c}{Max norm 1.0} \\
Epochs
    & --
    & 5
    & $\leq 50$; patience 5 \\
Batch size
    & --
    & 128
    & 128 \\
Min. events/batch
    & --
    & 16
    & 16 \\

\midrule
\multicolumn{4}{l}{\emph{Context}} \\
Max. support size
    & 512
    & 512
    & 512 \\
Context resampling
    & Per horizon
    & Per step
    & Per step \\
Inference contexts ($R$)
    & 5
    & 5
    & 5 \\

\midrule
\multicolumn{4}{l}{\emph{Protocol}} \\
CV folds
    & 5
    & 5
    & 5 \\
Hyperparameter search
    & None
    & None
    & None \\

\bottomrule
\end{tabular}
}
\end{table}

\subsection{Hyperparameter Search for Non-Pretrained Baselines}

The non-pretrained baselines with model-specific search configurations are tuned independently within each training fold using 20 Optuna trials. Table~\ref{tab:baseline_search_spaces} reports the exact search spaces defined in the experimental configuration files. Continuous parameters marked as log-uniform are sampled on a logarithmic scale; integer ranges are sampled with unit step unless otherwise noted; and categorical parameters are sampled from the listed values. 

\begin{table}[hbt]
\centering
\small
\caption{Hyperparameter search spaces for the tuned non-pretrained baselines. Each model is optimized using 20 Optuna trials within each training fold. }
\label{tab:baseline_search_spaces}
\begin{tabular}{lp{0.36\linewidth}p{0.38\linewidth}}
\toprule
\textbf{Model} & \textbf{Hyperparameter} & \textbf{Search space} \\
\midrule
\multirow{3}{*}{RSF}
& Number of trees & $[50,200]$ (integer) \\
& Minimum samples to split & $[5,20]$ (integer) \\
& Minimum samples per leaf & $[5,20]$ (integer) \\
\midrule
\multirow{3}{*}{GBSA}
& Learning rate & $[10^{-2},2\times10^{-1}]$ (log-uniform) \\
& Number of estimators & $[50,200]$ (integer) \\
& Maximum depth & $[2,5]$ (integer) \\
\midrule
\multirow{5}{*}{DeepSurv}
& Learning rate & $[10^{-4},5\times10^{-2}]$ (log-uniform) \\
& Dropout & $[0,0.5]$ (uniform) \\
& Number of hidden layers & $\{1,2,3\}$ \\
& Hidden width & $\{16,32,64\}$ \\
& Epochs & $\{10,20,\ldots,100\}$ \\
\midrule
\multirow{4}{*}{MLP-MTLR}
& Learning rate & $[10^{-4},10^{-1}]$ (log-uniform) \\
& Dropout & $[0,0.5]$ (uniform) \\
& Number of hidden layers & $[1,3]$ (integer) \\
& Hidden width & $\{32,64,128\}$ \\
\midrule
\multirow{6}{*}{MLP-DH}
& Learning rate & $[10^{-4},10^{-1}]$ (log-uniform) \\
& Dropout & $[0,0.5]$ (uniform) \\
& Number of hidden layers & $[1,3]$ (integer) \\
& Hidden width & $\{32,64,128\}$ \\
& DeepHit $\alpha$ & $[0,0.5]$ (uniform) \\
& DeepHit $\sigma$ & $[10^{-2},1]$ (log-uniform) \\
\midrule
\multirow{5}{*}{SurvTRACE}
& Learning rate & $[10^{-4},10^{-2}]$ (log-uniform) \\
& Hidden size & $\{8,16,32\}$ \\
& Hidden layers & $\{1,2,3,4\}$ \\
& Attention heads & $\{2,4,8\}$ \\
& Intermediate size & $\{32,64,128\}$ \\
\midrule
\multirow{3}{*}{DySurv}
& Learning rate & $[10^{-4},10^{-2}]$ (log-uniform) \\
& Encoded features & $\{16,32,64,128\}$ \\
& Batch size & $\{64,128,256\}$ \\
\bottomrule
\end{tabular}
\end{table}

\section{Supplementary Performance Comparisons}
\label{sec:pairwise_backbone}

The size-stratified mean-rank analysis in Section~\ref{sec:single_results} summarizes the relative ordering of models across data regimes, but it does not show which specific model-to-model comparisons drive those ranks. To provide a comprehensive picture of how adaptation interfaces and pretrained representations interact, this section presents: (i) full pairwise win matrices across all evaluated models; (ii) backbone gains over matched non-pretrained neural survival baselines; (iii) survival-head gains over same-backbone zero-shot reformulation; and (iv) survival-head gains over same-backbone classification fine-tuning.

\paragraph{Statistical Testing Methodology.}
All comparisons in this section are conducted at the data-set level after averaging performance metrics across the 5 repeated cross-validation folds for each data set. For any pair of models $A$ and $B$, we compute the paired difference $\Delta C_{\mathrm{td}} = C_{\mathrm{td}}(A) - C_{\mathrm{td}}(B)$. The data-set-level win score is defined as $P(\Delta C_{\mathrm{td}} > 0) + 0.5 P(|\Delta C_{\mathrm{td}}| \le \epsilon)$, where $\epsilon = 0.001$ denotes the absolute threshold below which differences are treated as ties. To evaluate statistical significance, we conduct one-sided paired Wilcoxon signed-rank tests assessing whether $\Delta C_{\mathrm{td}} > 0$. Multiplicity adjustment across models within each data-size regime is performed using the Benjamini--Hochberg False Discovery Rate (FDR) procedure. Statistically significant gains are denoted by an asterisk (\textbf{*}), indicating an FDR-adjusted significance threshold of $q \le 0.05$. To further test whether the choice of TabFM backbone (TabPFN, TabDPT, TabICL) significantly affects downstream performance within a given survival head, we perform non-parametric omnibus Friedman tests across backbones, with Benjamini--Hochberg correction across heads and data regimes.

\begin{figure}[p]
\centering
\includegraphics[width=\linewidth]{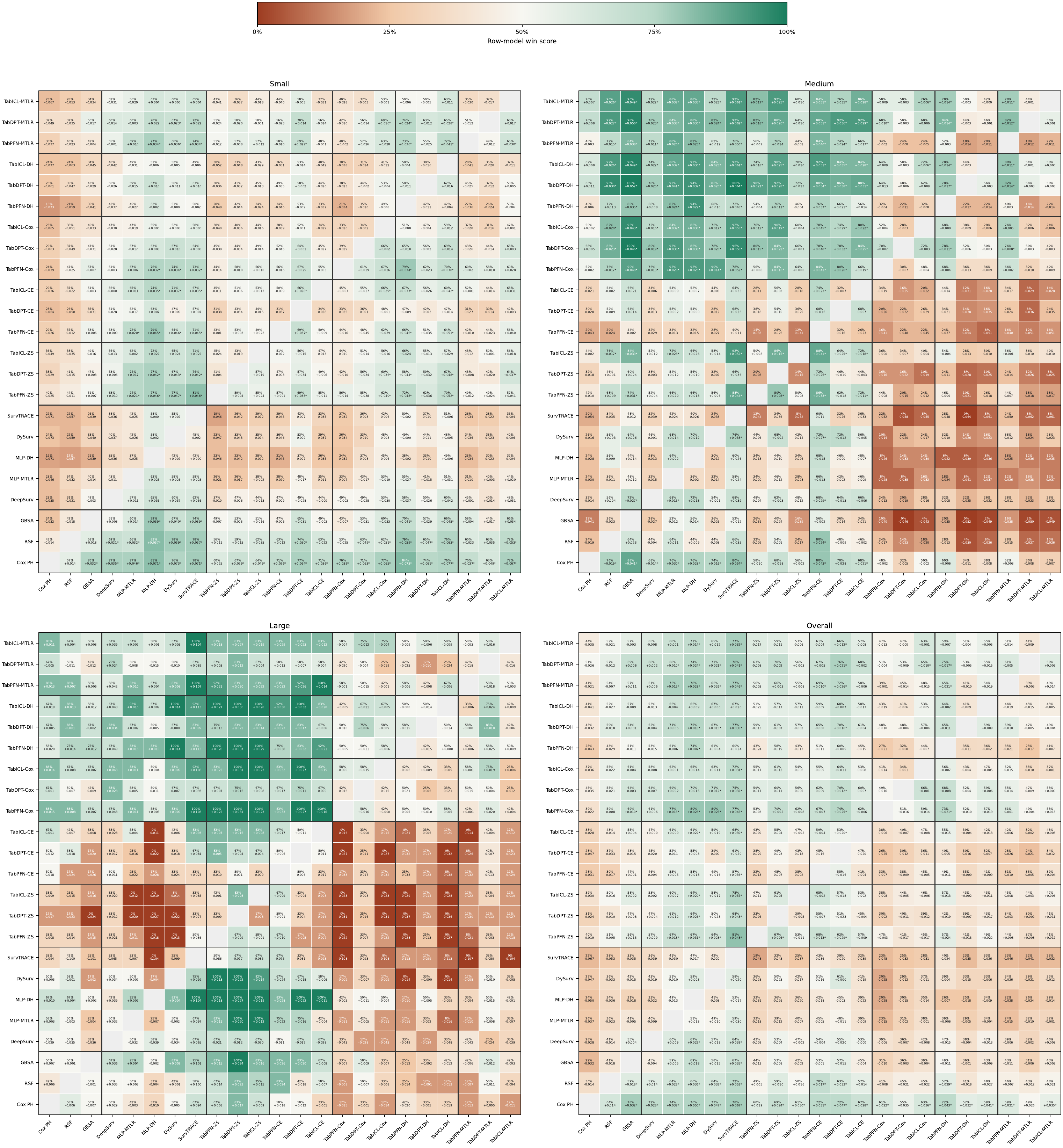}
\caption{\textbf{Pairwise model comparisons for $C_{\mathrm{td}}$ across data-size regimes.} Cells report data-set-level win scores and mean signed differences. $^*$ denotes Benjamini--Hochberg-corrected paired Wilcoxon tests with $q\le0.05$.}
\label{fig:pairwise_win_matrix_ctd}
\end{figure}

\paragraph{Pairwise Model Win Matrices.}
Figure~\ref{fig:pairwise_win_matrix_ctd} reports the complete pairwise win matrix for $C_{\mathrm{td}}$ across Small ($\le 500$ samples), Medium ($500 < N \le 4000$), Large ($> 4000$), and Overall regimes. The pairwise view confirms that no single adaptation interface dominates uniformly across all scales. On small data sets, zero-shot inference and classical baselines remain competitive against supervised neural adaptations. As data sets scale to medium and large regimes, supervised survival-head variants systematically gain advantage, with statistically significant win scores ($q \le 0.05$) over zero-shot and classification fine-tuning interfaces.

\begin{figure}[p]
\centering
\begin{subfigure}{\linewidth}
    \centering
    \includegraphics[width=0.88\linewidth]{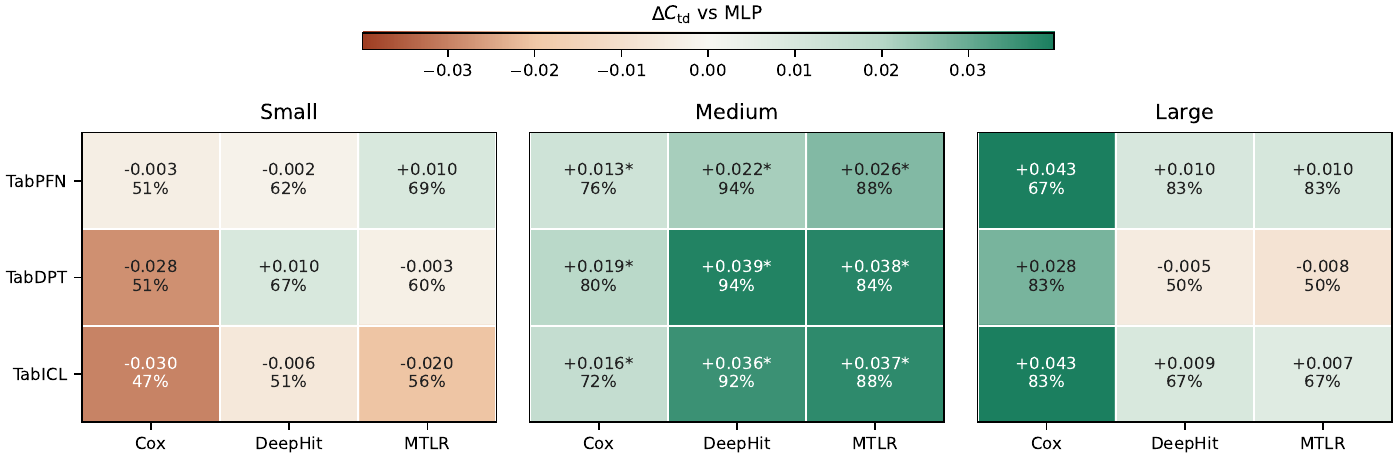}
    \caption{Backbone gains over matched MLP survival heads.}
    \label{fig:backbone_lift_ctd}
\end{subfigure}
\vspace{0.6em}

\begin{subfigure}{\linewidth}
    \centering
    \includegraphics[width=0.88\linewidth]{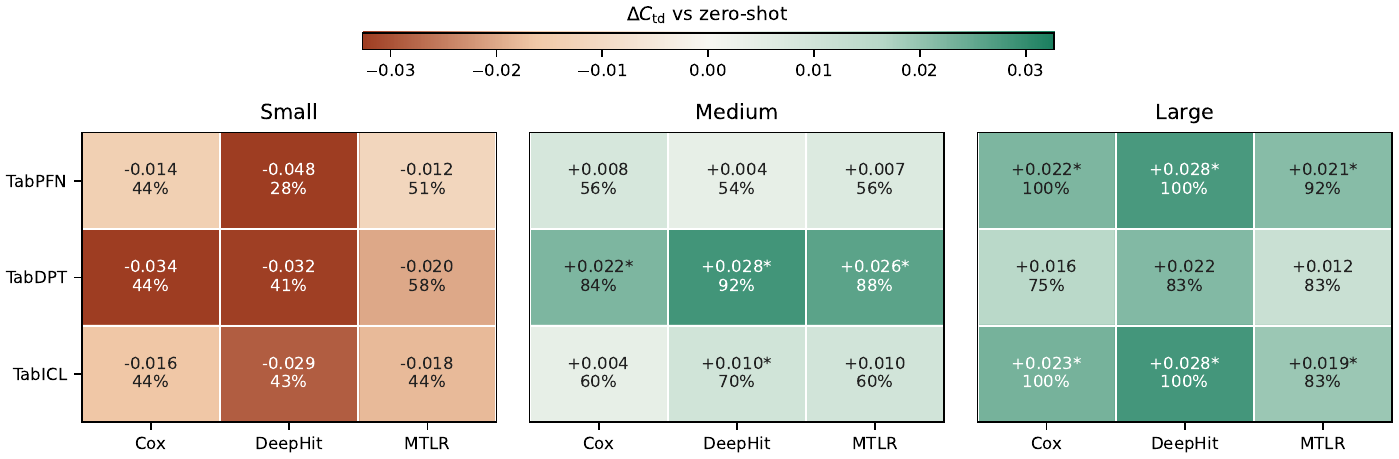}
    \caption{Survival-head gains over same-backbone zero-shot reformulation.}
    \label{fig:backbone_vs_zeroshot_ctd}
\end{subfigure}
\vspace{0.6em}

\begin{subfigure}{\linewidth}
    \centering
    \includegraphics[width=0.88\linewidth]{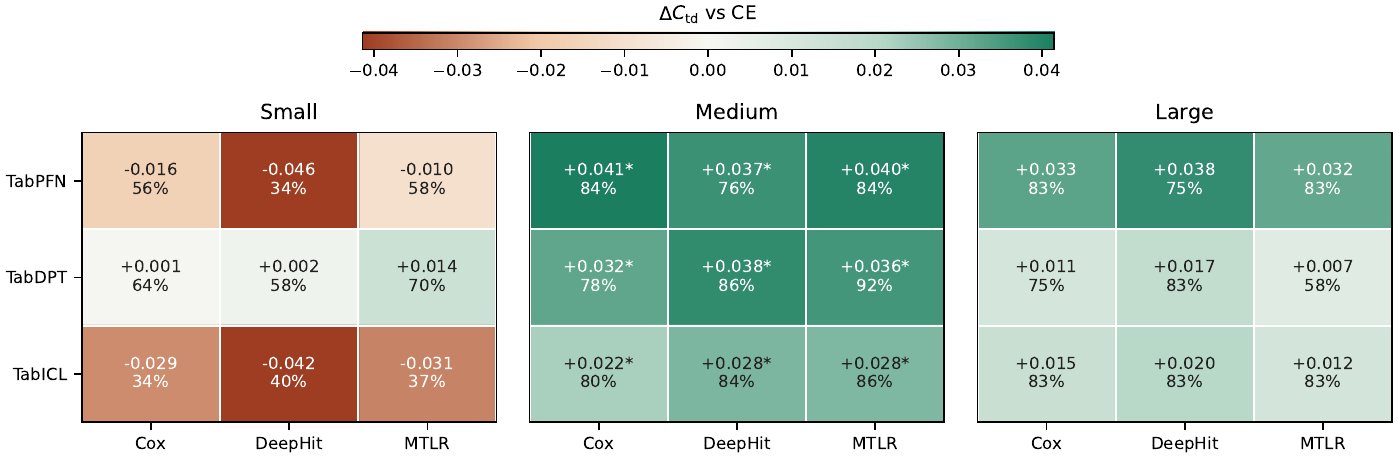}
    \caption{Survival-head gains over same-backbone classification fine-tuning.}
    \label{fig:backbone_vs_finetune_ctd}
\end{subfigure}

\caption{\textbf{Size-stratified $C_{\mathrm{td}}$ performance gains across TabFM adaptation interfaces.} (a) Backbone gains over matched non-pretrained survival models; (b) survival-head gains over same-backbone zero-shot reformulation; (c) survival-head gains over same-backbone classification fine-tuning. $^*$ indicates Benjamini--Hochberg-corrected one-sided paired Wilcoxon tests with $q\le0.05$.}
\label{fig:backbone_lift_all}
\end{figure}

\paragraph{Backbone gains over Matched MLP Baselines.}
Figure~\ref{fig:backbone_lift_all}(a) isolates the pure representation advantage of pretrained TabFMs by matching each TabFM survival head to a non-pretrained multilayer perceptron (MLP) sharing the exact same downstream survival objective: TabFM-Cox against DeepSurv, TabFM-DH against MLP-DeepHit, and TabFM-MTLR against MLP-MTLR. Positive $\Delta C_{\mathrm{td}}$ values indicate that substituting the learned MLP trunk with a frozen pretrained TabFM representation improves discrimination. In the small data regime, pretrained backbones show modest or mixed gains over MLP baselines. In medium and large regimes, however, pretrained representations provide consistent, statistically significant improvements (e.g., TabDPT-DH achieves $+0.039$ $\Delta C_{\mathrm{td}}$ with a $94\%$ win score, $q \le 0.05$). Friedman tests confirm that backbone representation choice significantly affects DeepHit adaptation ($q=0.030$), whereas tests for Cox ($q=0.053$) and MTLR ($q=0.149$) fall just outside the $q \le 0.05$ threshold.

\paragraph{Survival-Head Lift over Same-Backbone Zero-Shot Reformulation.}
Figure~\ref{fig:backbone_lift_all}(b) compares supervised survival-head adaptation directly against native zero-shot horizon-wise classification on the exact same pretrained backbone (TabPFN-ZS, TabDPT-ZS, TabICL-ZS). This directly answers whether adding an explicit survival head is superior to using the foundation model out-of-the-box. On small data sets, zero-shot inference is highly competitive, outperforming supervised heads across several combinations (win scores $28\%$--$58\%$, no significant survival-head advantage), highlighting that native zero-shot in-context inference is strong when supervision is limited. In medium and large regimes, however, explicit survival heads achieve substantial, statistically significant gains over zero-shot inference (e.g., TabDPT-DH achieves $+0.028$ $\Delta C_{\mathrm{td}}$ with a $92\%$ win score in medium data sets; in large data sets, win scores reach $75\%$--$100\%$ with extensive $q \le 0.05$ significance across all three backbones).

\paragraph{Survival-Head Lift over Same-Backbone Classification Fine-Tuning.}
Figure~\ref{fig:backbone_lift_all}(c) compares survival-head adaptation against horizon-expanded cross-entropy classification fine-tuning on the same backbone (TabPFN-CE, TabDPT-CE, TabICL-CE). While CE fine-tuning optimizes a new classification head on temporally expanded binary targets, it does not explicitly account for continuous hazard rates or joint discrete-time event distributions. In the small data regime, differences between CE fine-tuning and survival heads are mixed. In medium and large regimes, however, dedicated survival-head adaptations decisively outperform CE fine-tuning across all backbones and heads (e.g., across medium data sets, mean $\Delta C_{\mathrm{td}}$ gains range from $+0.022$ to $+0.041$ with $76\%$--$92\%$ win scores, and \emph{every} combination achieves $q \le 0.05$; in large data sets, win scores range from $75\%$ to $83\%$). These results demonstrate that transferring TabFMs to time-to-event prediction is not merely a matter of supervised fine-tuning; incorporating survival-specific adaptation heads is essential for maximizing predictive performance.

\section{Ablation Studies}

The following supplementary analyses probe complementary aspects of the main benchmark rather than introducing separate performance claims. Together they ask when additional supervision becomes useful, how sensitive the discrete-time interfaces are to temporal resolution, and how the three adaptation regimes differ in computational and qualitative behavior.

\subsection{Label Efficiency Under Increasing Supervision}
\label{sec:number_samples}
\begin{figure}[hbt]
\centering
\includegraphics[width=0.6\linewidth]{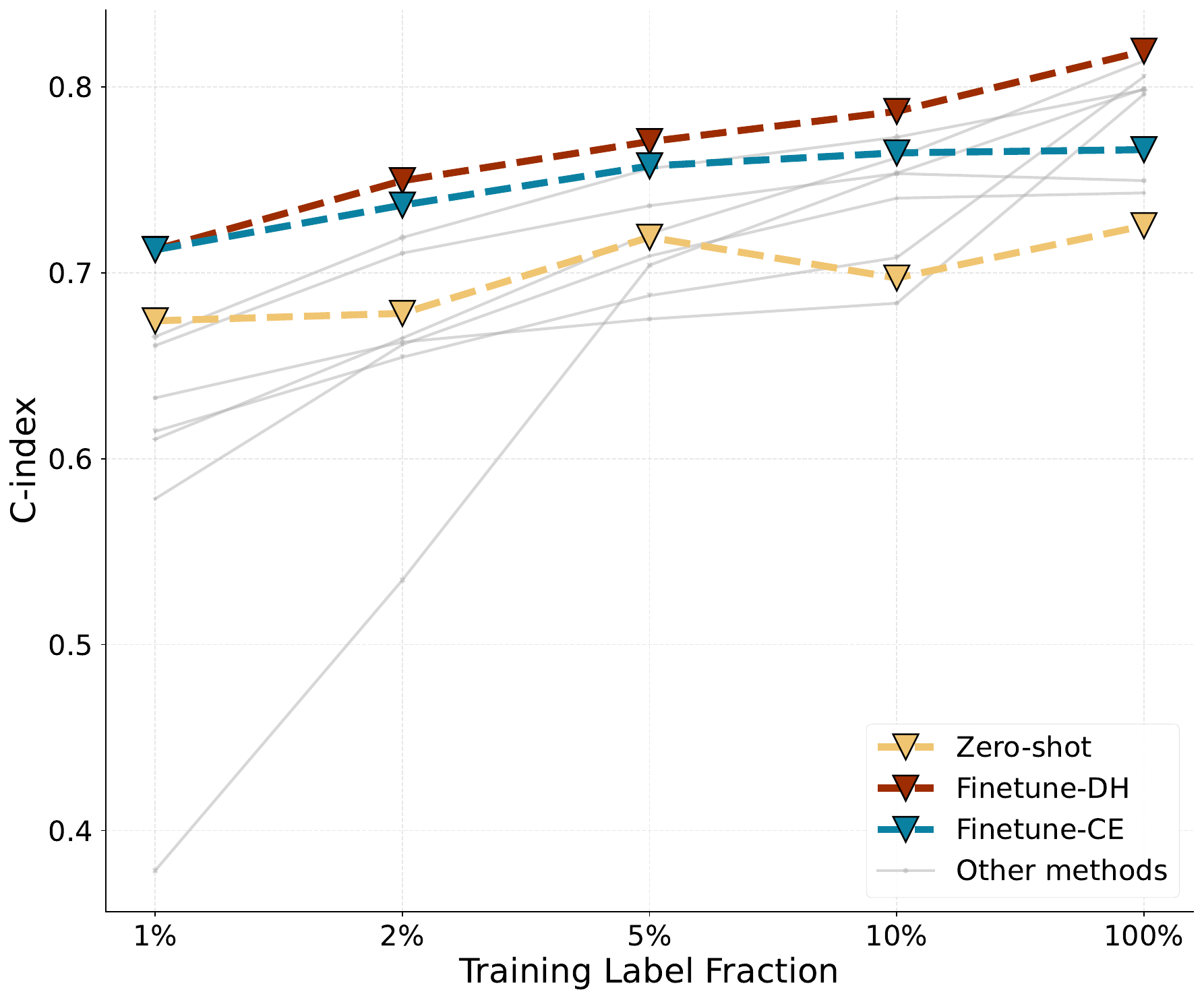}
\caption{Time-dependent $C_{td}$ as the available labeled training fraction increases from 1\% to 100\%.}
\label{fig:label_eff_cindex}
\end{figure}

Figure~\ref{fig:label_eff_cindex} provides a controlled complement to the data-regime analysis in Section~\ref{sec:single_results}. In the extreme low-label regime, zero-shot inference and DeepHit adaptation begin at essentially the same discrimination, but their trajectories diverge as more labeled supervision becomes available: zero-shot improves quickly and then saturates, whereas DeepHit continues to gain through the full-label setting. Classification fine-tuning remains lower throughout this experiment. The result is consistent with the main benchmark without duplicating its claim: native in-context inference can be useful when supervision is scarce, while DeepHit continues to improve as additional labeled data become available.


\subsection{Sensitivity to Temporal Discretization}
\label{sec:number_bins}

\begin{figure}[ht]
\centering
\includegraphics[width=0.6\linewidth]{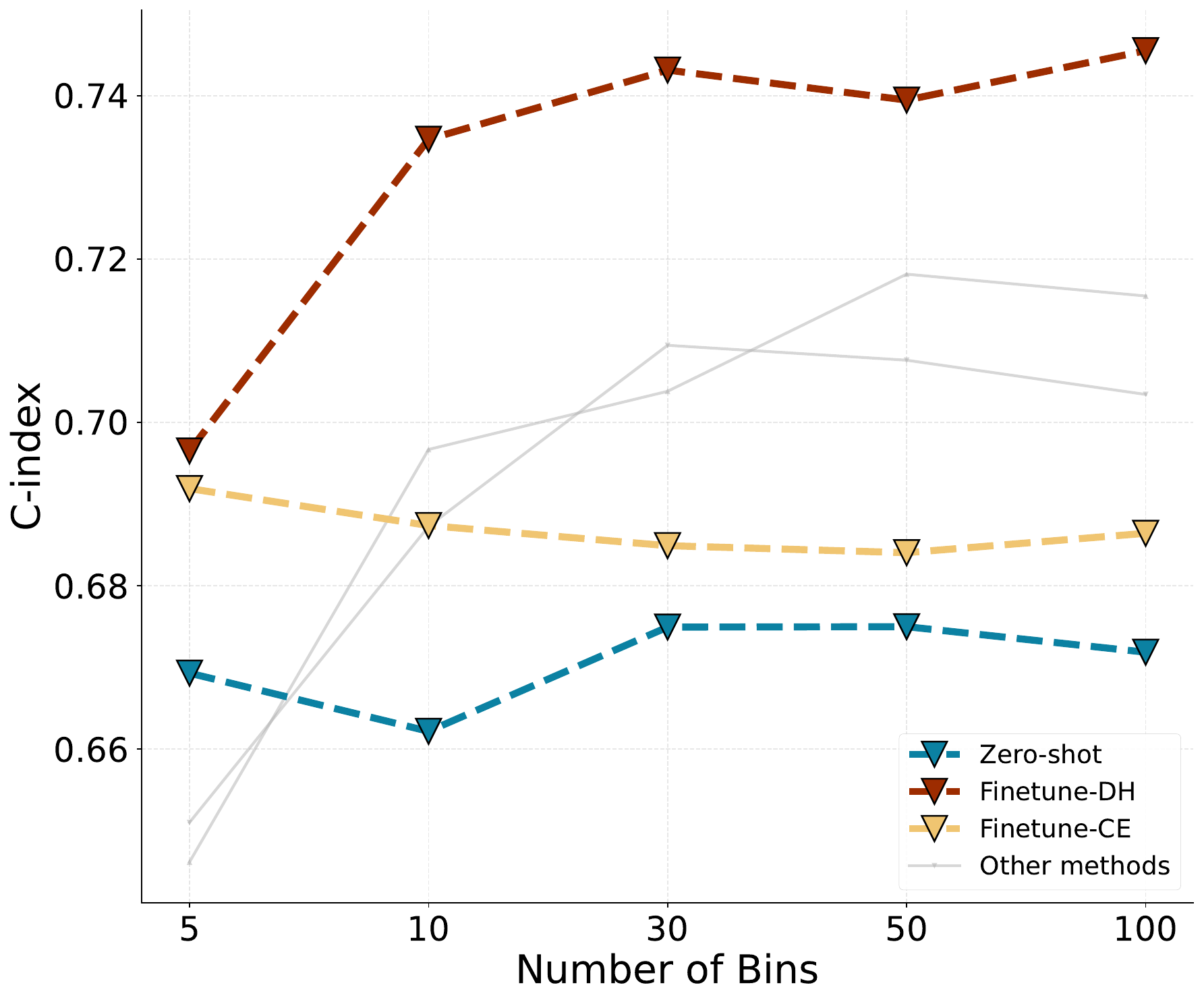}
\hfill
\caption{Sensitivity of the classification and discrete-time survival interfaces to the number of temporal bins.}
\label{fig:disc_sens}
\end{figure}

Figure~\ref{fig:disc_sens} tests whether the discrete-time conclusions depend strongly on a particular temporal resolution. DeepHit is most sensitive when moving from a very coarse grid to a moderate number of bins and then largely plateaus, whereas CE changes less in discrimination. The main implication is therefore robustness rather than a new ranking: temporal resolution matters most when the grid is too coarse, while increasing it beyond a moderate level produces diminishing returns. This supports the adaptive event-balanced discretization used in the benchmark as a way to avoid bins with too few events rather than as an explanation for the relative advantage of any particular interface.

\subsection{Efficiency--Performance Trade-Off}
\label{sec:efficiency}

The benchmark also exposes a computational distinction between the three interfaces. Zero-shot inference avoids task-specific optimization but performs native in-context prediction separately across time horizons and context samples. Classification adaptation pays an optimization cost on the temporally expanded data, whereas survival-head adaptation optimizes directly on subject-level representations and then averages predictions over five training-context samples at inference.

\begin{figure}[tb]
\centering
\begin{subfigure}{0.48\linewidth}
    \centering
    \includegraphics[width=\linewidth]{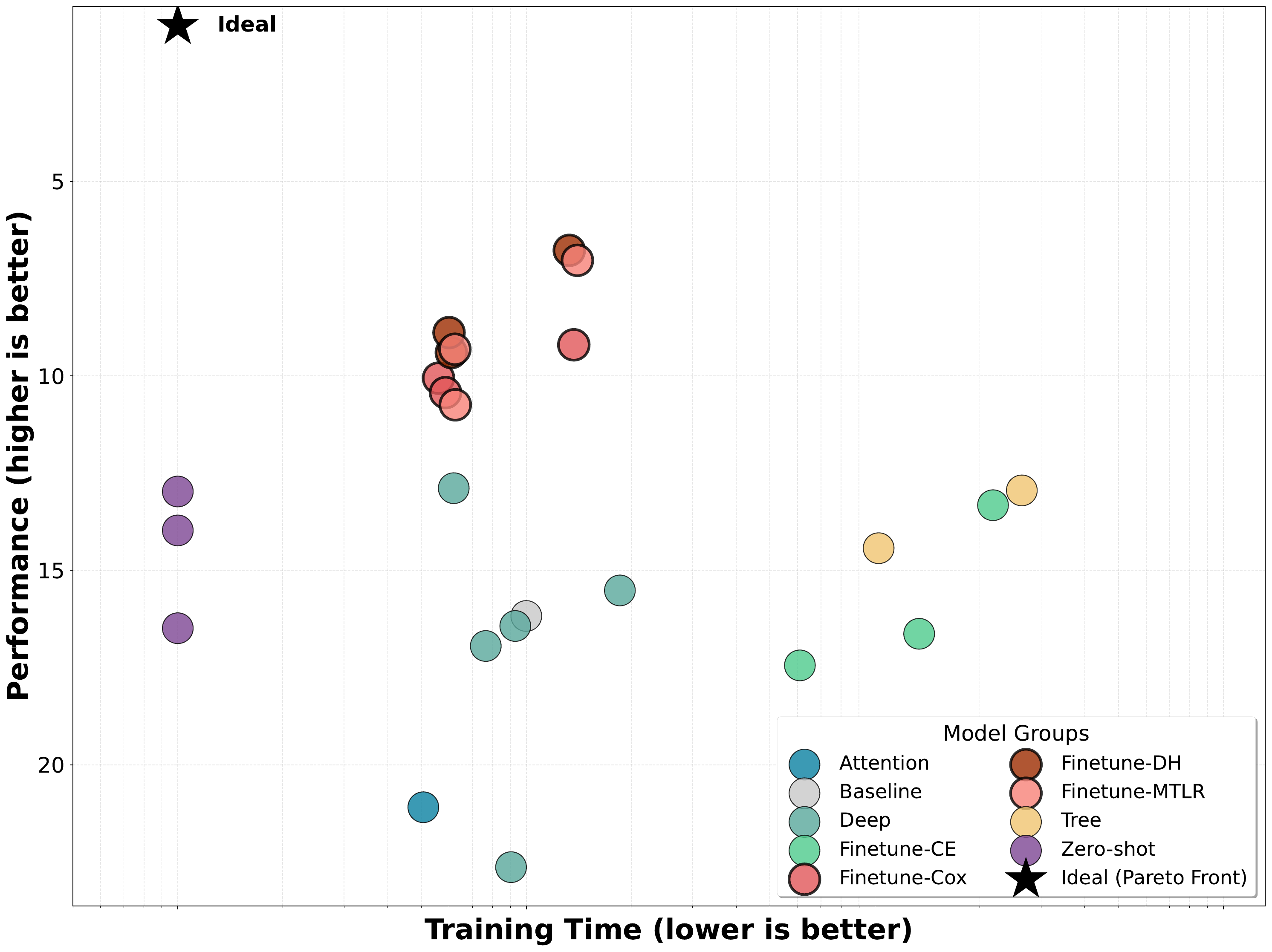}
    \caption{Training time vs. performance}
    \label{fig:efficiency_frontier}
\end{subfigure}
\hfill
\begin{subfigure}{0.48\linewidth}
    \centering
    \includegraphics[width=\linewidth]{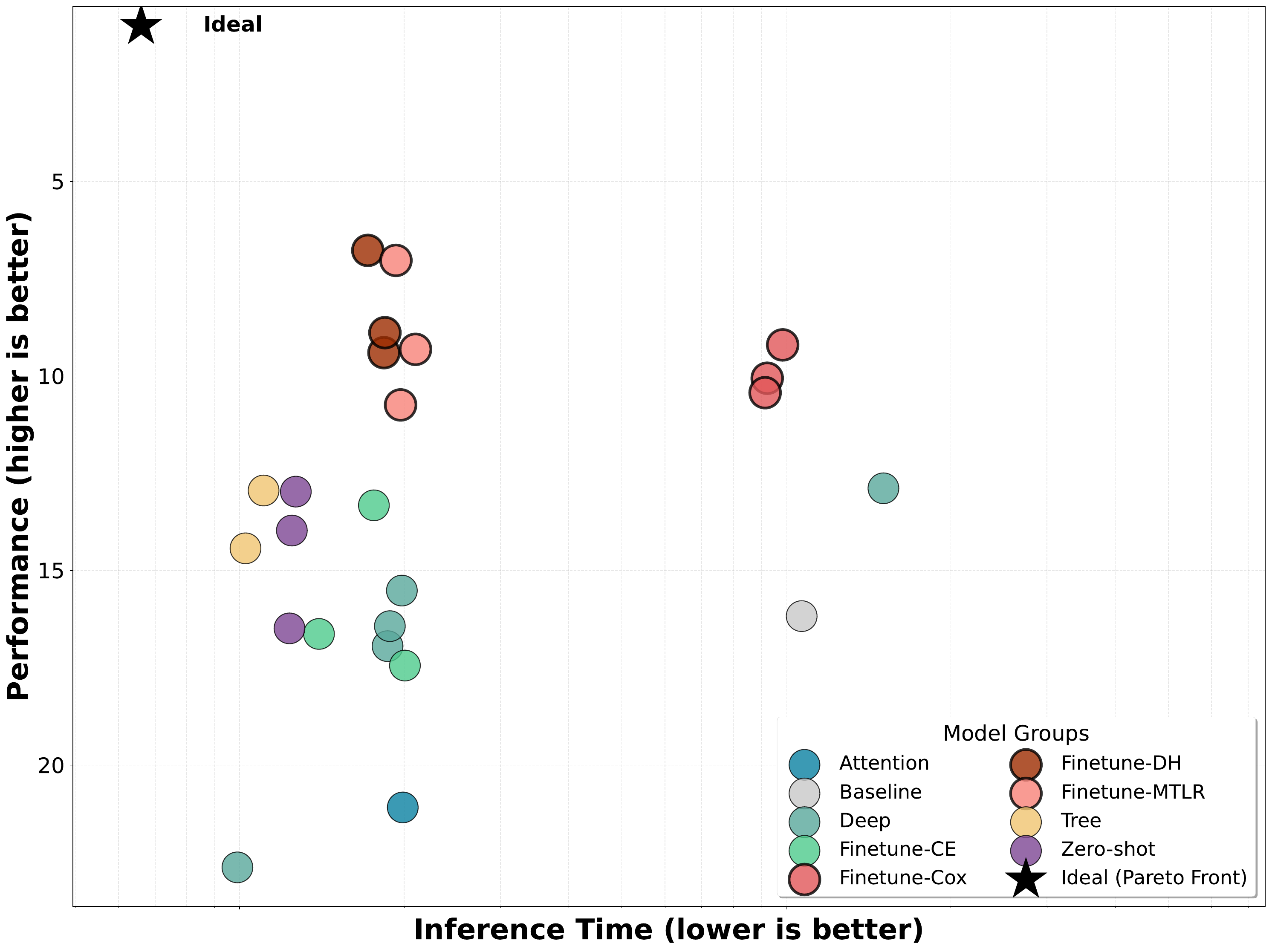}
    \caption{Inference time vs. performance}
    \label{fig:efficiency_frontier_inference}
\end{subfigure}
\caption{\textbf{Efficiency--performance trade-offs.} Performance versus computational cost on a log scale. Training time for the tuned non-pretrained baselines includes their per-fold Optuna search, which the TabFM variants do not perform, so these values describe the cost of the reported experimental pipelines rather than a direct comparison of optimization speed.}
\label{fig:efficiency}
\end{figure}

Figure~\ref{fig:efficiency} adds a practical perspective to the accuracy results rather than changing their ordering. The absence of optimization makes zero-shot attractive when fitting cost is the primary constraint, but repeated horizon-wise in-context prediction shifts part of that cost to inference. Supervised adaptation trades additional fitting time for the performance gains seen in the main benchmark, with the survival-head route avoiding the full subject--horizon expansion used by CE. Because the tuned baselines include their hyperparameter-search cost whereas the TabFM variants use fixed settings, the figure should be interpreted as a comparison of the full reported pipelines, not as evidence that one model family is intrinsically faster to optimize.

\subsection{Risk Stratification and Survival Curve Separation}
\label{sec:risk_stratification}
\begin{figure}[tb]
\centering
\textbf{Other methods} \\
\textbf{DeepHit} \hfill \textbf{DeepSurv} \hfill \textbf{DySurv} \\
\begin{subfigure}[t]{0.32\linewidth}
    \includegraphics[width=\linewidth]{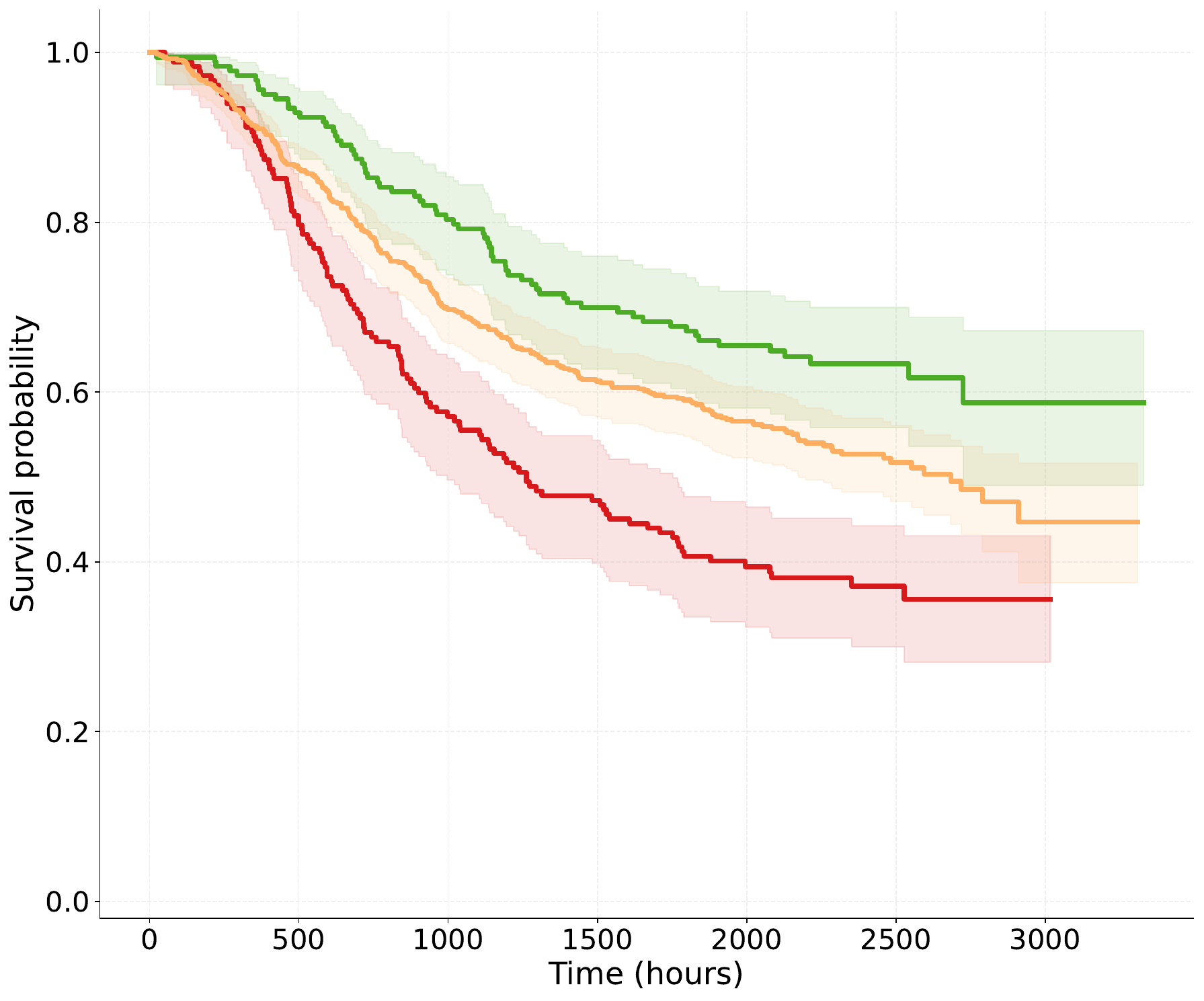}
\end{subfigure}
\hfill
\begin{subfigure}[t]{0.32\linewidth}
    \includegraphics[width=\linewidth]{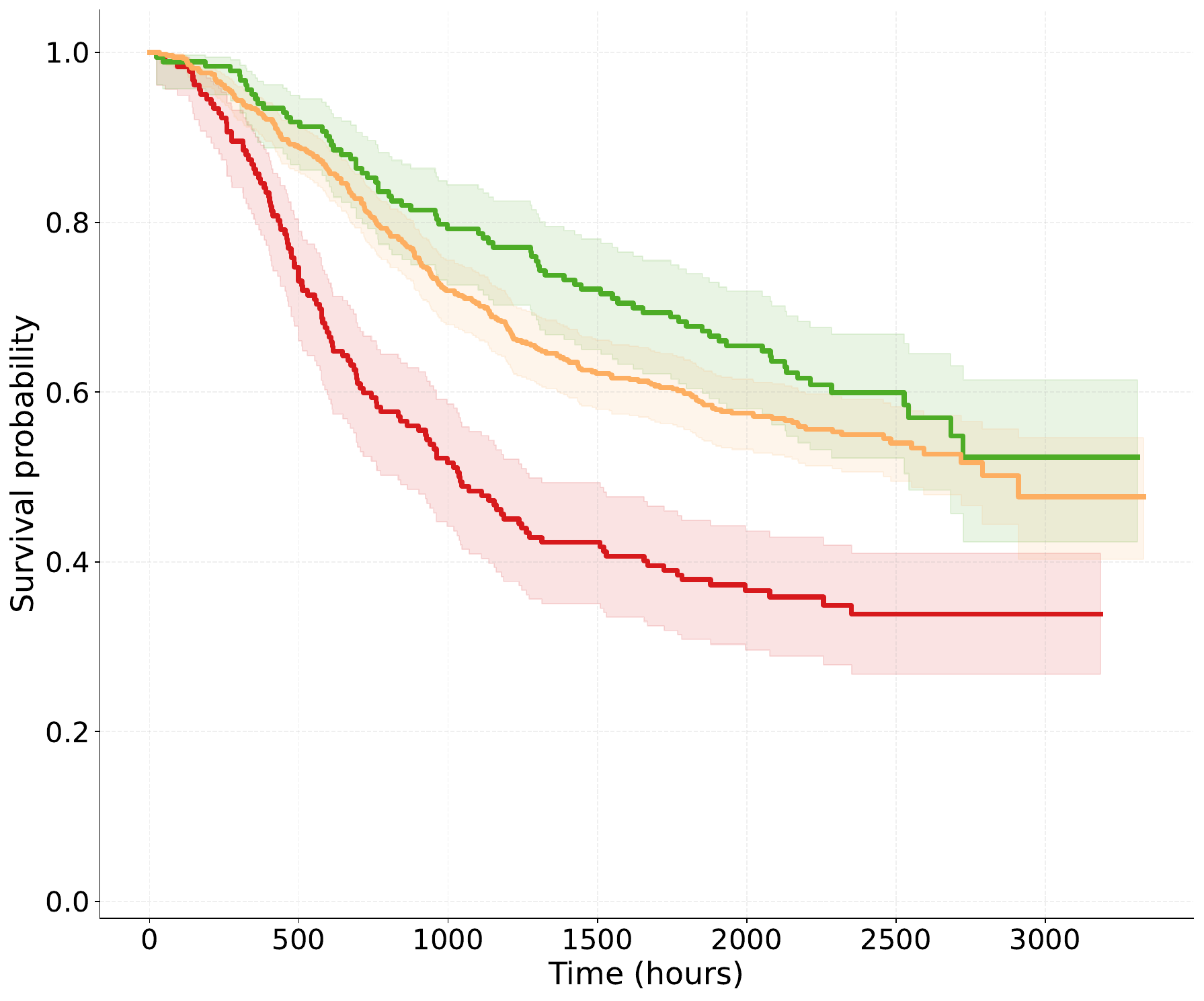}
\end{subfigure}
\hfill
\begin{subfigure}[t]{0.32\linewidth}
    \includegraphics[width=\linewidth]{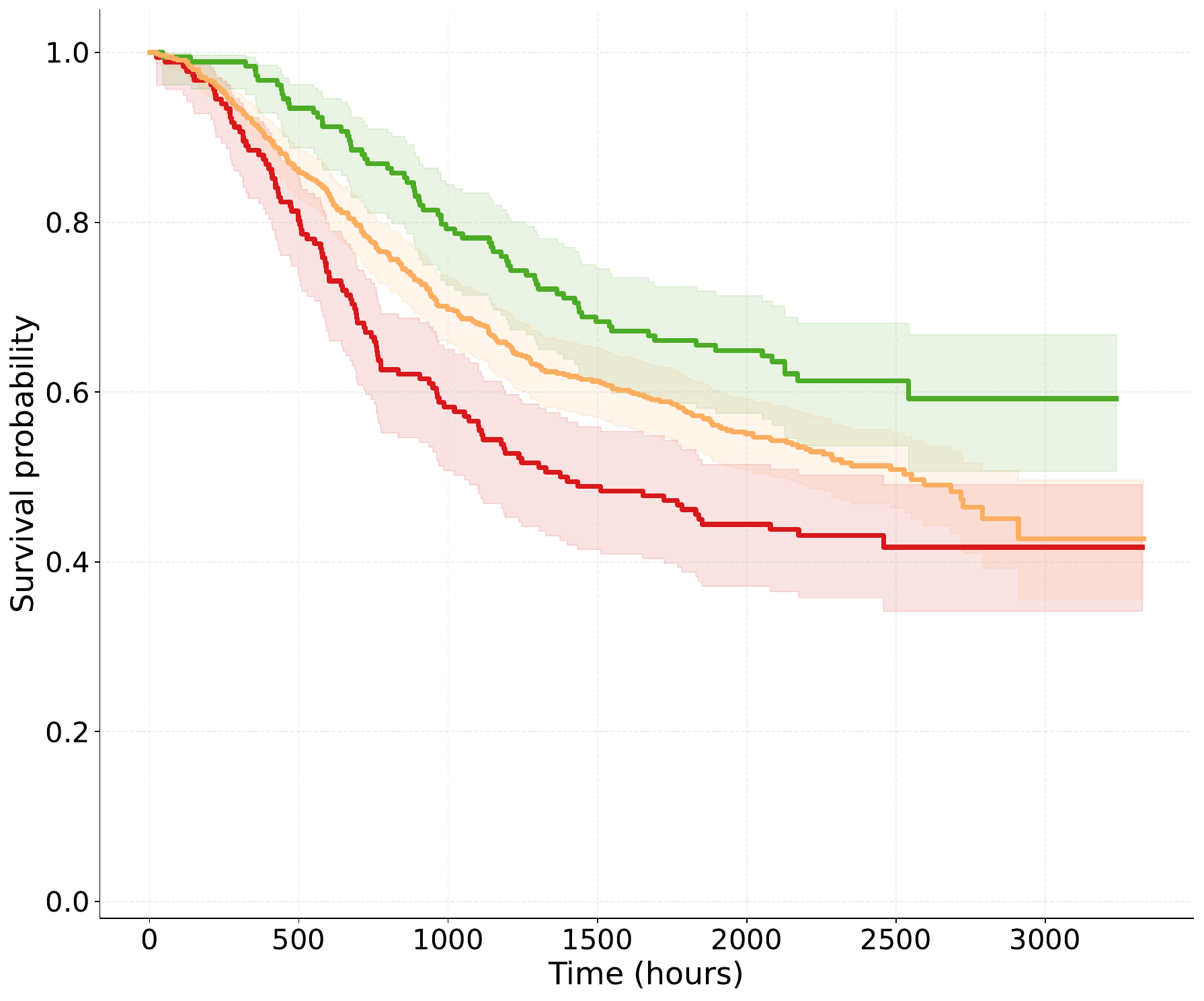}
\end{subfigure}
\vspace{0.4em}
\textbf{Tabular Foundation Models} \\
\textbf{TabPFN} \hfill \textbf{TabDPT} \hfill \textbf{TabICL} \\
\begin{subfigure}[t]{0.32\linewidth}
    \includegraphics[width=\linewidth]{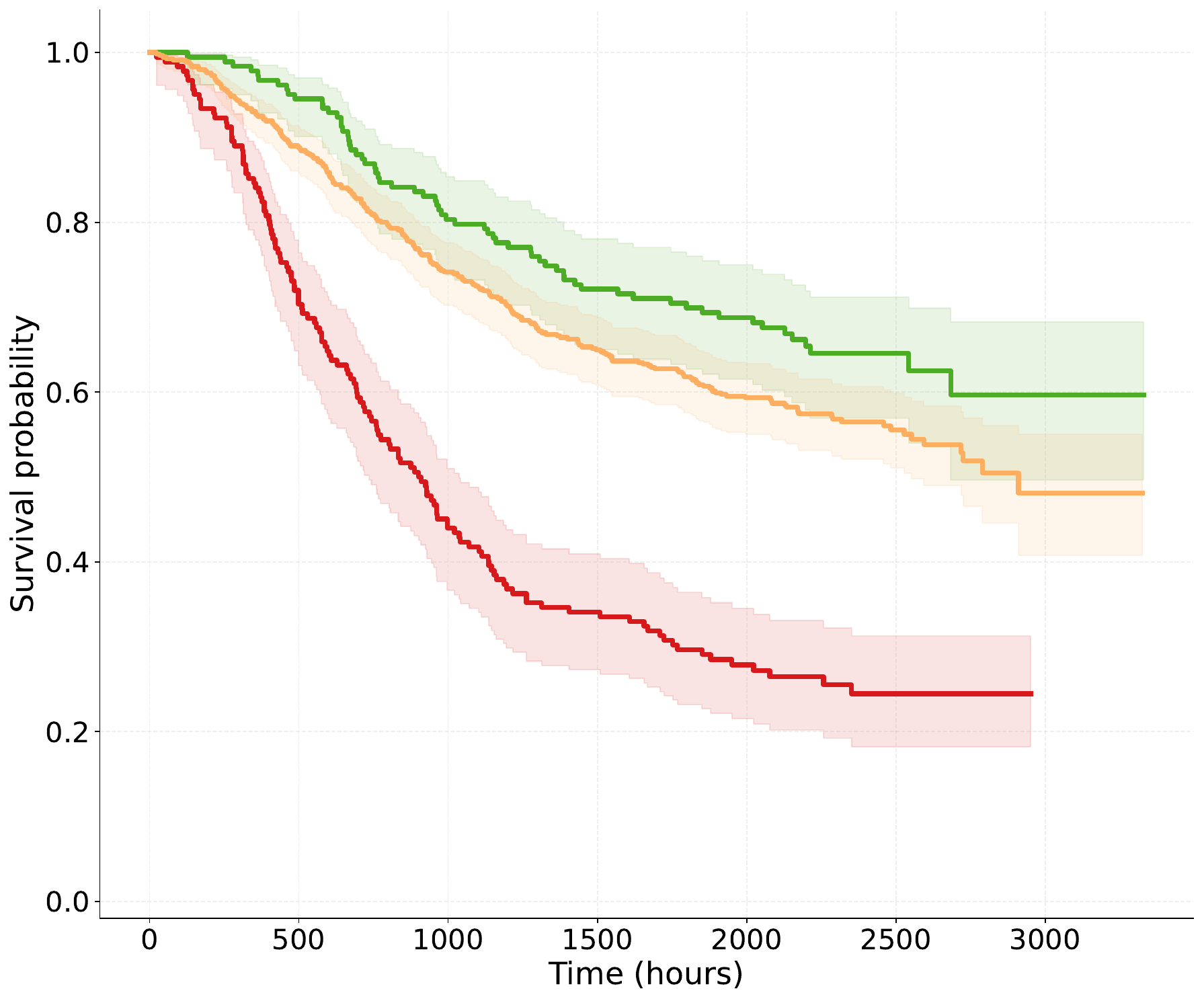}
\end{subfigure}
\hfill
\begin{subfigure}[t]{0.32\linewidth}
    \includegraphics[width=\linewidth]{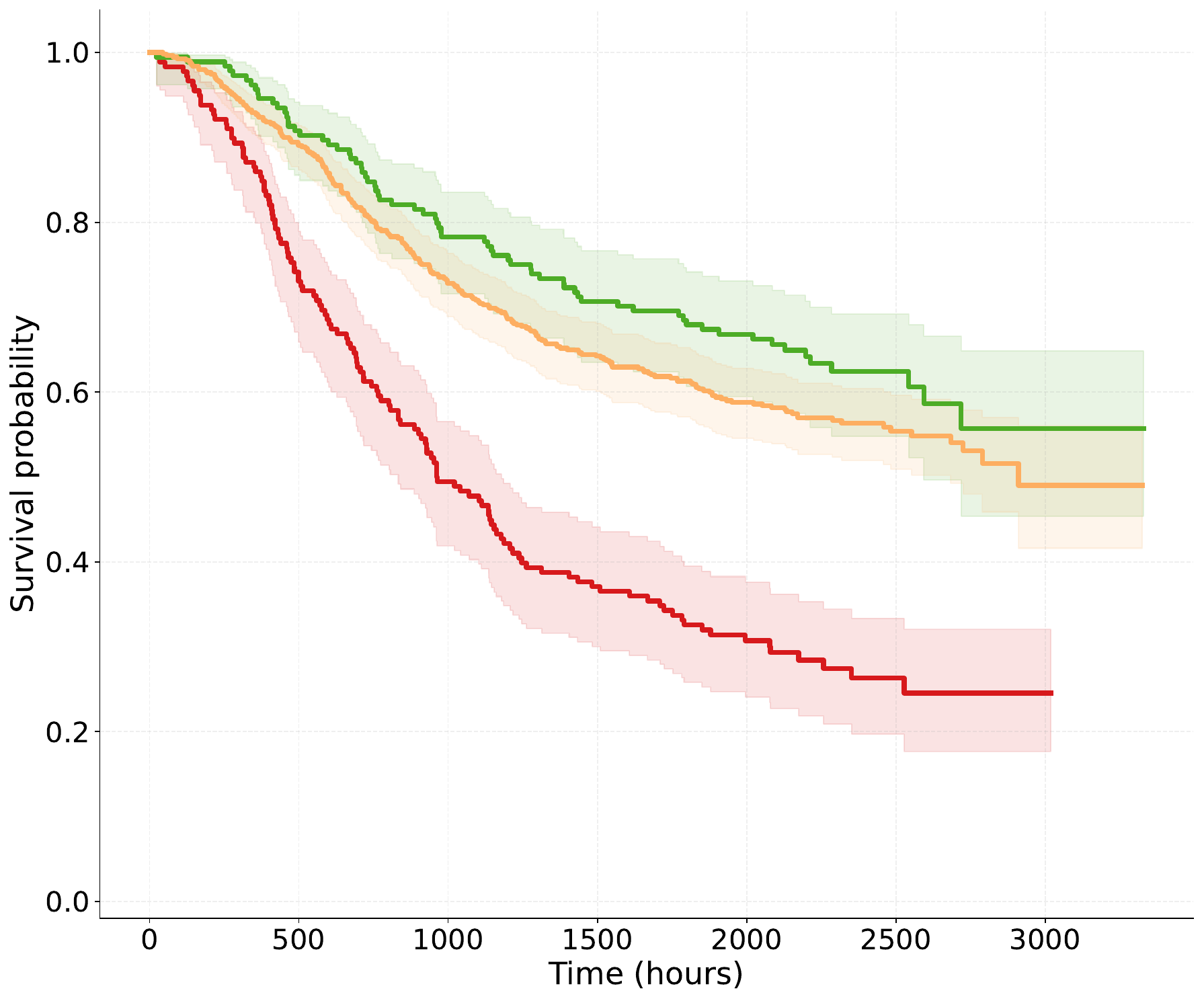}
\end{subfigure}
\hfill
\begin{subfigure}[t]{0.32\linewidth}
    \includegraphics[width=\linewidth]{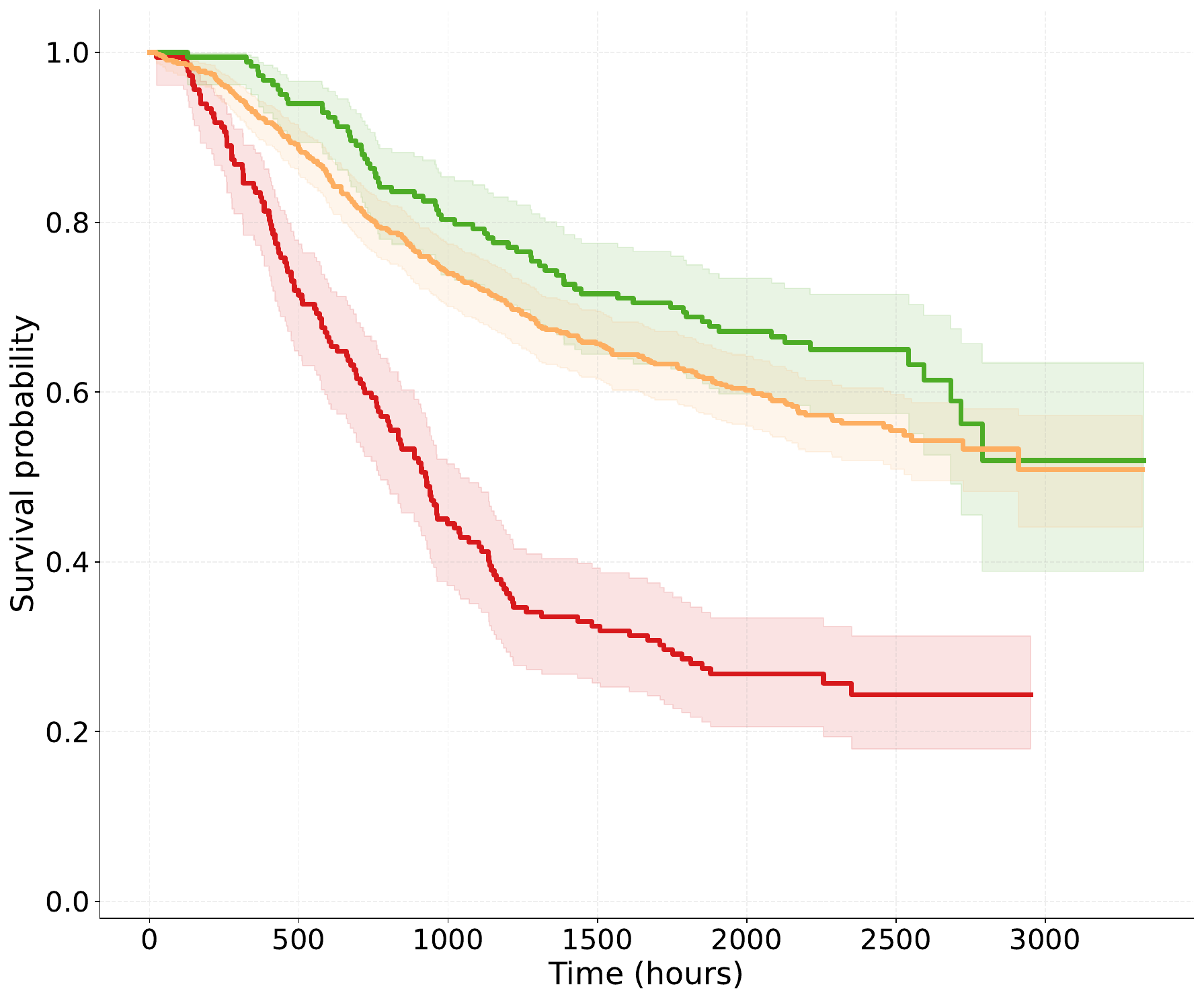}
\end{subfigure}
\vspace{0.3em}
\textit{(a) Zero-shot} \\
\vspace{0.4em}
\begin{subfigure}[t]{0.32\linewidth}
    \includegraphics[width=\linewidth]{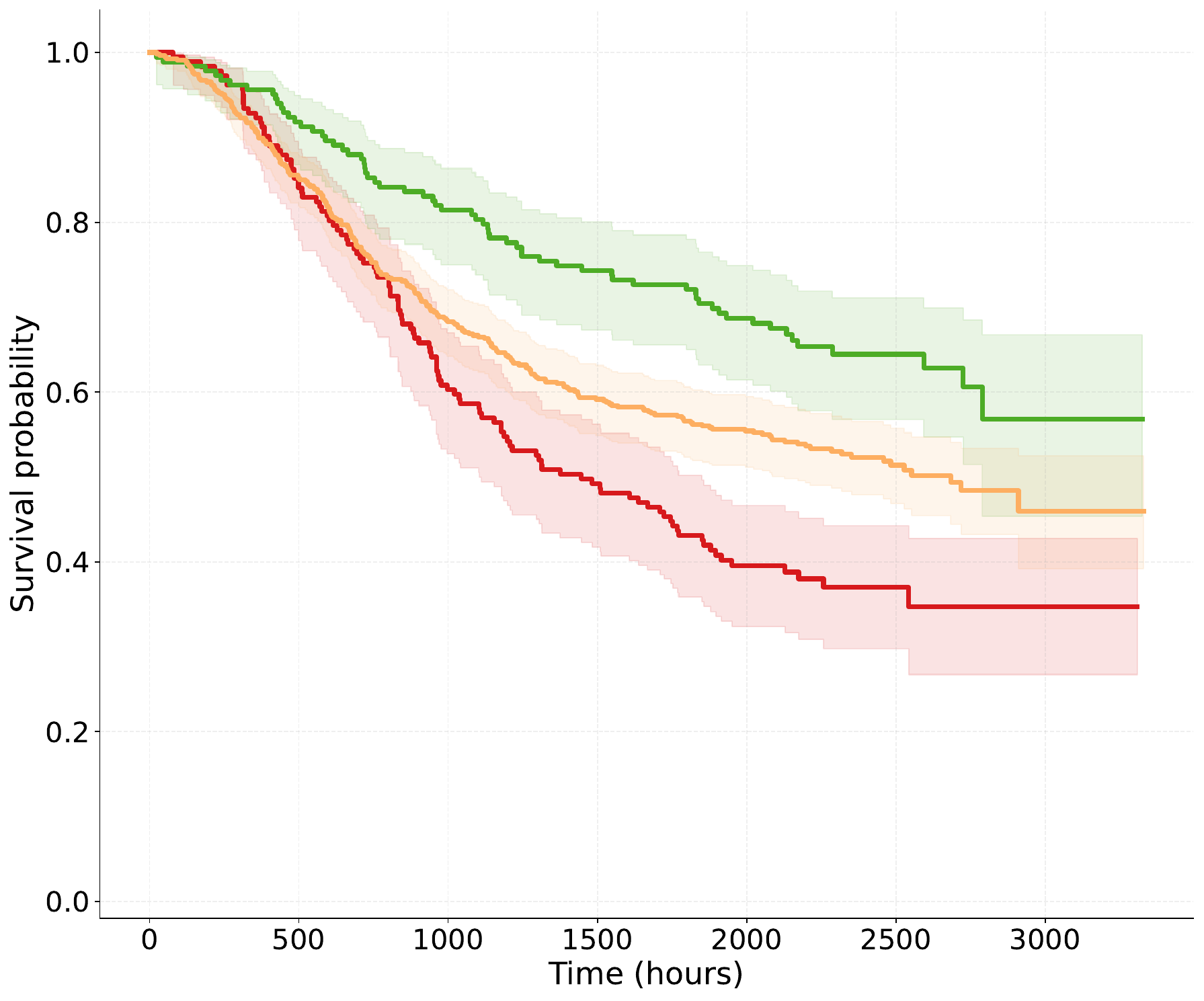}
\end{subfigure}
\hfill
\begin{subfigure}[t]{0.32\linewidth}
    \includegraphics[width=\linewidth]{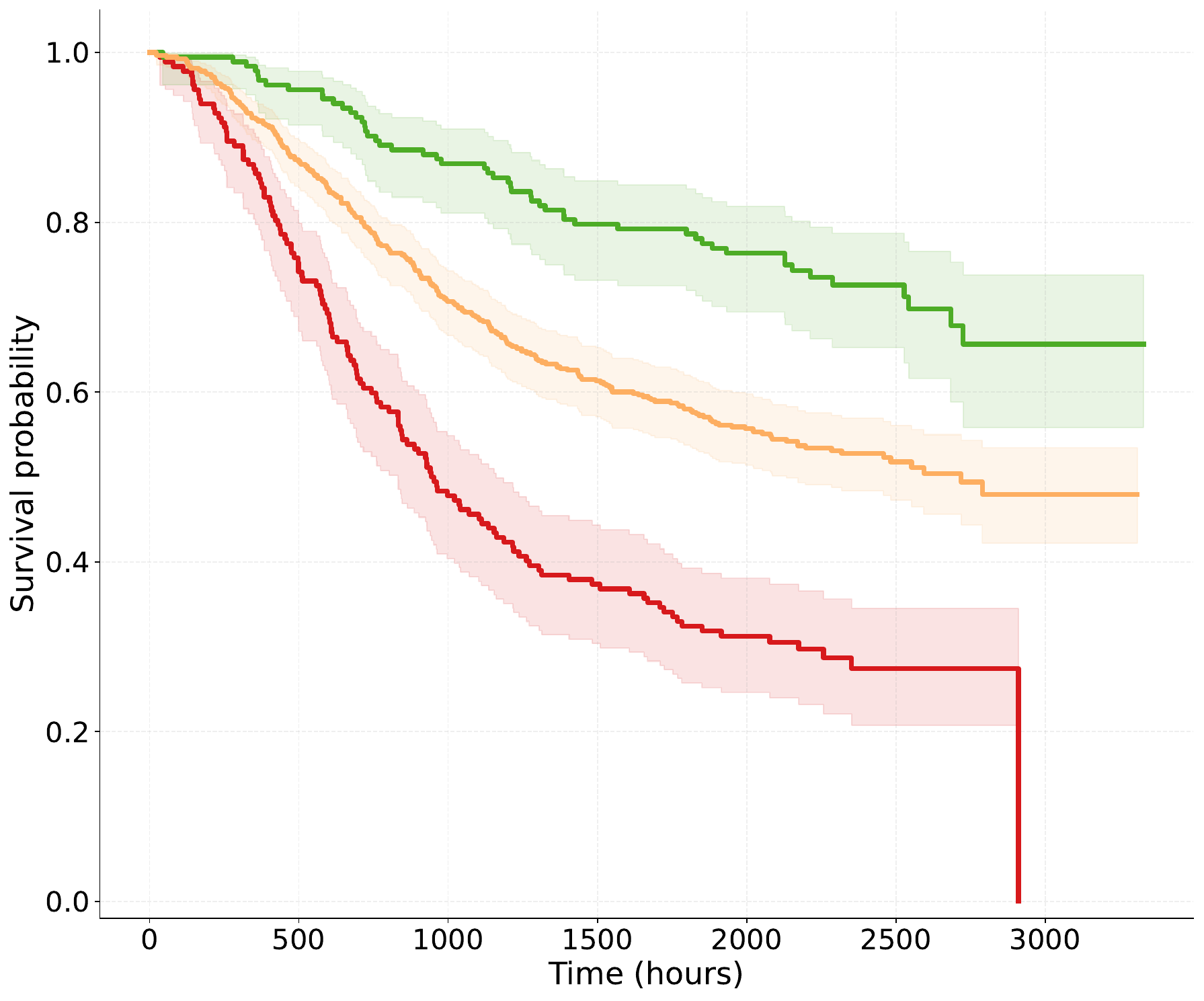}
\end{subfigure}
\hfill
\begin{subfigure}[t]{0.32\linewidth}
    \includegraphics[width=\linewidth]{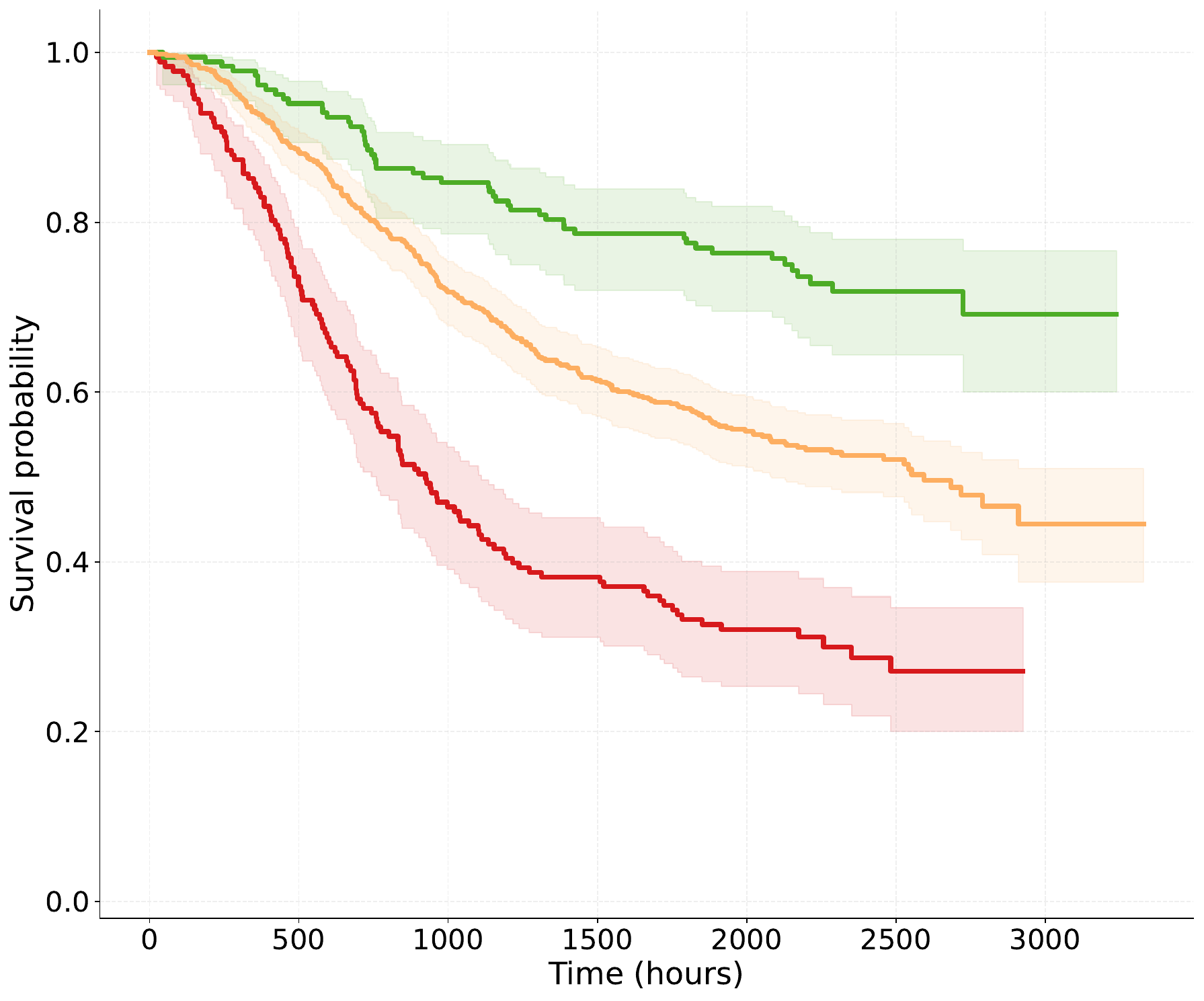}
\end{subfigure}
\vspace{0.3em}
\textit{(b) Classification fine-tuning} \\
\vspace{0.6em}
\begin{subfigure}[t]{0.32\linewidth}
    \includegraphics[width=\linewidth]{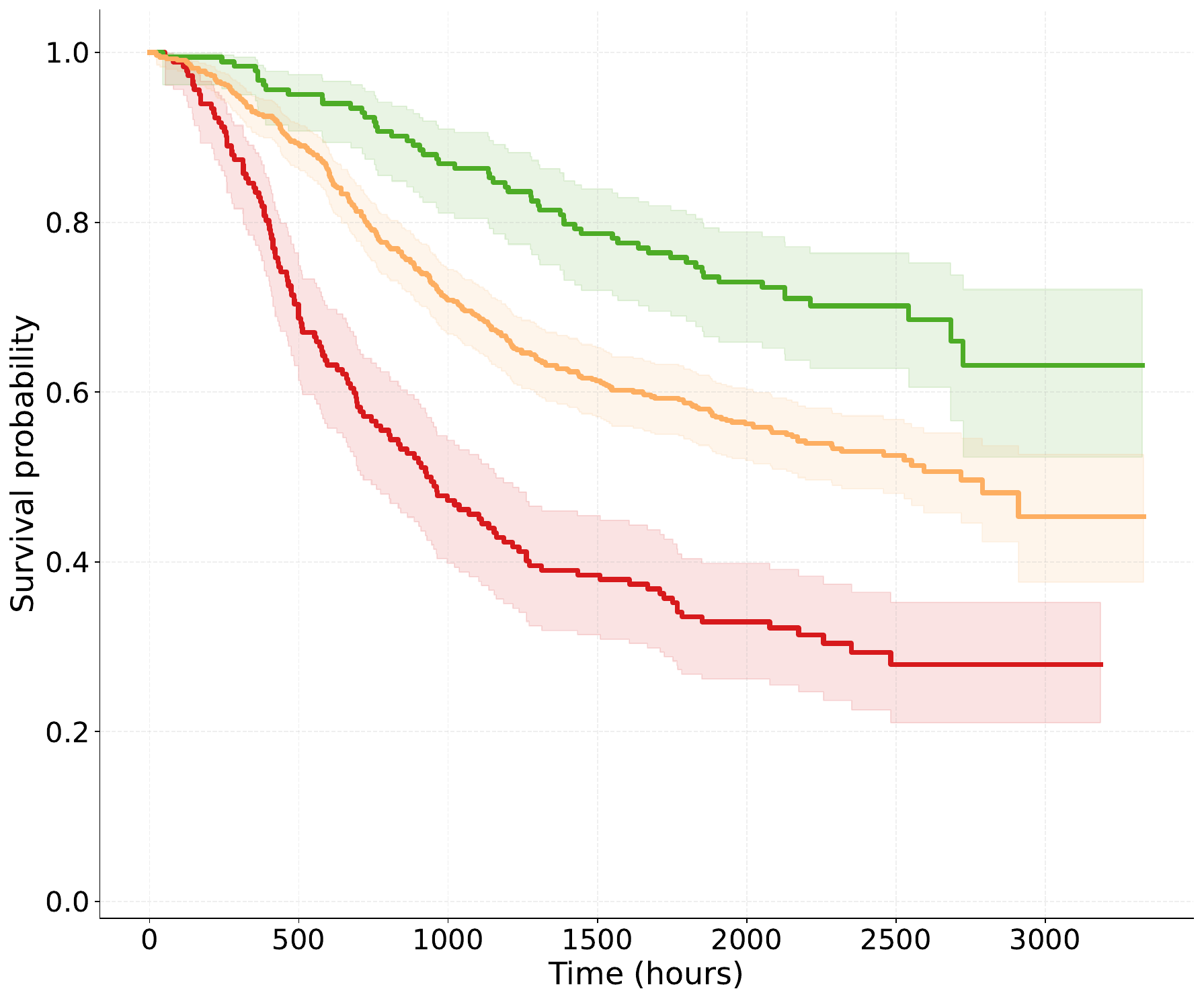}
\end{subfigure}
\hfill
\begin{subfigure}[t]{0.32\linewidth}
    \includegraphics[width=\linewidth]{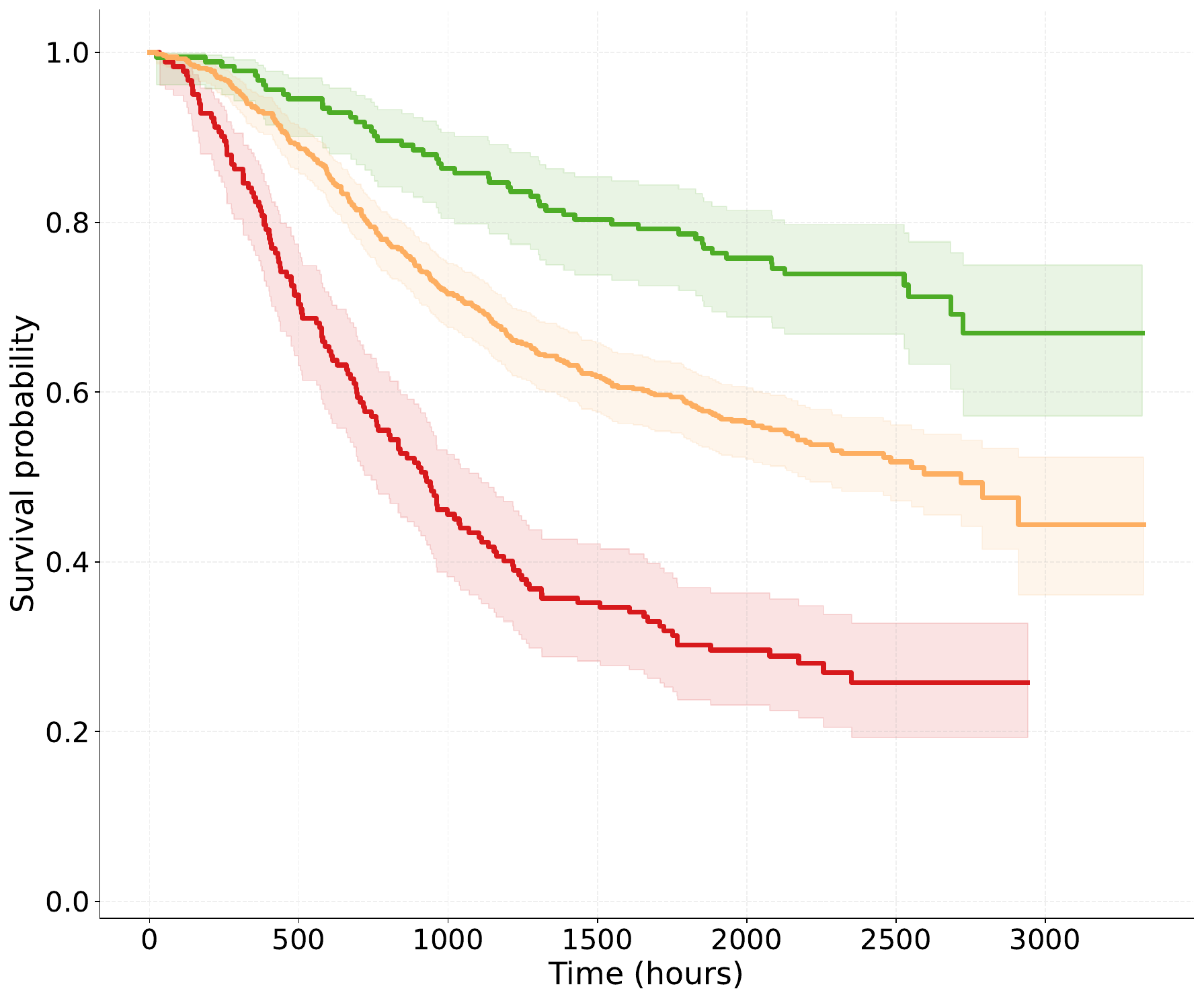}
\end{subfigure}
\hfill
\begin{subfigure}[t]{0.32\linewidth}
    \includegraphics[width=\linewidth]{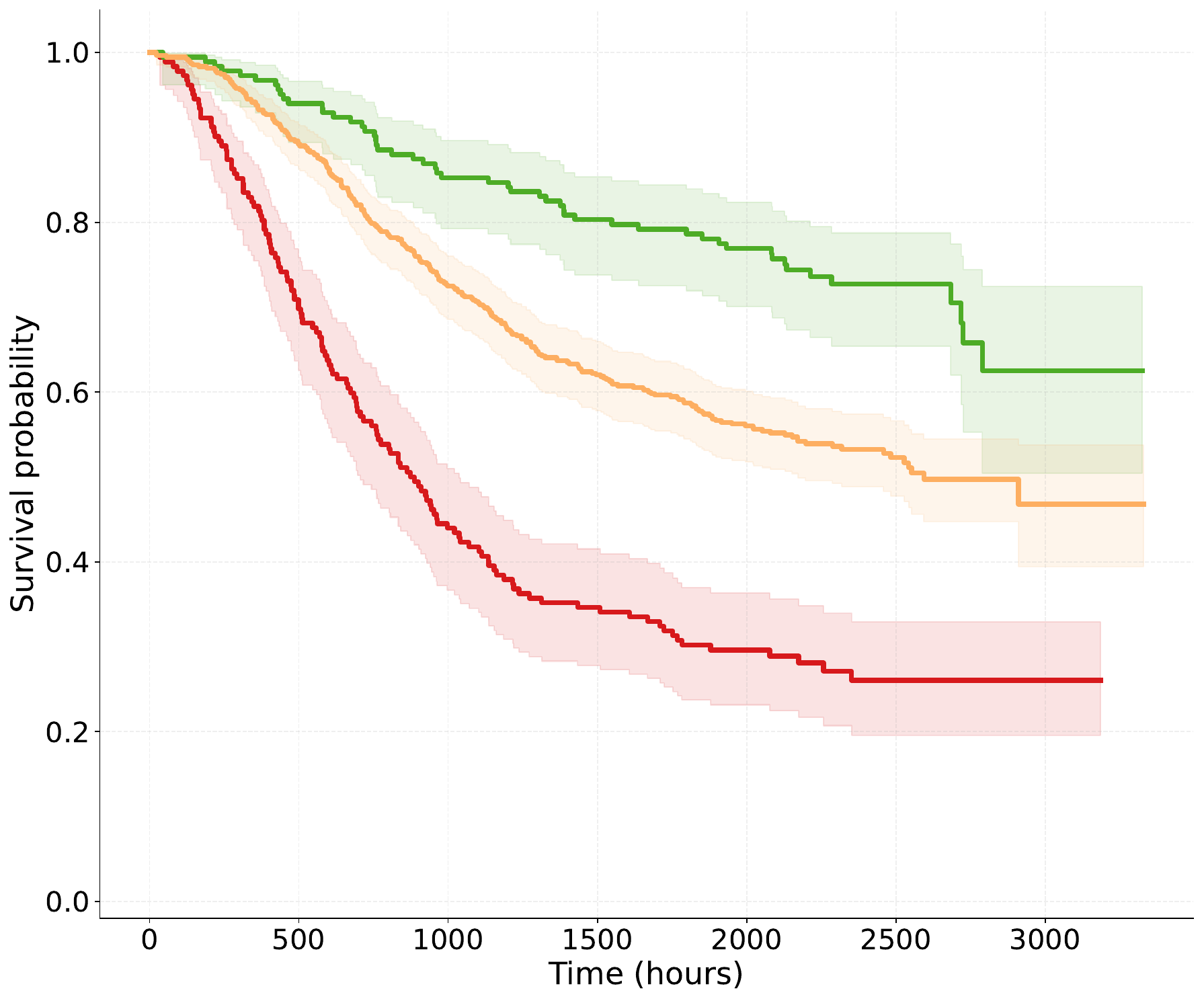}
\end{subfigure}
\vspace{0.3em}
\textit{(c) Survival head adaptation (Cox; frozen backbone)}
\caption{\textbf{Risk-stratified Kaplan--Meier curves on the COLON data set.}}
\label{fig:risk_stratification_COLON}
\end{figure}

\begin{figure}[tb]
\centering
\textbf{Other methods} \\
\textbf{CoxPH} \hfill \textbf{Gradient Boosting} \hfill \textbf{DeepSurv} \\
\begin{subfigure}[t]{0.32\linewidth}
    \includegraphics[width=\linewidth]{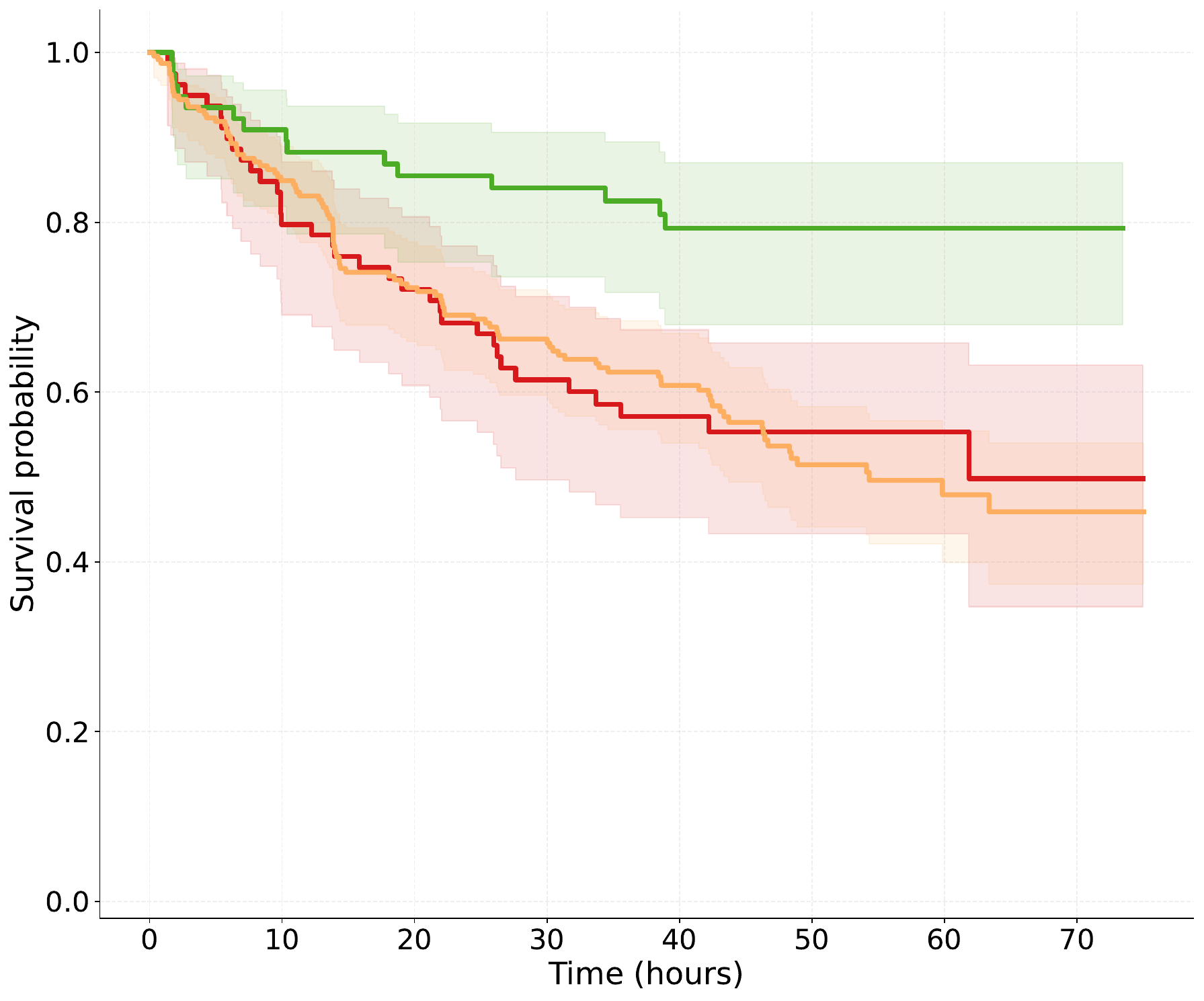}
\end{subfigure}
\hfill
\begin{subfigure}[t]{0.32\linewidth}
    \includegraphics[width=\linewidth]{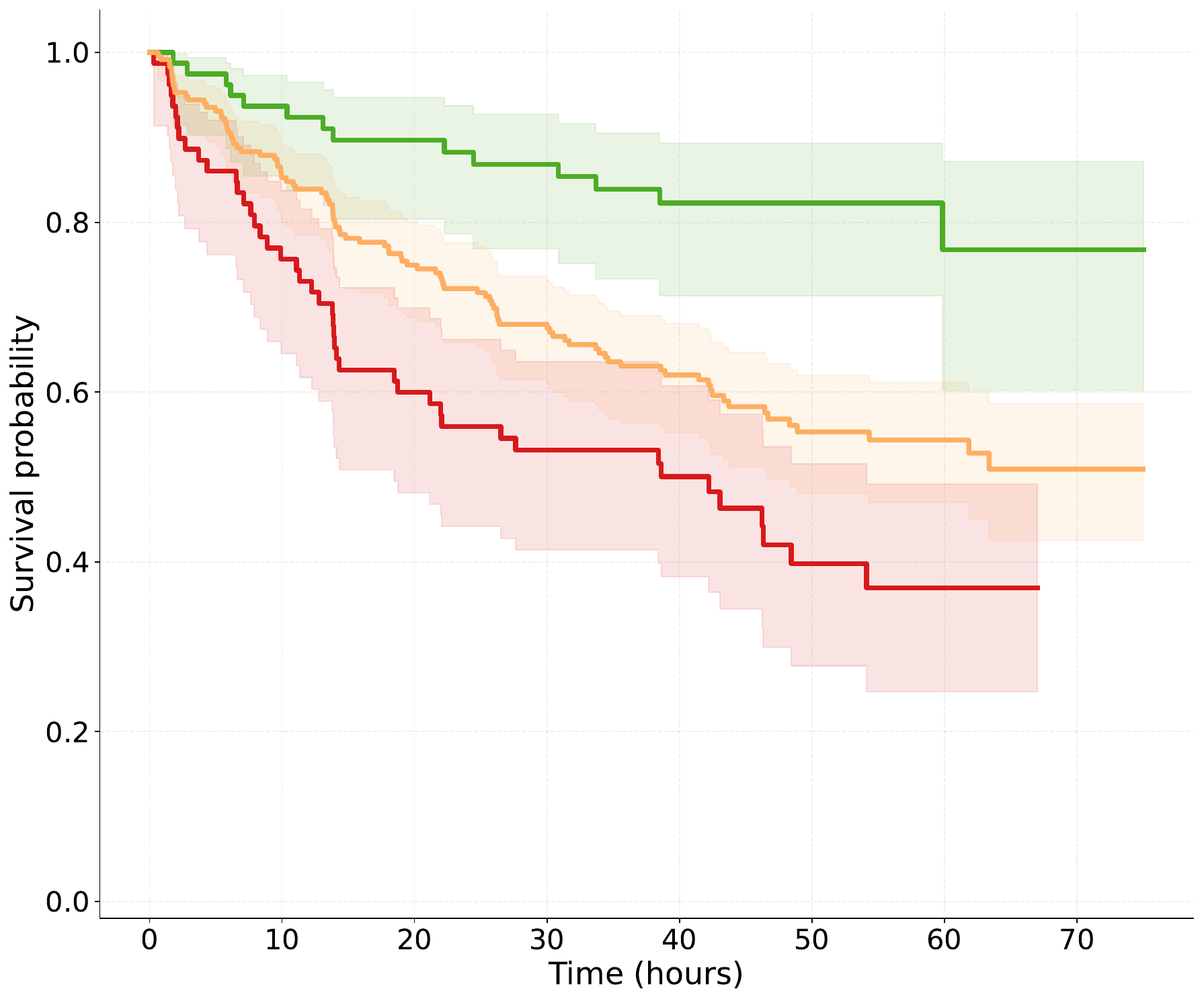}
\end{subfigure}
\hfill
\begin{subfigure}[t]{0.32\linewidth}
    \includegraphics[width=\linewidth]{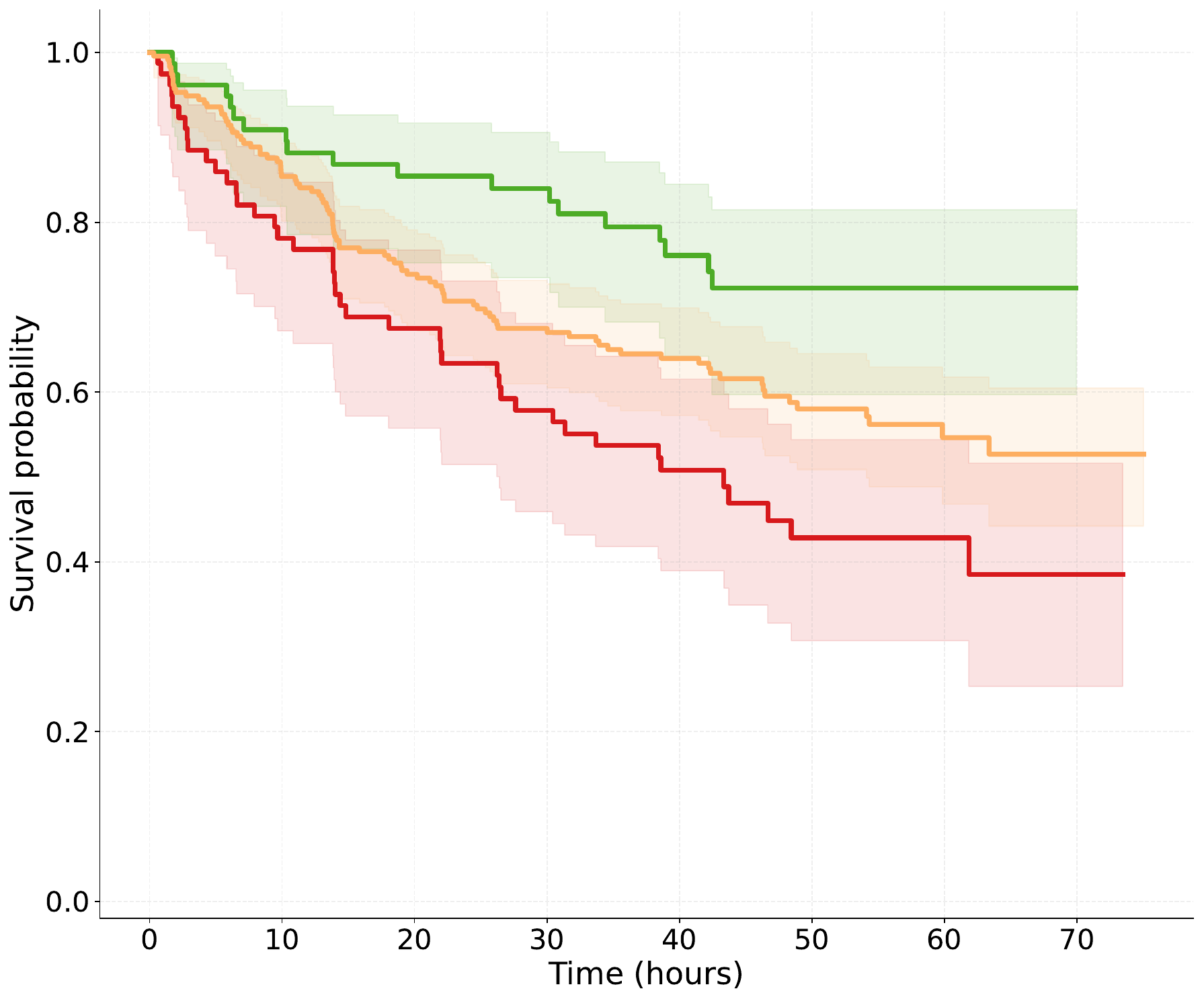}
\end{subfigure}
\vspace{0.4em}
\textbf{Tabular Foundation Models} \\
\textbf{TabPFN} \hfill \textbf{TabDPT} \hfill \textbf{TabICL} \\
\begin{subfigure}[t]{0.32\linewidth}
    \includegraphics[width=\linewidth]{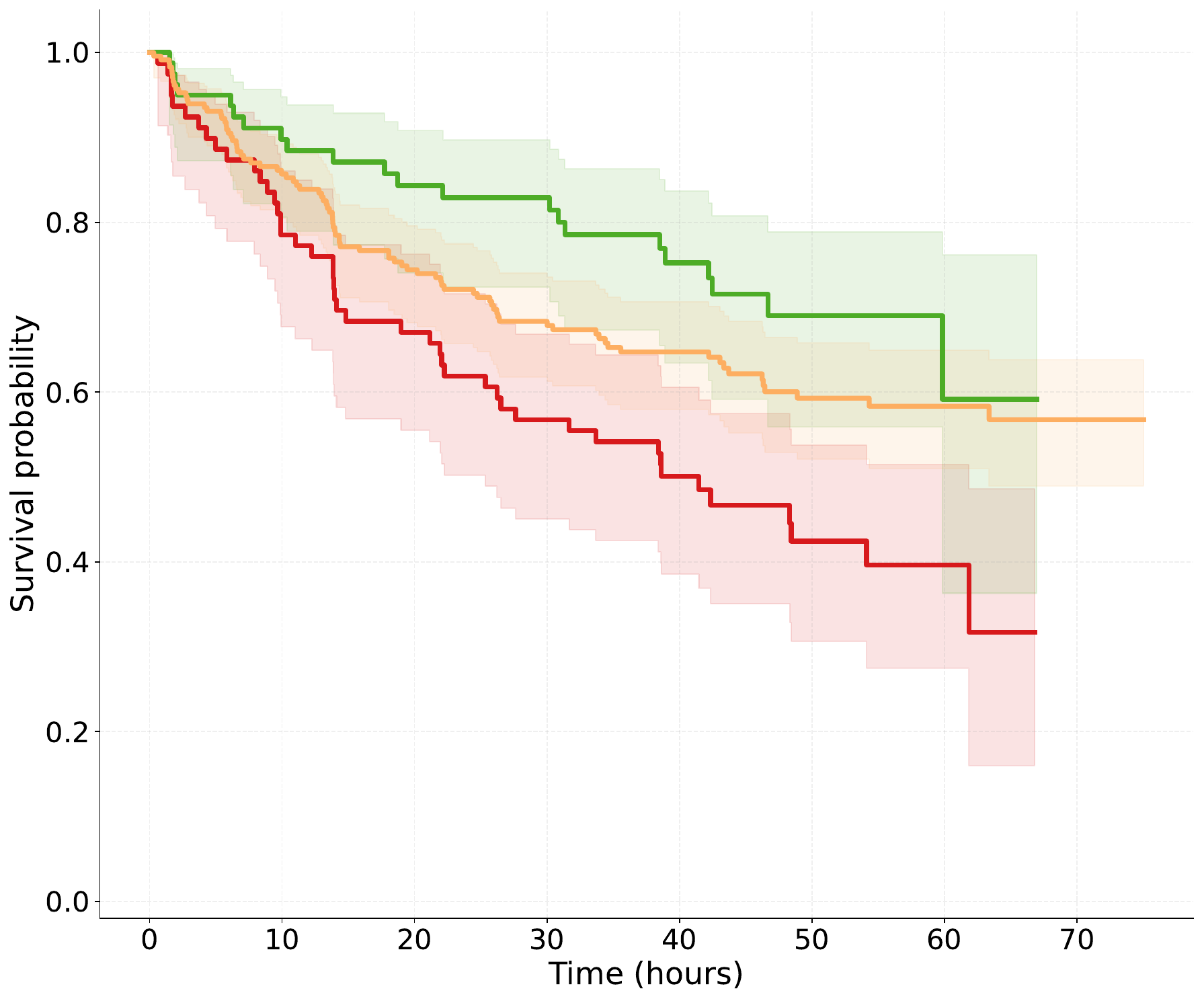}
\end{subfigure}
\hfill
\begin{subfigure}[t]{0.32\linewidth}
    \includegraphics[width=\linewidth]{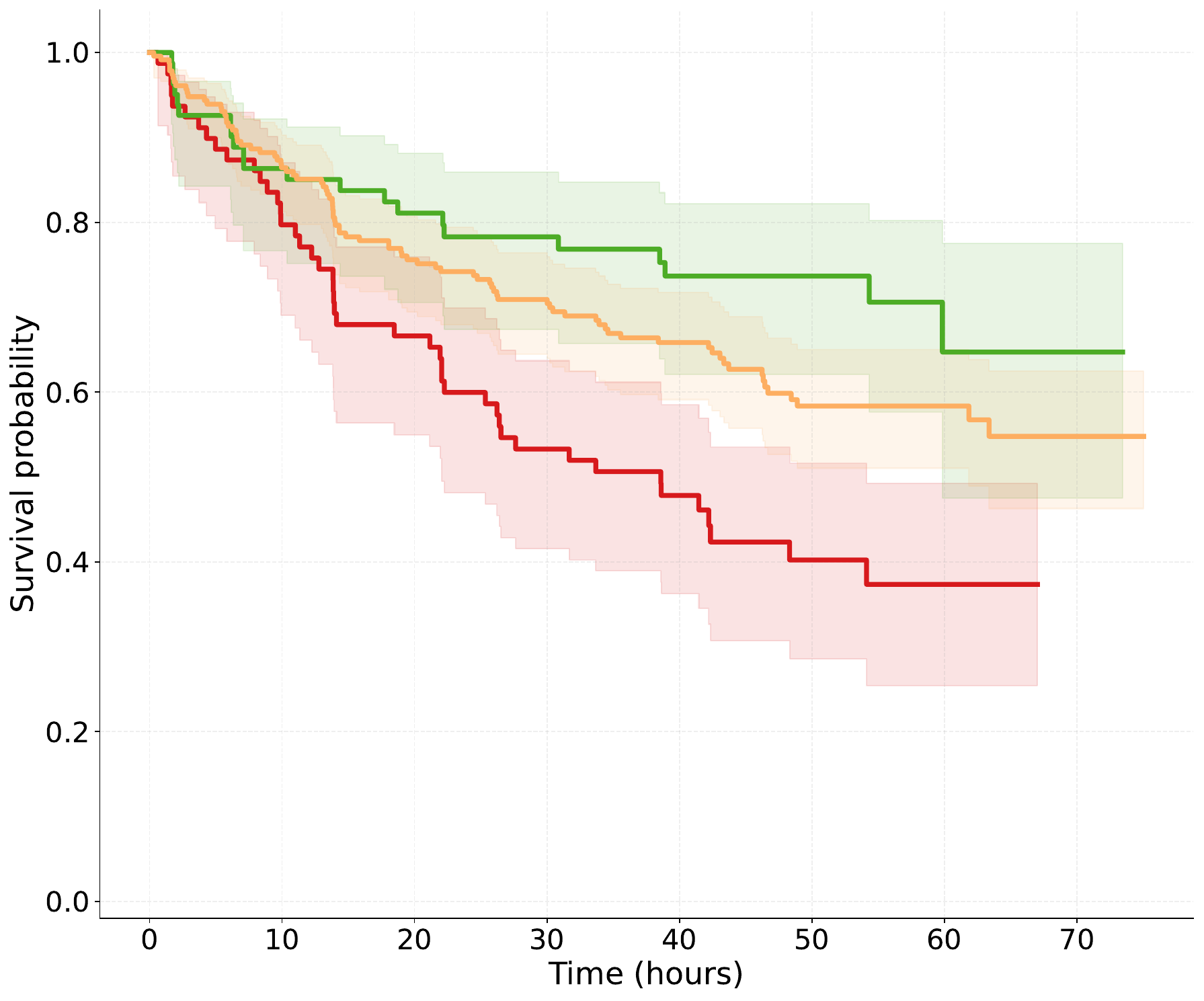}
\end{subfigure}
\hfill
\begin{subfigure}[t]{0.32\linewidth}
    \includegraphics[width=\linewidth]{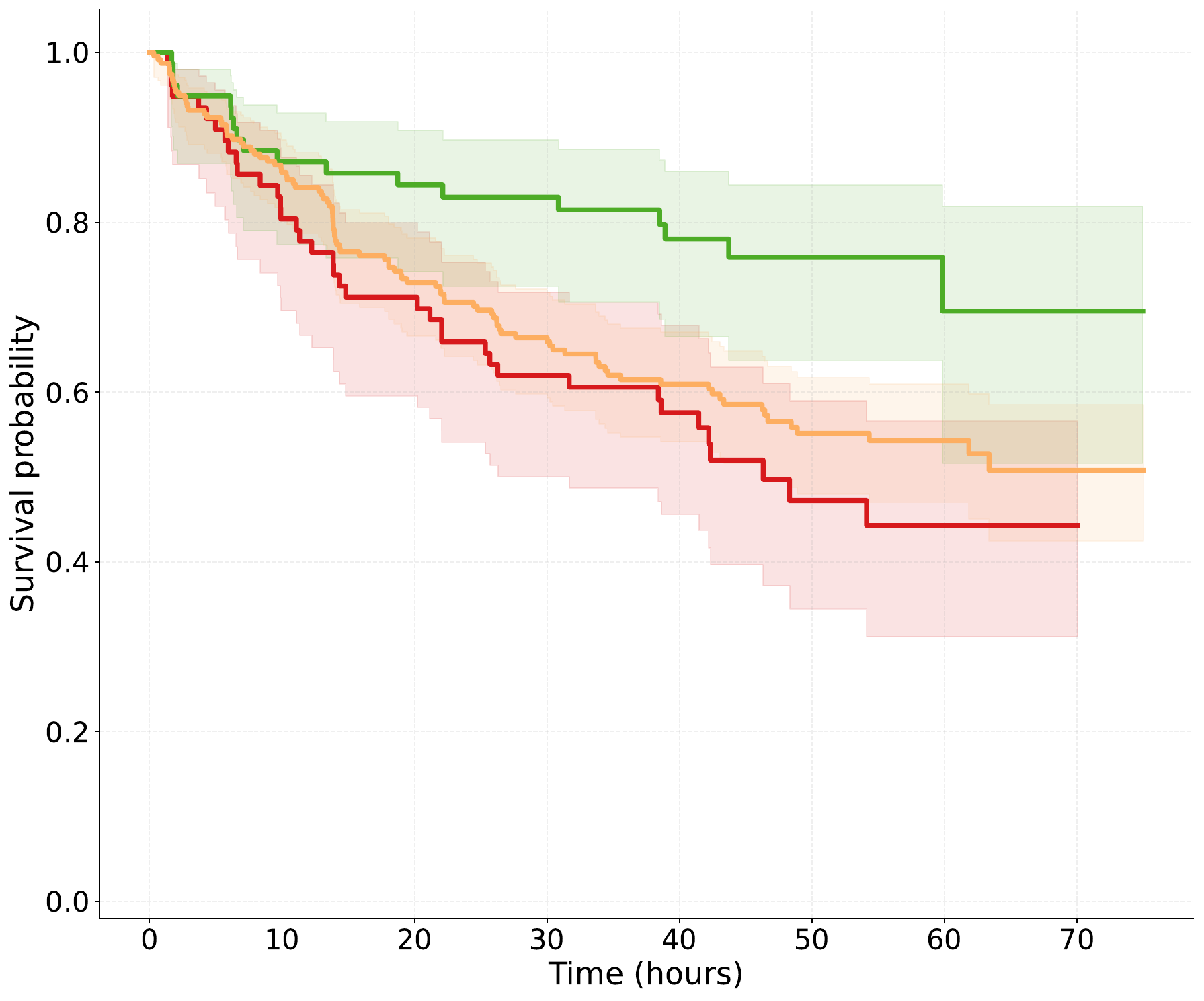}
\end{subfigure}
\vspace{0.3em}
\textit{(a) Zero-shot)} \\
\vspace{0.4em}
\begin{subfigure}[t]{0.32\linewidth}
    \includegraphics[width=\linewidth]{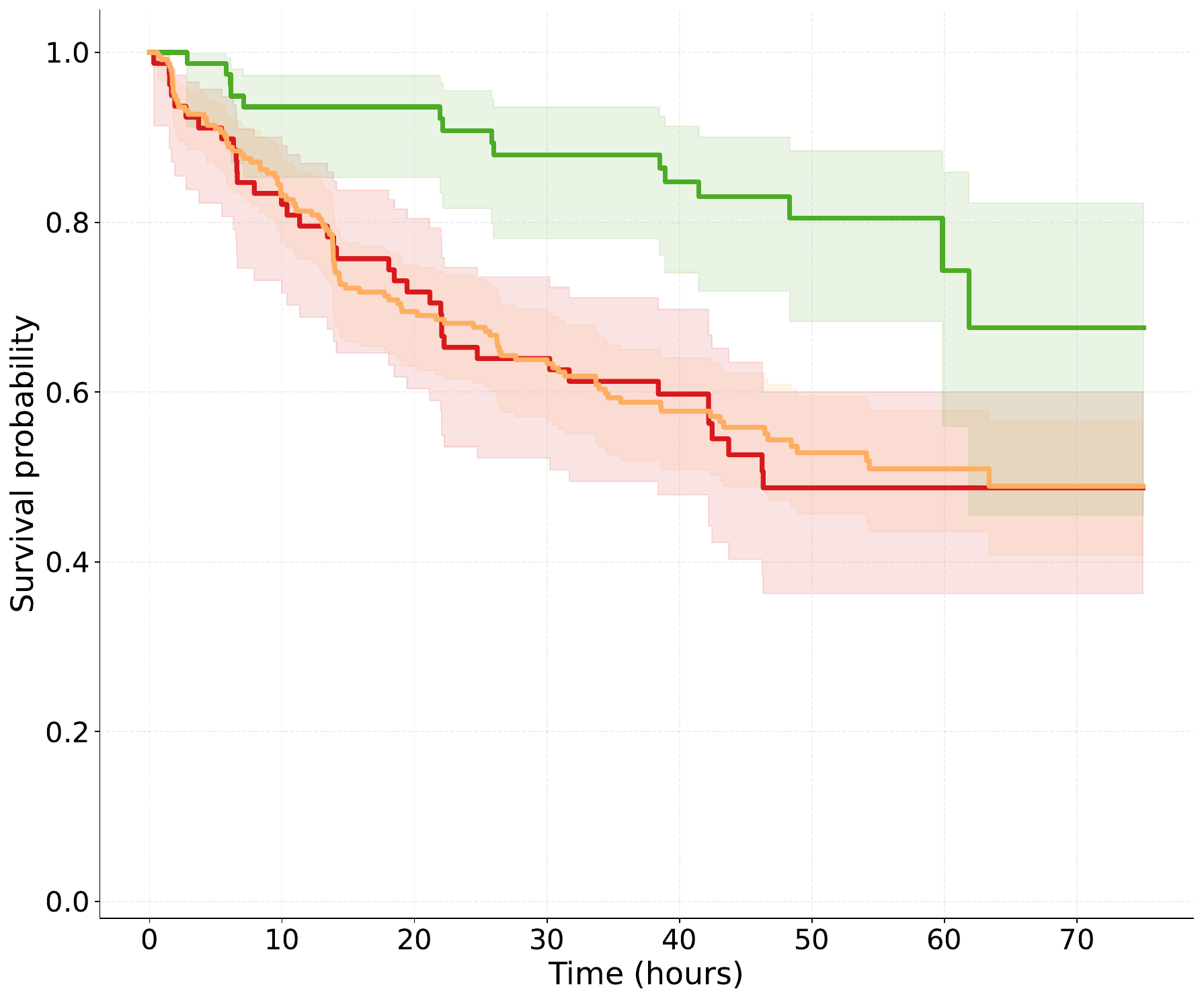}
\end{subfigure}
\hfill
\begin{subfigure}[t]{0.32\linewidth}
    \includegraphics[width=\linewidth]{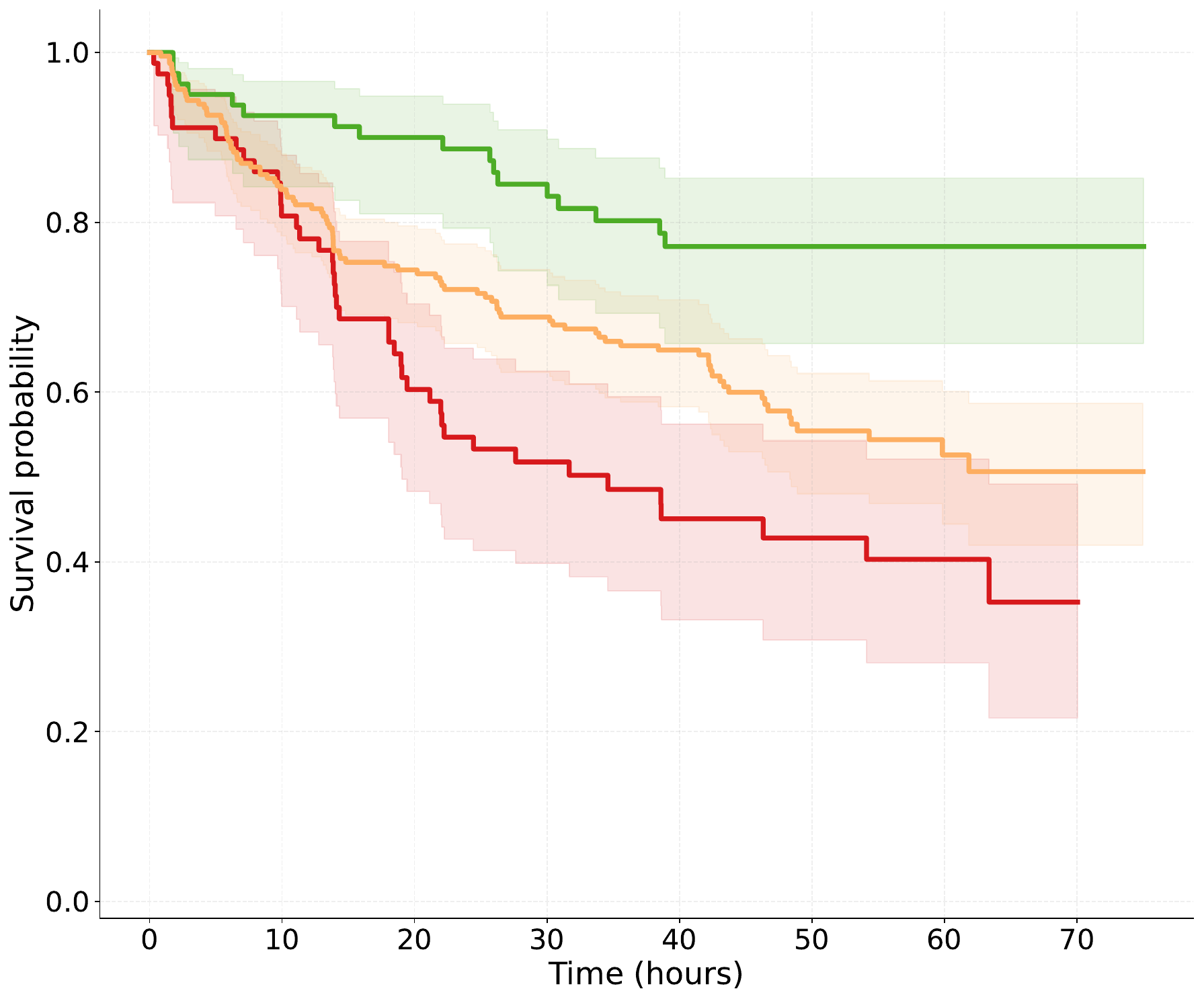}
\end{subfigure}
\hfill
\begin{subfigure}[t]{0.32\linewidth}
    \includegraphics[width=\linewidth]{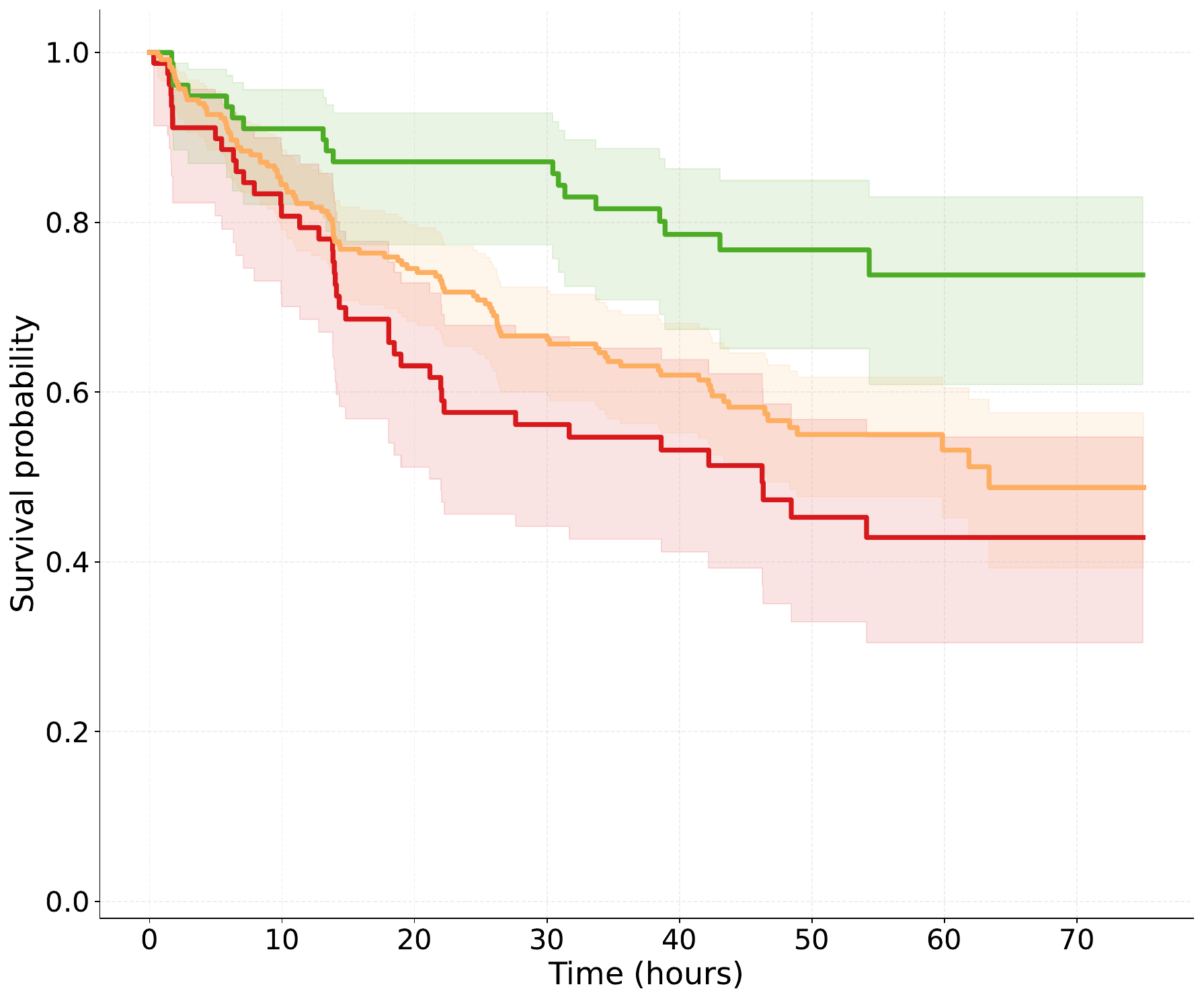}
\end{subfigure}
\vspace{0.3em}
\textit{(b) Classification fine-tuning} \\
\vspace{0.6em}
\begin{subfigure}[t]{0.32\linewidth}
    \includegraphics[width=\linewidth]{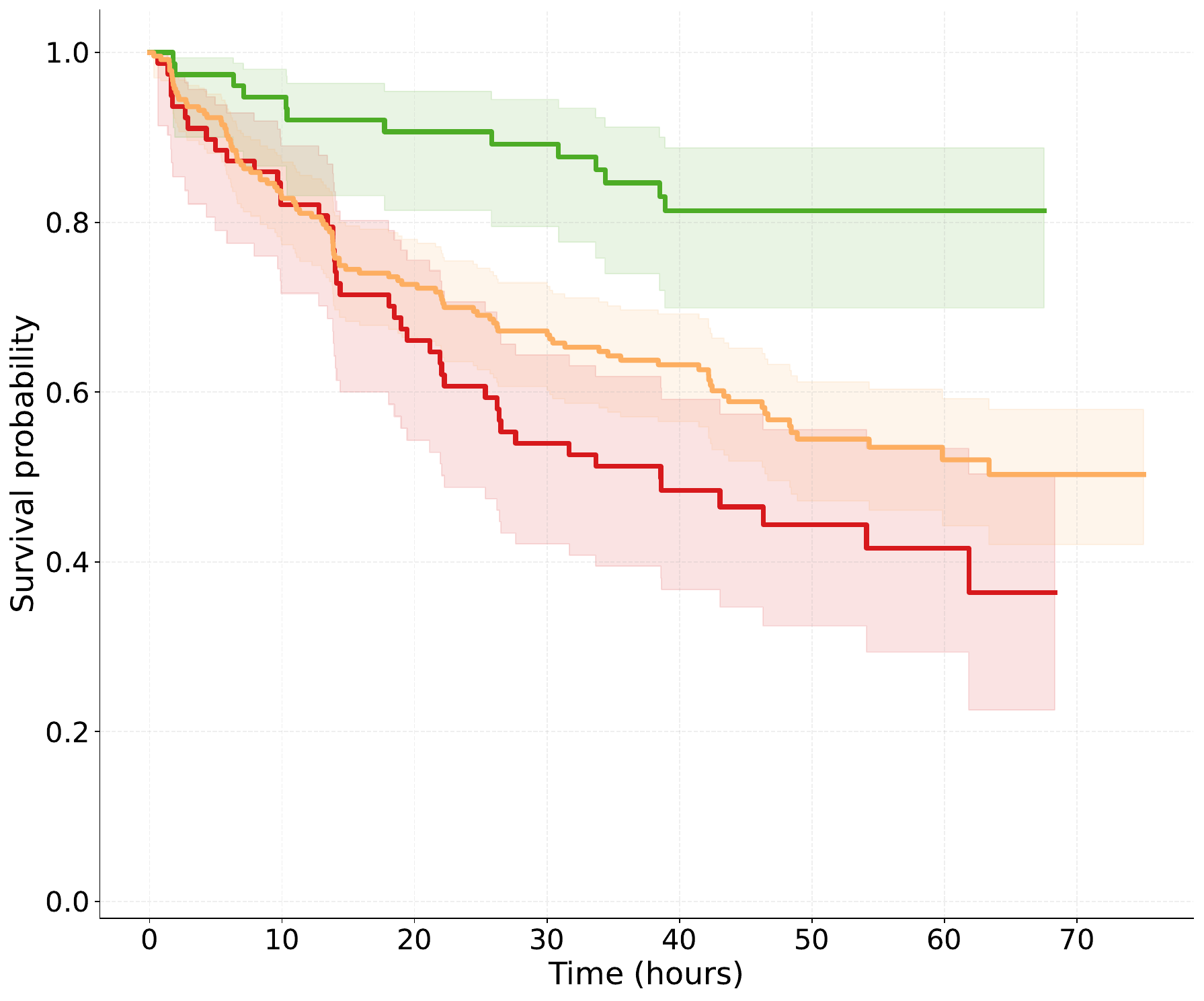}
\end{subfigure}
\hfill
\begin{subfigure}[t]{0.32\linewidth}
    \includegraphics[width=\linewidth]{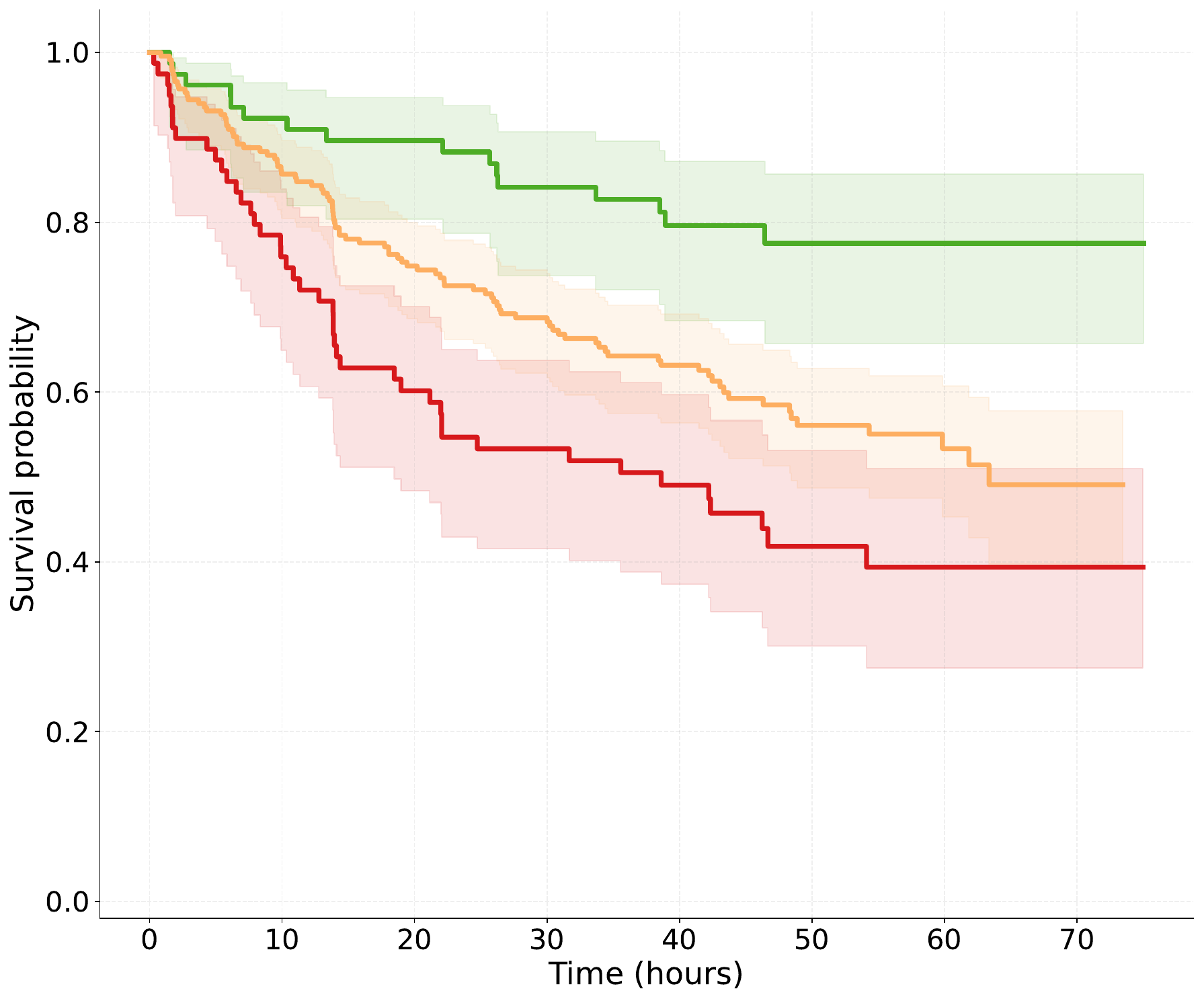}
\end{subfigure}
\hfill
\begin{subfigure}[t]{0.32\linewidth}
    \includegraphics[width=\linewidth]{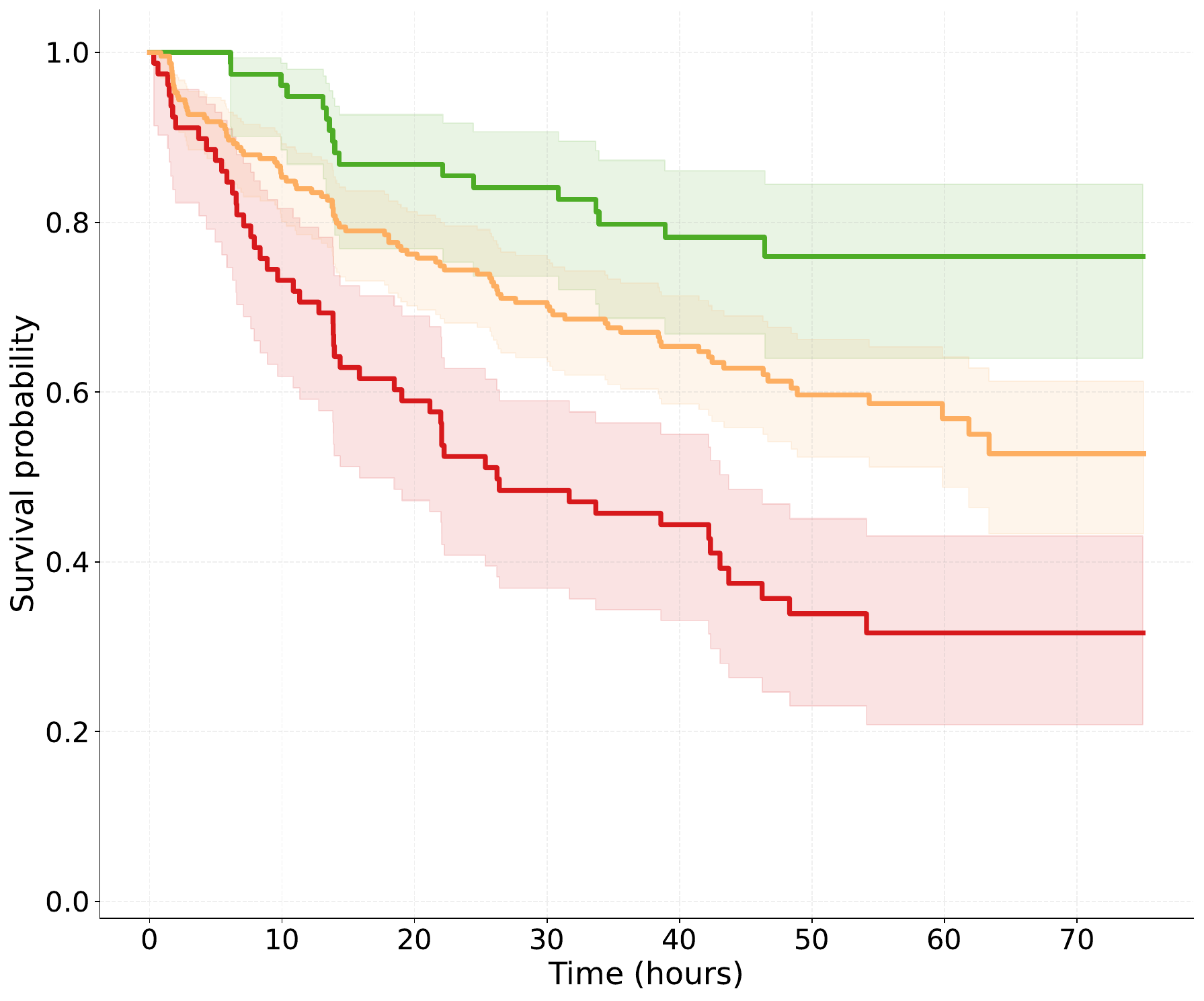}
\end{subfigure}
\vspace{0.3em}
\textit{(c) Survival head adaptation (MTLR; frozen backbone)}
\caption{\textbf{Risk-stratified Kaplan--Meier curves on the DIABETES data set.} }
\label{fig:risk_stratification_DIABETES}
\end{figure}

The aggregate benchmark evaluates ranking and probabilistic accuracy, but it does not show how those differences appear when predictions are used to form risk groups. Figures~\ref{fig:risk_stratification_COLON} and~\ref{fig:risk_stratification_DIABETES} therefore provide a qualitative view on two data sets from different size regimes: COLON ($N=911$, medium) and DIABETES ($N=394$, small). Across these examples, the adaptation strategies differ not only in scalar performance but also in the degree to which predicted risk tiers translate into ordered Kaplan--Meier curves. Survival-head variants generally produce clearer separation, while zero-shot and classification-based predictions show more overlap in some panels. These examples are intended to illustrate the practical manifestation of the benchmark-wide discrimination results, not to establish a separate ranking of interfaces or to imply that the same degree of separation holds on every data set.

\clearpage
\section{Full Results}
\begingroup
\footnotesize
\setlength{\tabcolsep}{4pt}
\begin{longtable}{lccclccc}
\caption[Core survival metrics]{Detailed Single-risk datasets Performance Metrics (mean~$\pm$~std across 5 folds). $C_{td}$: Antolini time-dependent concordance; AUC: mean dynamic AUC across q25/q50/q75. IBS: Integrated Brier Score.}\label{tab:core_metrics}\\
\toprule
\multicolumn{4}{c}{\textit{AML-BULL}} & \multicolumn{4}{c}{\textit{AIDS2}} \\
\midrule
Model & $C_{td}$ $\uparrow$ & IBS $\downarrow$ & AUC $\uparrow$ & Model & $C_{td}$ $\uparrow$ & IBS $\downarrow$ & AUC $\uparrow$ \\
\midrule
Cox PH & \ensuremath{0.549_{\textcolor{gray}{\pm 0.078}}} & \ensuremath{0.242_{\textcolor{gray}{\pm 0.077}}} & \ensuremath{0.519_{\textcolor{gray}{\pm 0.132}}} & Cox PH & \ensuremath{0.547_{\textcolor{gray}{\pm 0.019}}} & \ensuremath{0.138_{\textcolor{gray}{\pm 0.015}}} & \ensuremath{0.579_{\textcolor{gray}{\pm 0.025}}} \\
RSF & \ensuremath{\textbf{0.602}_{\textcolor{gray}{\pm 0.100}}} & \ensuremath{\textbf{0.184}_{\textcolor{gray}{\pm 0.046}}} & \ensuremath{\textbf{0.642}_{\textcolor{gray}{\pm 0.134}}} & RSF & \ensuremath{0.542_{\textcolor{gray}{\pm 0.004}}} & \ensuremath{\textbf{0.136}_{\textcolor{gray}{\pm 0.015}}} & \ensuremath{\underline{0.590}_{\textcolor{gray}{\pm 0.019}}} \\
GBSA & \ensuremath{\underline{0.600}_{\textcolor{gray}{\pm 0.121}}} & \ensuremath{0.210_{\textcolor{gray}{\pm 0.078}}} & \ensuremath{0.593_{\textcolor{gray}{\pm 0.167}}} & GBSA & \ensuremath{0.458_{\textcolor{gray}{\pm 0.053}}} & \ensuremath{0.138_{\textcolor{gray}{\pm 0.014}}} & \ensuremath{0.558_{\textcolor{gray}{\pm 0.021}}} \\
DeepSurv & \ensuremath{0.568_{\textcolor{gray}{\pm 0.044}}} & \ensuremath{0.206_{\textcolor{gray}{\pm 0.044}}} & \ensuremath{0.572_{\textcolor{gray}{\pm 0.078}}} & DeepSurv & \ensuremath{0.550_{\textcolor{gray}{\pm 0.012}}} & \ensuremath{0.139_{\textcolor{gray}{\pm 0.013}}} & \ensuremath{0.582_{\textcolor{gray}{\pm 0.017}}} \\
\bottomrule
\end{longtable}
\endgroup
\begingroup
\footnotesize
\setlength{\tabcolsep}{3pt}
\begin{longtable}{llcccccc}
\caption{\textbf{Competing-risk results per cause} (mean~$\pm$~std across 5 folds). Superscript $(1)$ and $(2)$ index the two competing causes. Best value per column within a data set is \textbf{bold}, second best \underline{underlined}.}\label{tab:detailed_metrics_cr}\\
\toprule
\textbf{Dataset} & \textbf{Model} & \textbf{$C_{td}^{(1)}$} $\uparrow$ & \textbf{IBS$^{(1)}$} $\downarrow$ & \textbf{AUC$^{(1)}$} $\uparrow$ & \textbf{$C_{td}^{(2)}$} $\uparrow$ & \textbf{IBS$^{(2)}$} $\downarrow$ & \textbf{AUC$^{(2)}$} $\uparrow$ \\
\midrule
\endfirsthead
\toprule
\textbf{Dataset} & \textbf{Model} & \textbf{$C_{td}^{(1)}$} $\uparrow$ & \textbf{IBS$^{(1)}$} $\downarrow$ & \textbf{AUC$^{(1)}$} $\uparrow$ & \textbf{$C_{td}^{(2)}$} $\uparrow$ & \textbf{IBS$^{(2)}$} $\downarrow$ & \textbf{AUC$^{(2)}$} $\uparrow$ \\
\midrule
\endhead
\bottomrule
\endfoot
\textbf{Fram-CR} & Cox PH & \ensuremath{\textbf{0.744}_{\textcolor{gray}{\pm 0.009}}} & \ensuremath{0.177_{\textcolor{gray}{\pm 0.007}}} & \ensuremath{\underline{0.639}_{\textcolor{gray}{\pm 0.017}}} & \ensuremath{\textbf{0.713}_{\textcolor{gray}{\pm 0.020}}} & \ensuremath{0.146_{\textcolor{gray}{\pm 0.005}}} & \ensuremath{0.592_{\textcolor{gray}{\pm 0.042}}} \\
 & SurvBoost & \ensuremath{0.730_{\textcolor{gray}{\pm 0.004}}} & \ensuremath{0.180_{\textcolor{gray}{\pm 0.007}}} & \ensuremath{0.623_{\textcolor{gray}{\pm 0.025}}} & \ensuremath{0.701_{\textcolor{gray}{\pm 0.018}}} & \ensuremath{0.146_{\textcolor{gray}{\pm 0.005}}} & \ensuremath{0.591_{\textcolor{gray}{\pm 0.046}}} \\
 & DeepSurv & \ensuremath{0.691_{\textcolor{gray}{\pm 0.057}}} & \ensuremath{0.185_{\textcolor{gray}{\pm 0.010}}} & \ensuremath{0.606_{\textcolor{gray}{\pm 0.030}}} & \ensuremath{0.568_{\textcolor{gray}{\pm 0.138}}} & \ensuremath{0.159_{\textcolor{gray}{\pm 0.010}}} & \ensuremath{0.546_{\textcolor{gray}{\pm 0.068}}} \\
 & MLP-DH & \ensuremath{0.718_{\textcolor{gray}{\pm 0.012}}} & \ensuremath{0.188_{\textcolor{gray}{\pm 0.008}}} & \ensuremath{0.621_{\textcolor{gray}{\pm 0.010}}} & \ensuremath{0.673_{\textcolor{gray}{\pm 0.019}}} & \ensuremath{0.152_{\textcolor{gray}{\pm 0.005}}} & \ensuremath{\textbf{0.597}_{\textcolor{gray}{\pm 0.040}}} \\
 & DySurv & \ensuremath{0.720_{\textcolor{gray}{\pm 0.010}}} & \ensuremath{0.175_{\textcolor{gray}{\pm 0.011}}} & \ensuremath{0.608_{\textcolor{gray}{\pm 0.026}}} & \ensuremath{0.679_{\textcolor{gray}{\pm 0.029}}} & \ensuremath{0.147_{\textcolor{gray}{\pm 0.005}}} & \ensuremath{0.537_{\textcolor{gray}{\pm 0.050}}} \\
 & SurvTrace & \ensuremath{0.728_{\textcolor{gray}{\pm 0.005}}} & \ensuremath{0.176_{\textcolor{gray}{\pm 0.010}}} & \ensuremath{0.615_{\textcolor{gray}{\pm 0.017}}} & \ensuremath{0.702_{\textcolor{gray}{\pm 0.027}}} & \ensuremath{0.144_{\textcolor{gray}{\pm 0.005}}} & \ensuremath{0.569_{\textcolor{gray}{\pm 0.045}}} \\
 & TabPFN-DH & \ensuremath{0.713_{\textcolor{gray}{\pm 0.007}}} & \ensuremath{0.181_{\textcolor{gray}{\pm 0.005}}} & \ensuremath{0.611_{\textcolor{gray}{\pm 0.019}}} & \ensuremath{0.697_{\textcolor{gray}{\pm 0.017}}} & \ensuremath{0.148_{\textcolor{gray}{\pm 0.005}}} & \ensuremath{0.575_{\textcolor{gray}{\pm 0.030}}} \\
 & TabDPT-DH & \ensuremath{0.726_{\textcolor{gray}{\pm 0.005}}} & \ensuremath{0.182_{\textcolor{gray}{\pm 0.004}}} & \ensuremath{0.612_{\textcolor{gray}{\pm 0.011}}} & \ensuremath{0.709_{\textcolor{gray}{\pm 0.025}}} & \ensuremath{0.150_{\textcolor{gray}{\pm 0.006}}} & \ensuremath{0.574_{\textcolor{gray}{\pm 0.049}}} \\
 & TabICL-DH & \ensuremath{0.719_{\textcolor{gray}{\pm 0.014}}} & \ensuremath{0.185_{\textcolor{gray}{\pm 0.007}}} & \ensuremath{0.596_{\textcolor{gray}{\pm 0.017}}} & \ensuremath{0.706_{\textcolor{gray}{\pm 0.032}}} & \ensuremath{0.150_{\textcolor{gray}{\pm 0.003}}} & \ensuremath{0.566_{\textcolor{gray}{\pm 0.052}}} \\
 & TabPFN-MTLR & \ensuremath{0.728_{\textcolor{gray}{\pm 0.008}}} & \ensuremath{0.165_{\textcolor{gray}{\pm 0.007}}} & \ensuremath{0.634_{\textcolor{gray}{\pm 0.018}}} & \ensuremath{0.701_{\textcolor{gray}{\pm 0.026}}} & \ensuremath{0.135_{\textcolor{gray}{\pm 0.005}}} & \ensuremath{\underline{0.594}_{\textcolor{gray}{\pm 0.045}}} \\
 & TabDPT-MTLR & \ensuremath{\underline{0.732}_{\textcolor{gray}{\pm 0.010}}} & \ensuremath{\underline{0.163}_{\textcolor{gray}{\pm 0.009}}} & \ensuremath{0.627_{\textcolor{gray}{\pm 0.022}}} & \ensuremath{\underline{0.710}_{\textcolor{gray}{\pm 0.014}}} & \ensuremath{\textbf{0.133}_{\textcolor{gray}{\pm 0.005}}} & \ensuremath{0.590_{\textcolor{gray}{\pm 0.043}}} \\
 & TabICL-MTLR & \ensuremath{0.729_{\textcolor{gray}{\pm 0.010}}} & \ensuremath{\textbf{0.162}_{\textcolor{gray}{\pm 0.007}}} & \ensuremath{0.631_{\textcolor{gray}{\pm 0.021}}} & \ensuremath{0.706_{\textcolor{gray}{\pm 0.023}}} & \ensuremath{0.134_{\textcolor{gray}{\pm 0.003}}} & \ensuremath{0.589_{\textcolor{gray}{\pm 0.041}}} \\
 & TabPFN-Cox & \ensuremath{0.726_{\textcolor{gray}{\pm 0.014}}} & \ensuremath{0.166_{\textcolor{gray}{\pm 0.007}}} & \ensuremath{\textbf{0.639}_{\textcolor{gray}{\pm 0.008}}} & \ensuremath{0.666_{\textcolor{gray}{\pm 0.013}}} & \ensuremath{\underline{0.133}_{\textcolor{gray}{\pm 0.007}}} & \ensuremath{0.588_{\textcolor{gray}{\pm 0.047}}} \\
 & TabDPT-Cox & \ensuremath{0.701_{\textcolor{gray}{\pm 0.022}}} & \ensuremath{0.198_{\textcolor{gray}{\pm 0.058}}} & \ensuremath{0.622_{\textcolor{gray}{\pm 0.015}}} & \ensuremath{0.625_{\textcolor{gray}{\pm 0.048}}} & \ensuremath{0.146_{\textcolor{gray}{\pm 0.015}}} & \ensuremath{0.594_{\textcolor{gray}{\pm 0.029}}} \\
 & TabICL-Cox & \ensuremath{0.712_{\textcolor{gray}{\pm 0.009}}} & \ensuremath{0.203_{\textcolor{gray}{\pm 0.079}}} & \ensuremath{0.622_{\textcolor{gray}{\pm 0.020}}} & \ensuremath{0.631_{\textcolor{gray}{\pm 0.046}}} & \ensuremath{0.147_{\textcolor{gray}{\pm 0.016}}} & \ensuremath{0.573_{\textcolor{gray}{\pm 0.038}}} \\
\midrule
\textbf{PBC2-CR} & Cox PH & \ensuremath{0.790_{\textcolor{gray}{\pm 0.008}}} & \ensuremath{0.127_{\textcolor{gray}{\pm 0.010}}} & \ensuremath{0.829_{\textcolor{gray}{\pm 0.015}}} & \ensuremath{0.830_{\textcolor{gray}{\pm 0.047}}} & \ensuremath{0.034_{\textcolor{gray}{\pm 0.001}}} & \ensuremath{0.839_{\textcolor{gray}{\pm 0.042}}} \\
 & SurvBoost & \ensuremath{0.837_{\textcolor{gray}{\pm 0.014}}} & \ensuremath{0.102_{\textcolor{gray}{\pm 0.005}}} & \ensuremath{\underline{0.869}_{\textcolor{gray}{\pm 0.016}}} & \ensuremath{0.872_{\textcolor{gray}{\pm 0.032}}} & \ensuremath{\underline{0.028}_{\textcolor{gray}{\pm 0.001}}} & \ensuremath{\textbf{0.886}_{\textcolor{gray}{\pm 0.031}}} \\
 & DeepSurv & \ensuremath{0.815_{\textcolor{gray}{\pm 0.013}}} & \ensuremath{0.117_{\textcolor{gray}{\pm 0.011}}} & \ensuremath{0.851_{\textcolor{gray}{\pm 0.013}}} & \ensuremath{0.867_{\textcolor{gray}{\pm 0.023}}} & \ensuremath{0.030_{\textcolor{gray}{\pm 0.002}}} & \ensuremath{0.863_{\textcolor{gray}{\pm 0.031}}} \\
 & MLP-DH & \ensuremath{0.802_{\textcolor{gray}{\pm 0.026}}} & \ensuremath{\textbf{0.098}_{\textcolor{gray}{\pm 0.009}}} & \ensuremath{0.863_{\textcolor{gray}{\pm 0.032}}} & \ensuremath{0.801_{\textcolor{gray}{\pm 0.057}}} & \ensuremath{\textbf{0.026}_{\textcolor{gray}{\pm 0.002}}} & \ensuremath{\underline{0.884}_{\textcolor{gray}{\pm 0.044}}} \\
 & DySurv & \ensuremath{0.765_{\textcolor{gray}{\pm 0.010}}} & \ensuremath{0.142_{\textcolor{gray}{\pm 0.009}}} & \ensuremath{0.811_{\textcolor{gray}{\pm 0.011}}} & \ensuremath{0.677_{\textcolor{gray}{\pm 0.044}}} & \ensuremath{0.041_{\textcolor{gray}{\pm 0.007}}} & \ensuremath{0.708_{\textcolor{gray}{\pm 0.076}}} \\
 & SurvTrace & \ensuremath{0.801_{\textcolor{gray}{\pm 0.015}}} & \ensuremath{0.109_{\textcolor{gray}{\pm 0.009}}} & \ensuremath{0.849_{\textcolor{gray}{\pm 0.022}}} & \ensuremath{0.840_{\textcolor{gray}{\pm 0.019}}} & \ensuremath{0.030_{\textcolor{gray}{\pm 0.003}}} & \ensuremath{0.856_{\textcolor{gray}{\pm 0.028}}} \\
 & TabPFN-DH & \ensuremath{0.815_{\textcolor{gray}{\pm 0.013}}} & \ensuremath{0.108_{\textcolor{gray}{\pm 0.014}}} & \ensuremath{0.852_{\textcolor{gray}{\pm 0.020}}} & \ensuremath{0.866_{\textcolor{gray}{\pm 0.034}}} & \ensuremath{0.030_{\textcolor{gray}{\pm 0.002}}} & \ensuremath{0.847_{\textcolor{gray}{\pm 0.031}}} \\
 & TabDPT-DH & \ensuremath{\textbf{0.846}_{\textcolor{gray}{\pm 0.016}}} & \ensuremath{\underline{0.098}_{\textcolor{gray}{\pm 0.010}}} & \ensuremath{\textbf{0.874}_{\textcolor{gray}{\pm 0.014}}} & \ensuremath{\underline{0.880}_{\textcolor{gray}{\pm 0.036}}} & \ensuremath{0.031_{\textcolor{gray}{\pm 0.002}}} & \ensuremath{0.869_{\textcolor{gray}{\pm 0.020}}} \\
 & TabICL-DH & \ensuremath{0.828_{\textcolor{gray}{\pm 0.021}}} & \ensuremath{0.109_{\textcolor{gray}{\pm 0.012}}} & \ensuremath{0.858_{\textcolor{gray}{\pm 0.019}}} & \ensuremath{0.816_{\textcolor{gray}{\pm 0.058}}} & \ensuremath{0.033_{\textcolor{gray}{\pm 0.003}}} & \ensuremath{0.825_{\textcolor{gray}{\pm 0.039}}} \\
 & TabPFN-MTLR & \ensuremath{0.807_{\textcolor{gray}{\pm 0.012}}} & \ensuremath{0.123_{\textcolor{gray}{\pm 0.012}}} & \ensuremath{0.845_{\textcolor{gray}{\pm 0.013}}} & \ensuremath{0.844_{\textcolor{gray}{\pm 0.020}}} & \ensuremath{0.035_{\textcolor{gray}{\pm 0.002}}} & \ensuremath{0.850_{\textcolor{gray}{\pm 0.031}}} \\
 & TabDPT-MTLR & \ensuremath{\underline{0.837}_{\textcolor{gray}{\pm 0.011}}} & \ensuremath{0.108_{\textcolor{gray}{\pm 0.012}}} & \ensuremath{0.862_{\textcolor{gray}{\pm 0.008}}} & \ensuremath{0.867_{\textcolor{gray}{\pm 0.042}}} & \ensuremath{0.034_{\textcolor{gray}{\pm 0.005}}} & \ensuremath{0.873_{\textcolor{gray}{\pm 0.039}}} \\
 & TabICL-MTLR & \ensuremath{0.833_{\textcolor{gray}{\pm 0.013}}} & \ensuremath{0.111_{\textcolor{gray}{\pm 0.013}}} & \ensuremath{0.858_{\textcolor{gray}{\pm 0.011}}} & \ensuremath{\textbf{0.887}_{\textcolor{gray}{\pm 0.024}}} & \ensuremath{0.033_{\textcolor{gray}{\pm 0.001}}} & \ensuremath{0.879_{\textcolor{gray}{\pm 0.036}}} \\
 & TabPFN-Cox & \ensuremath{0.735_{\textcolor{gray}{\pm 0.047}}} & \ensuremath{0.399_{\textcolor{gray}{\pm 0.058}}} & \ensuremath{0.821_{\textcolor{gray}{\pm 0.018}}} & \ensuremath{0.781_{\textcolor{gray}{\pm 0.038}}} & \ensuremath{0.099_{\textcolor{gray}{\pm 0.023}}} & \ensuremath{0.760_{\textcolor{gray}{\pm 0.072}}} \\
 & TabDPT-Cox & \ensuremath{0.765_{\textcolor{gray}{\pm 0.119}}} & \ensuremath{0.298_{\textcolor{gray}{\pm 0.130}}} & \ensuremath{0.844_{\textcolor{gray}{\pm 0.013}}} & \ensuremath{0.771_{\textcolor{gray}{\pm 0.092}}} & \ensuremath{0.080_{\textcolor{gray}{\pm 0.028}}} & \ensuremath{0.714_{\textcolor{gray}{\pm 0.117}}} \\
 & TabICL-Cox & \ensuremath{0.781_{\textcolor{gray}{\pm 0.066}}} & \ensuremath{0.341_{\textcolor{gray}{\pm 0.068}}} & \ensuremath{0.860_{\textcolor{gray}{\pm 0.010}}} & \ensuremath{0.683_{\textcolor{gray}{\pm 0.080}}} & \ensuremath{0.065_{\textcolor{gray}{\pm 0.027}}} & \ensuremath{0.664_{\textcolor{gray}{\pm 0.084}}} \\
\midrule
\textbf{Supp-CR} & Cox PH & \ensuremath{0.776_{\textcolor{gray}{\pm 0.013}}} & \ensuremath{0.154_{\textcolor{gray}{\pm 0.005}}} & \ensuremath{0.795_{\textcolor{gray}{\pm 0.011}}} & \ensuremath{0.804_{\textcolor{gray}{\pm 0.005}}} & \ensuremath{0.175_{\textcolor{gray}{\pm 0.009}}} & \ensuremath{0.846_{\textcolor{gray}{\pm 0.009}}} \\
 & SurvBoost & \ensuremath{0.813_{\textcolor{gray}{\pm 0.008}}} & \ensuremath{0.131_{\textcolor{gray}{\pm 0.006}}} & \ensuremath{0.834_{\textcolor{gray}{\pm 0.015}}} & \ensuremath{0.815_{\textcolor{gray}{\pm 0.006}}} & \ensuremath{\underline{0.170}_{\textcolor{gray}{\pm 0.008}}} & \ensuremath{\textbf{0.855}_{\textcolor{gray}{\pm 0.007}}} \\
 & DeepSurv & \ensuremath{\textbf{0.844}_{\textcolor{gray}{\pm 0.009}}} & \ensuremath{\textbf{0.120}_{\textcolor{gray}{\pm 0.009}}} & \ensuremath{\textbf{0.861}_{\textcolor{gray}{\pm 0.010}}} & \ensuremath{0.813_{\textcolor{gray}{\pm 0.005}}} & \ensuremath{\textbf{0.170}_{\textcolor{gray}{\pm 0.008}}} & \ensuremath{0.850_{\textcolor{gray}{\pm 0.006}}} \\
 & MLP-DH & \ensuremath{0.821_{\textcolor{gray}{\pm 0.017}}} & \ensuremath{0.129_{\textcolor{gray}{\pm 0.003}}} & \ensuremath{0.827_{\textcolor{gray}{\pm 0.015}}} & \ensuremath{0.745_{\textcolor{gray}{\pm 0.007}}} & \ensuremath{0.202_{\textcolor{gray}{\pm 0.007}}} & \ensuremath{0.798_{\textcolor{gray}{\pm 0.010}}} \\
 & DySurv & \ensuremath{0.480_{\textcolor{gray}{\pm 0.015}}} & \ensuremath{0.173_{\textcolor{gray}{\pm 0.005}}} & \ensuremath{0.759_{\textcolor{gray}{\pm 0.019}}} & \ensuremath{0.357_{\textcolor{gray}{\pm 0.013}}} & \ensuremath{0.201_{\textcolor{gray}{\pm 0.006}}} & \ensuremath{0.803_{\textcolor{gray}{\pm 0.013}}} \\
 & SurvTrace & \ensuremath{0.565_{\textcolor{gray}{\pm 0.026}}} & \ensuremath{0.137_{\textcolor{gray}{\pm 0.006}}} & \ensuremath{0.805_{\textcolor{gray}{\pm 0.008}}} & \ensuremath{0.363_{\textcolor{gray}{\pm 0.010}}} & \ensuremath{0.179_{\textcolor{gray}{\pm 0.007}}} & \ensuremath{0.834_{\textcolor{gray}{\pm 0.007}}} \\
 & TabPFN-DH & \ensuremath{0.816_{\textcolor{gray}{\pm 0.016}}} & \ensuremath{0.139_{\textcolor{gray}{\pm 0.005}}} & \ensuremath{0.782_{\textcolor{gray}{\pm 0.014}}} & \ensuremath{0.804_{\textcolor{gray}{\pm 0.004}}} & \ensuremath{0.172_{\textcolor{gray}{\pm 0.007}}} & \ensuremath{0.836_{\textcolor{gray}{\pm 0.008}}} \\
 & TabDPT-DH & \ensuremath{0.805_{\textcolor{gray}{\pm 0.023}}} & \ensuremath{0.144_{\textcolor{gray}{\pm 0.013}}} & \ensuremath{0.777_{\textcolor{gray}{\pm 0.026}}} & \ensuremath{0.804_{\textcolor{gray}{\pm 0.005}}} & \ensuremath{0.173_{\textcolor{gray}{\pm 0.007}}} & \ensuremath{0.841_{\textcolor{gray}{\pm 0.011}}} \\
 & TabICL-DH & \ensuremath{0.825_{\textcolor{gray}{\pm 0.020}}} & \ensuremath{0.136_{\textcolor{gray}{\pm 0.013}}} & \ensuremath{0.797_{\textcolor{gray}{\pm 0.018}}} & \ensuremath{0.800_{\textcolor{gray}{\pm 0.008}}} & \ensuremath{0.176_{\textcolor{gray}{\pm 0.008}}} & \ensuremath{0.839_{\textcolor{gray}{\pm 0.010}}} \\
 & TabPFN-MTLR & \ensuremath{\underline{0.840}_{\textcolor{gray}{\pm 0.010}}} & \ensuremath{0.131_{\textcolor{gray}{\pm 0.007}}} & \ensuremath{0.841_{\textcolor{gray}{\pm 0.012}}} & \ensuremath{0.816_{\textcolor{gray}{\pm 0.003}}} & \ensuremath{0.171_{\textcolor{gray}{\pm 0.009}}} & \ensuremath{0.850_{\textcolor{gray}{\pm 0.008}}} \\
 & TabDPT-MTLR & \ensuremath{0.835_{\textcolor{gray}{\pm 0.017}}} & \ensuremath{0.131_{\textcolor{gray}{\pm 0.006}}} & \ensuremath{\underline{0.843}_{\textcolor{gray}{\pm 0.011}}} & \ensuremath{\textbf{0.821}_{\textcolor{gray}{\pm 0.007}}} & \ensuremath{0.172_{\textcolor{gray}{\pm 0.011}}} & \ensuremath{\underline{0.853}_{\textcolor{gray}{\pm 0.008}}} \\
 & TabICL-MTLR & \ensuremath{0.835_{\textcolor{gray}{\pm 0.018}}} & \ensuremath{0.136_{\textcolor{gray}{\pm 0.006}}} & \ensuremath{0.840_{\textcolor{gray}{\pm 0.025}}} & \ensuremath{\underline{0.819}_{\textcolor{gray}{\pm 0.005}}} & \ensuremath{0.172_{\textcolor{gray}{\pm 0.010}}} & \ensuremath{0.850_{\textcolor{gray}{\pm 0.009}}} \\
 & TabPFN-Cox & \ensuremath{0.783_{\textcolor{gray}{\pm 0.020}}} & \ensuremath{0.133_{\textcolor{gray}{\pm 0.006}}} & \ensuremath{0.816_{\textcolor{gray}{\pm 0.013}}} & \ensuremath{0.715_{\textcolor{gray}{\pm 0.005}}} & \ensuremath{0.175_{\textcolor{gray}{\pm 0.008}}} & \ensuremath{0.841_{\textcolor{gray}{\pm 0.005}}} \\
 & TabDPT-Cox & \ensuremath{0.778_{\textcolor{gray}{\pm 0.033}}} & \ensuremath{0.133_{\textcolor{gray}{\pm 0.013}}} & \ensuremath{0.813_{\textcolor{gray}{\pm 0.031}}} & \ensuremath{0.716_{\textcolor{gray}{\pm 0.008}}} & \ensuremath{0.177_{\textcolor{gray}{\pm 0.007}}} & \ensuremath{0.840_{\textcolor{gray}{\pm 0.005}}} \\
 & TabICL-Cox & \ensuremath{0.783_{\textcolor{gray}{\pm 0.033}}} & \ensuremath{\underline{0.127}_{\textcolor{gray}{\pm 0.015}}} & \ensuremath{0.821_{\textcolor{gray}{\pm 0.029}}} & \ensuremath{0.708_{\textcolor{gray}{\pm 0.009}}} & \ensuremath{0.183_{\textcolor{gray}{\pm 0.009}}} & \ensuremath{0.836_{\textcolor{gray}{\pm 0.005}}} \\
\midrule
\textbf{Syn-CR} & Cox PH & \ensuremath{0.580_{\textcolor{gray}{\pm 0.007}}} & \ensuremath{0.188_{\textcolor{gray}{\pm 0.001}}} & \ensuremath{0.573_{\textcolor{gray}{\pm 0.007}}} & \ensuremath{0.590_{\textcolor{gray}{\pm 0.008}}} & \ensuremath{0.189_{\textcolor{gray}{\pm 0.004}}} & \ensuremath{0.577_{\textcolor{gray}{\pm 0.011}}} \\
 & SurvBoost & \ensuremath{0.686_{\textcolor{gray}{\pm 0.009}}} & \ensuremath{0.169_{\textcolor{gray}{\pm 0.003}}} & \ensuremath{0.744_{\textcolor{gray}{\pm 0.014}}} & \ensuremath{0.683_{\textcolor{gray}{\pm 0.014}}} & \ensuremath{0.173_{\textcolor{gray}{\pm 0.004}}} & \ensuremath{0.737_{\textcolor{gray}{\pm 0.014}}} \\
 & DeepSurv & \ensuremath{\textbf{0.747}_{\textcolor{gray}{\pm 0.007}}} & \ensuremath{\textbf{0.150}_{\textcolor{gray}{\pm 0.004}}} & \ensuremath{\textbf{0.807}_{\textcolor{gray}{\pm 0.009}}} & \ensuremath{\textbf{0.749}_{\textcolor{gray}{\pm 0.005}}} & \ensuremath{\textbf{0.152}_{\textcolor{gray}{\pm 0.003}}} & \ensuremath{\textbf{0.805}_{\textcolor{gray}{\pm 0.006}}} \\
 & MLP-DH & \ensuremath{\underline{0.739}_{\textcolor{gray}{\pm 0.005}}} & \ensuremath{0.239_{\textcolor{gray}{\pm 0.005}}} & \ensuremath{0.750_{\textcolor{gray}{\pm 0.019}}} & \ensuremath{\underline{0.743}_{\textcolor{gray}{\pm 0.005}}} & \ensuremath{0.243_{\textcolor{gray}{\pm 0.005}}} & \ensuremath{0.752_{\textcolor{gray}{\pm 0.012}}} \\
 & DySurv & \ensuremath{0.504_{\textcolor{gray}{\pm 0.012}}} & \ensuremath{0.267_{\textcolor{gray}{\pm 0.007}}} & \ensuremath{\underline{0.804}_{\textcolor{gray}{\pm 0.011}}} & \ensuremath{0.498_{\textcolor{gray}{\pm 0.012}}} & \ensuremath{0.268_{\textcolor{gray}{\pm 0.007}}} & \ensuremath{\underline{0.803}_{\textcolor{gray}{\pm 0.005}}} \\
 & SurvTrace & \ensuremath{0.496_{\textcolor{gray}{\pm 0.013}}} & \ensuremath{0.253_{\textcolor{gray}{\pm 0.009}}} & \ensuremath{0.733_{\textcolor{gray}{\pm 0.014}}} & \ensuremath{0.491_{\textcolor{gray}{\pm 0.014}}} & \ensuremath{0.255_{\textcolor{gray}{\pm 0.014}}} & \ensuremath{0.720_{\textcolor{gray}{\pm 0.016}}} \\
 & TabPFN-DH & \ensuremath{0.608_{\textcolor{gray}{\pm 0.004}}} & \ensuremath{0.239_{\textcolor{gray}{\pm 0.004}}} & \ensuremath{0.660_{\textcolor{gray}{\pm 0.012}}} & \ensuremath{0.610_{\textcolor{gray}{\pm 0.011}}} & \ensuremath{0.244_{\textcolor{gray}{\pm 0.008}}} & \ensuremath{0.673_{\textcolor{gray}{\pm 0.017}}} \\
 & TabDPT-DH & \ensuremath{0.538_{\textcolor{gray}{\pm 0.009}}} & \ensuremath{0.250_{\textcolor{gray}{\pm 0.005}}} & \ensuremath{0.603_{\textcolor{gray}{\pm 0.005}}} & \ensuremath{0.540_{\textcolor{gray}{\pm 0.018}}} & \ensuremath{0.251_{\textcolor{gray}{\pm 0.009}}} & \ensuremath{0.602_{\textcolor{gray}{\pm 0.024}}} \\
 & TabICL-DH & \ensuremath{0.504_{\textcolor{gray}{\pm 0.007}}} & \ensuremath{0.251_{\textcolor{gray}{\pm 0.004}}} & \ensuremath{0.575_{\textcolor{gray}{\pm 0.007}}} & \ensuremath{0.513_{\textcolor{gray}{\pm 0.011}}} & \ensuremath{0.254_{\textcolor{gray}{\pm 0.006}}} & \ensuremath{0.583_{\textcolor{gray}{\pm 0.019}}} \\
 & TabPFN-MTLR & \ensuremath{0.714_{\textcolor{gray}{\pm 0.007}}} & \ensuremath{\underline{0.166}_{\textcolor{gray}{\pm 0.004}}} & \ensuremath{0.796_{\textcolor{gray}{\pm 0.007}}} & \ensuremath{0.717_{\textcolor{gray}{\pm 0.006}}} & \ensuremath{\underline{0.167}_{\textcolor{gray}{\pm 0.004}}} & \ensuremath{0.798_{\textcolor{gray}{\pm 0.007}}} \\
 & TabDPT-MTLR & \ensuremath{0.593_{\textcolor{gray}{\pm 0.011}}} & \ensuremath{0.197_{\textcolor{gray}{\pm 0.003}}} & \ensuremath{0.610_{\textcolor{gray}{\pm 0.016}}} & \ensuremath{0.600_{\textcolor{gray}{\pm 0.014}}} & \ensuremath{0.203_{\textcolor{gray}{\pm 0.006}}} & \ensuremath{0.608_{\textcolor{gray}{\pm 0.021}}} \\
 & TabICL-MTLR & \ensuremath{0.569_{\textcolor{gray}{\pm 0.013}}} & \ensuremath{0.200_{\textcolor{gray}{\pm 0.004}}} & \ensuremath{0.577_{\textcolor{gray}{\pm 0.010}}} & \ensuremath{0.579_{\textcolor{gray}{\pm 0.014}}} & \ensuremath{0.202_{\textcolor{gray}{\pm 0.003}}} & \ensuremath{0.598_{\textcolor{gray}{\pm 0.017}}} \\
 & TabPFN-Cox & \ensuremath{0.714_{\textcolor{gray}{\pm 0.006}}} & \ensuremath{0.227_{\textcolor{gray}{\pm 0.002}}} & \ensuremath{0.711_{\textcolor{gray}{\pm 0.020}}} & \ensuremath{0.716_{\textcolor{gray}{\pm 0.009}}} & \ensuremath{0.244_{\textcolor{gray}{\pm 0.025}}} & \ensuremath{0.701_{\textcolor{gray}{\pm 0.025}}} \\
 & TabDPT-Cox & \ensuremath{0.669_{\textcolor{gray}{\pm 0.013}}} & \ensuremath{0.272_{\textcolor{gray}{\pm 0.041}}} & \ensuremath{0.604_{\textcolor{gray}{\pm 0.030}}} & \ensuremath{0.679_{\textcolor{gray}{\pm 0.011}}} & \ensuremath{0.234_{\textcolor{gray}{\pm 0.028}}} & \ensuremath{0.630_{\textcolor{gray}{\pm 0.038}}} \\
 & TabICL-Cox & \ensuremath{0.616_{\textcolor{gray}{\pm 0.012}}} & \ensuremath{0.278_{\textcolor{gray}{\pm 0.038}}} & \ensuremath{0.585_{\textcolor{gray}{\pm 0.008}}} & \ensuremath{0.621_{\textcolor{gray}{\pm 0.014}}} & \ensuremath{0.235_{\textcolor{gray}{\pm 0.015}}} & \ensuremath{0.593_{\textcolor{gray}{\pm 0.017}}} \\
\midrule
\end{longtable}
\endgroup

\vskip 0.2in
\bibliography{sample}

\end{document}